\documentclass[sn-mathphys,Numbered]{sn-jnl}

\usepackage{graphicx}%
\usepackage{multirow}%
\usepackage{amsmath,amssymb,amsfonts}%
\usepackage{amsthm}%
\usepackage{mathrsfs}%
\usepackage[title]{appendix}%
\usepackage{xcolor}%
\usepackage{textcomp}%
\usepackage{manyfoot}%
\usepackage{booktabs}%
\usepackage{algorithm}%
\usepackage{algorithmicx}%
\usepackage{algpseudocode}%
\usepackage{listings}%
\usepackage{comment}
\usepackage{dblfnote}
\usepackage[hyphenbreaks]{breakurl}
\begin{document}

\title[Article Title]{Federated Battery Diagnosis and Prognosis}


\author[1]{\fnm{Nur Banu} \sur{Altinpulluk}}\email{nurbanu@wayne.edu}

\author[1]{\fnm{Deniz} \sur{Altinpulluk}}\email{deniz@wayne.edu}

\author[2]{\fnm{Paritosh} \sur{Ramanan}}\email{paritosh.ramanan@okstate.edu}

\author[4]{\fnm{Noah} \sur{Paulson}}\email{npaulson@anl.gov}


\author[3]{\fnm{Feng} \sur{Qiu}}\email{fqiu@anl.gov}

\author[4]{\fnm{Susan} \sur{Babinec}}\email{sbabinec@anl.gov}

\author*[1]{\fnm{Murat} \sur{Yildirim}}\email{murat@wayne.edu}

\affil[1]{\orgdiv{Industrial Engineering Department}, \orgname{Wayne State University}, \orgaddress{\street{4815 4th St}, \city{Detroit}, \postcode{48201}, \state{MI}, \country{USA}}}

\affil[2]{\orgdiv{Industrial Engineering \& Management Department}, \orgname{Oklahoma State University}, \orgaddress{\street{354 Engineering North}, \city{Stillwater}, \postcode{74078}, \state{OK}, \country{USA}}}

\affil[3]{\orgdiv{Energy Systems Department}, \orgname{Argonne National Laboratory}, \orgaddress{\street{9700 S Cass Ave}, \city{Lemont}, \postcode{60439}, \state{IL}, \country{USA}}}

\affil[4]{\orgdiv{Stationary Storage Department}, \orgname{Argonne National Laboratory}, \orgaddress{\street{9700 S Cass Ave}, \city{Lemont}, \postcode{60439}, \state{IL}, \country{USA}}}


\abstract{Battery {diagnosis, prognosis} and health management models play a critical role in the integration of battery systems in energy and mobility fields. However, large-scale deployment of these models is hindered by a myriad of challenges centered around data ownership, privacy, communication, and processing. State-of-the-art battery diagnosis and prognosis methods require centralized collection of data, which further aggravates these challenges. Here we propose a federated battery prognosis model, which distributes the processing of battery {standard current-voltage-time-usage data} in a privacy-preserving manner. Instead of exchanging raw {standard current-voltage-time-usage data}, our model communicates only the model parameters, thus reducing communication load and preserving data confidentiality. The proposed model offers a paradigm shift in battery health management through privacy-preserving distributed methods for battery data processing and remaining lifetime prediction.}

\maketitle

\section{Introduction}
%
%
%
%
{
Climate change is a pressing global issue that requires widespread efforts across disciplines to develop technologies capable of significantly reducing or eliminating greenhouse gas emissions. Large-scale adoption of renewable energy sources and electric mobility are expected to be the main drivers toward this goal. The success of this transition hinges on the efficient integration of these technologies into the existing electricity infrastructure, which requires lithium-ion batteries as a vital storage medium, capturing and storing excess energy during peak production periods for use during times of low production or high demand. This energy storage capability is pivotal in maintaining a stable and reliable grid, to mitigate the intermittent nature of generation and demand in these technologies. Such energy storage capabilities are essential for sustaining a stable and reliable grid, particularly in mitigating the intermittent nature inherent in the generation and demand patterns associated with these technologies. As the demand for clean energy and electric mobility rises, ensuring large-scale adoption of lithium-ion batteries becomes paramount in forging a sustainable future and combating the adverse effects of climate change \cite{UN2023},\cite{ward2022principles}.}








{
The industrial adoption of lithium-ion batteries revolves around the implementation of efficient battery health management strategies. Streaming data from the battery systems, such as standard current-voltage-time-usage information, carries a wealth of information and enables sensor-driven methods called \textit{battery diagnosis and prognosis}, that predicts the current battery condition and future trajectory of degradation \cite{severson2019data}. These methods generate significant economic and societal value from streaming standard current-voltage-time-usage information data. Stringent reliability and performance requirements, and an increasing volume of data, further amplify the importance of this data as a critical asset. 
Effective management of this asset class requires new generations of diagnosis and prognosis methods that can ease deployment and reduce implementation risks \cite{EnergyGov2018}.
}



\begin{figure*}[tb]
\centering
\includegraphics[width=0.9\textwidth]{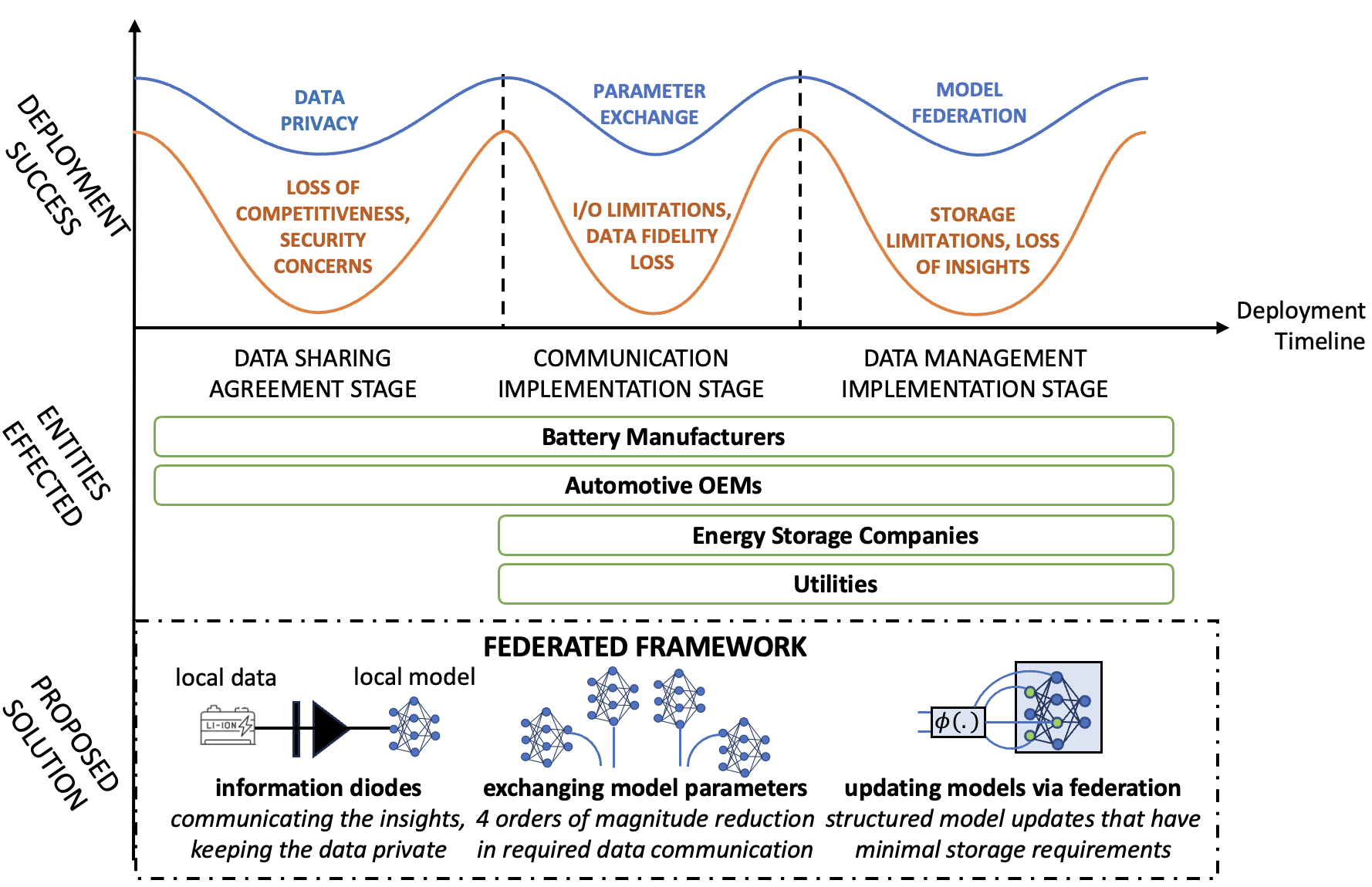}
\caption{An overview of the deployment challenges impacted entities. The proposed federated solution inherently addresses these challenges.}
\label{fig1}
\end{figure*}

Current approaches for battery diagnosis and prognosis build on continuous collection and processing of the {standard current-voltage-time-usage data} in a central server - a class of methods called centralized battery diagnosis and prognosis. 
These approaches require a massive amount of data transfer across batteries in the field and a central database, which congests computation and communication channels, and subjects the operators to significant vulnerabilities in terms of data residency and privacy. 
As highlighted in Fig. \ref{fig1}, these limitations create significant deployment challenges that occur at different stages of the deployment timeline and impact different entities including battery manufacturers, automotive original equipment manufacturers (OEMs), energy storage companies, and utilities. 

%

Data sharing proves to be a critical entry barrier in the deployment of any monitoring system. This barrier becomes particularly restrictive when the battery operator and the central database belong to different entities, as in the example of a battery manufacturer collecting data from different energy companies. In these settings, data sharing raises fundamental security concerns and vulnerabilities in case of data breach incidents \cite{oecdenhancing}, \cite{kaissis2020secure}. Additionally, when there are competing business interests, even minor data leaks can threaten an industrial organization's competitive edge while unleashing devastating legal and financial consequences \cite{cheng2017enterprise}. In our federated diagnosis and prognosis framework, we propose the use of "information diodes" that are situated between the local data and the local model. Like the diodes in electrochemical applications, our information diodes trap the raw data transfer and only enable the insights to pass from local data modules to local models; which inherently ensures privacy of the data. 

Once the data-sharing issues have been successfully addressed and the data architecture is established, conventional diagnosis and prognosis systems confront a second but equally central challenge: data communication capacity. Batteries generate a substantial volume of data, and when this data is transferred to a central database, it can quickly saturate communication channels, owing to inherent input/output limitations \cite{verbraeken2020survey}. Typically, operators resolve this constraint by significantly diminishing the fidelity of the transmitted data. This is typically achieved by performing data averaging over extended periods (e.g., an hour) and subsequently transmitting the processed data to the central dataset. This compromise causes a significant loss of data that carries a wealth of information on battery condition. In our framework, we address this challenge by processing the high-fidelity data locally and communicating only a random subset of parameters across channels connecting the batteries to the central database. This change in operations enables us to harness the full information contained in the data while realizing orders of magnitude reduction in the size of the communicated data. 

The data management and processing stage generates an additional layer of implementation challenges. In conventional diagnosis and prognosis implementations, a central unit collects, stores and processes high volumes of data coming from a fleet of batteries. This operational mode significantly increases the need for processing power in the central node and requires a data storage capability to host an ever-increasing amount of data. To address this issue, operators typically execute periodic data purges, resulting in the loss of certain insights; however, this practice is implemented to maintain data storage requirements within predefined parameters. The proposed federated battery prognosis inherently addresses these problems by distributing the processing and storage of {standard current-voltage-time-usage data} and updating the central model exclusively via structured model updates, which is computationally efficient and requires only minimal storage on the federated server.

\begin{figure*}[tb]
\centering
\includegraphics[width=0.98\textwidth]{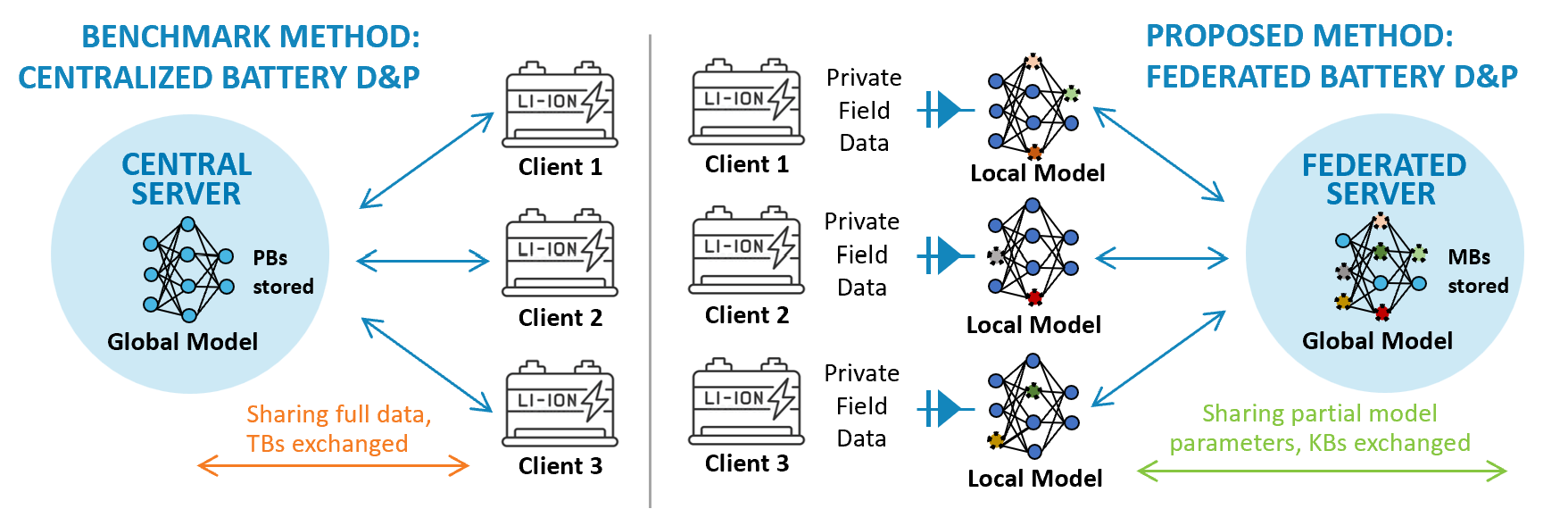}
\caption{Comparative analysis for the conventional centralized battery diagnosis and prognosis models, and the proposed federated battery diagnosis and prognosis models.}
\label{fig1_2}
\end{figure*}

Federated battery diagnosis and prognosis employs a unique perspective, which regards sensor-driven insights as the real asset ({as opposed to {standard current-voltage-time-usage data}), and builds a framework that focuses on collecting insights rather than the data. A comparison of centralized and federated battery models is shown in Fig. \ref{fig1_2}. Centralized battery diagnosis and prognosis require a central server to collect data from every battery (referred to as a client), to train a central server. This mode of operation could incur data storage complexities in the order of petabytes while demanding high bandwidth communication channels capable of streaming data in the order of terabytes. Federated models offer an inherently different framework that focuses on distributing data processing in a privacy-preserving manner to collect insights, rather than data. Specifically, federated diagnosis and prognosis use local {standard current-voltage-time-usage data} to update local models for each battery (or client) and only collect a random subset of the locally updated model parameters (i.e. insights) within a global server. In Fig. \ref{fig1_2}, the updated parameters for the client and federated models are highlighted with dashed lines. Our information diode approach passes the insights but traps the raw data in a local server, hence ensuring data privacy. These model parameters are used to update a federated model that continuously revises the prognosis predictions on battery lifetime. 
}


A central advantage of employing the proposed federated approach lies in its remarkably low entry barrier for deployment, training, and inference. Challenges related to model deployment disproportionately affect businesses of varying sizes. Small-to-medium enterprises with limited resources do not have the capability to invest in high-throughput data communication/computation systems and lack the necessary in-house technical expertise to help deploy and maintain these complex data pipelines and cloud architectures. Even when they invest, they typically have a limited number of assets and failure instances to learn from, thus needing to rely on other entities, such as original equipment manufacturers with access to much more information to generate these insights. 
Nonetheless, it is precisely these enterprises that stand to derive the greatest advantages from the insights that can be harnessed through these systems. They are subjected to the highest risks for not having access to the predictive capabilities of these systems; due to significant operational risks associated with asset failures and limited financial buffers to mitigate these risks.

The overarching research theme in our study attempts to examine whether the battery prognosis method can be fundamentally changed to enable the widespread adoption of our information diode approach. More specifically, our underlying research hinges on three questions: “Can we build federated battery prognosis models that can (i) significantly reduce the computational load by distributing computations across the edge devices and the central server, (ii) require an order of magnitude less communication requirement across the central server and the edge battery devices, and (iii) ensure the privacy of the battery {standard current-voltage-time-usage data} while enabling the prognosis models to learn from a wide range of clients. 
To address this question, we propose an end-to-end federated learning framework for predicting battery remaining life. The proposed framework is composed of two federated stages: The first stage develops a federated-autoencoder model to reduce the dimensionality of the multi-stream {standard current-voltage-time-usage data} space in a privacy-preserving manner. Transformation of input data through federated-autoencoder boosts the performance of the subsequent prognosis task, which revolves around a federated neural network structure to predict battery remaining life. The proposed framework prevents data leakage and enables fully autonomous prognosis tasks, making it applicable to small-to-large industrial systems. The proposed models are evaluated using extensive battery datasets that incorporate accelerated life testing data for lithium-ion batteries, with different chemistries, including HE5050 and NMC532 \cite{severson2019data} \cite{paulson2022feature}.

\section{State of the Art}
\label{2}
In the realm of battery diagnosis and prognosis, various techniques have been employed, including model-based, data-driven, and hybrid models. Model-based approaches employ a physics-based or a mathematical model to represent degradation behavior. These approaches include circuit models \cite{guha2017remaining}, empirical models based on methods such as Kalman filters \cite{duan2020remaining} or particle filtering \cite{walker2015comparison}, \cite{miao2013remaining}, mechanistic models \cite{ramadass2004development} or fused approaches \cite{zhang2017remaining}, \cite{guha2017state}. Data-driven methods shift the focus to the analysis of data trends. The widespread use of sensor technology has provided a vast data source, making data-driven methods more attractive, as they are flexible and capable of implicitly capturing complex degradation behaviors. The main techniques for a data-driven approach to battery diagnosis and prognosis include signal processing \cite{wang2017remaining}, ML techniques such as neural networks \cite{ren2018remaining}, \cite{zhang2018long}, \cite{li2020state}, support vector regression \cite{patil2015novel}, and autoregressive integrated moving average (ARIMA) models \cite{saha2009comparison}. Hybrid models fuse model-based and data-driven approaches to leverage the performance of the prediction models \cite{li2019remaining}, \cite{xue2020remaining}, \cite{zheng2015integrated}. For more extensive reviews on lithium-ion battery diagnosis and prognosis, readers are referred to \cite{hu2020battery}, \cite{li2019data}, \cite{lipu2018review}, \cite{meng2019review}, \cite{ng2020predicting}.

{
Traditional ML methodologies for lithium-ion battery diagnosis and prognosis often entail the transfer of data to a central node for analysis and model training. However, this centralized approach introduces significant risks to data privacy and security while requiring extensive computation and communication bandwidths. To address these concerns, federated learning emerged as a leading method for safeguarding data privacy during the training of ML models. By conducting model training locally on client devices and aggregating only the encrypted model updates, federated learning ensures the protection and decentralization of sensitive data. This makes it an ideal solution for privacy-preserving battery diagnosis and prognosis. Notably, scalable privacy-preserving solutions, including federated learning, have gained momentum in recent years \cite{kairouz2021advances} with emerging applications such as Google GBoard \cite{hard2018federated}. These scalable federated learning solutions have been successfully applied in various domains, including healthcare \cite{rieke2020future}, \cite{pfitzner2021federated}, cybersecurity \cite{huong2021detecting}, manufacturing \cite{ge2021failure}, \cite{mehta2022federated}, and energy systems \cite{saputra2019energy}, \cite{savi2021short}, effectively mitigating risks associated with user data privacy.  To ensure the protection of users' privacy in the implementation of federated learning, it is crucial to address pre-processing steps like dimensionality reduction and feature extraction. Often, these steps are overlooked, which poses a risk of data leakage and compromises the privacy of sensitive information. To mitigate this risk, it is imperative to perform federated dimensionality reduction and feature extraction, keeping the private data at its source and preventing data leakage. Several techniques have been proposed, including federated PCA \cite{grammenos2020federated}, federated autoencoder \cite{novoa2022fast}, federated feature selection based on mutual information \cite{banerjee2021fed}, secure federated feature selection based on $\chi^2$-test protocol \cite{wangsecure}, federated feature selection based on particle swarm optimization \cite{hu2022multi}, and federated tensor decomposition model for feature extraction \cite{gao2021federated}. The utilization of federated feature extraction and selection has extended to various application domains, such as the electricity market \cite{wang2021electricity}, autonomous vehicles \cite{cassara2022federated}, and human activity recognition \cite{xiao2021federated}. By incorporating these methodologies, researchers can maintain the privacy of sensitive data while extracting meaningful features for accurate predictions, thereby upholding the confidentiality and security of client data. Leveraging the inherent advantages of federated learning, this approach holds immense promise in maintaining battery diagnosis and prognosis accuracy while prioritizing data privacy and security, and to the best of our knowledge, it represents an unexplored application of federated learning for lithium-ion battery diagnosis and prognosis.}

Our federated learning strategy is specifically geared towards a low-overhead approach for battery prognosis, that is amenable to the cost and resource constraints of small and medium-sized enterprises as well as OEMs. Our holistic federated learning strategy consists of two distinct phases: federated autoencoder training for dimensionality reduction in a fully distributed setting; and a Federated Deep Neural Network (DNN) training for battery prognosis. {During the first phase, we employ a federated autoencoder to reduce the dimensionality of the feature set to efficiently capture the essential information while minimizing the data transmission overhead. In the second phase, the transformed features are fed into a federated DNN model for predicting the remaining useful life of lithium-ion batteries. Collectively, this framework is referred to as the Federated Holistic Battery Prognosis and Evaluation Framework (HOPE-FED). It offers a holistic solution that addresses the unique needs of battery prognosis in a federated learning setting, providing an effective and practical approach for industry applications.}




\section{Federated Learning Framework for Battery Diagnosis and Prognosis} \label{3}

We consider a conventional battery asset management setting, where the communication and processing of {standard current-voltage-time-usage data} occur across two types of entities. The first entity type, called \textit{the clients}, refers to battery operators that may either be operating a single battery, or a small fleet of batteries. The second entity type is called \textit{the monitoring service providers}. In many applications, these are either original equipment manufacturers, third-party monitoring solution providers, or large-scale fleet operators (e.g. large mobility as service providers). These entities often have access to {standard current-voltage-time-usage data} from a large fleet of battery devices, and provide insights or alerts to clients when their batteries exhibit significant failure risks.

In this section, we propose a holistic federated battery prognosis approach for this setting. Our approach federates both the dimensionality reduction and remaining life prediction tasks. Through a fully federated approach, HOPE-FED strikes a fine balance toward achieving enhanced quality with respect to prognosis across a fleet of batteries owned by multiple stakeholders. A significant benefit of our approach is that it eliminates the need to collect sample datasets for learning the low dimensional embeddings of the underlying distributed performance data. As a result, the entire data analysis pipeline can be federated using our methodology resulting in complete localization of stakeholder datasets.




The proposed HOPE-FED framework has two distinct federation stages: (i) a federated autoencoder for dimensionality reduction, and (ii) a Deep Neural Network (DNN) for predicting the RUL for lithium-ion batteries. Owing to their sequential nature, both phases are able to leverage a similar algorithmic structure along with a shared computational architecture to carry out federation among a diverse set of clients. In the following subsections, we detail the localized dimensionality reduction (DR) training step followed by the local prediction training step before discussing the federated methodology comprising of both steps.

\subsection{Local Autoencoder for Reducing Dimension}
The first stage of the HOPE-FED framework involves modeling a federated autoencoder structure to reduce the dimensionality of the feature set before predicting RUL. Compared to traditional techniques such as principal component analysis (PCA), autoencoders are a powerful type of artificial neural network that can capture nonlinear feature relationships and effectively reduce data dimensionality. Autoencoders consist of an encoder and a decoder, which convert high-dimensional data into low-dimensional encodings and reconstruct the original data using the encodings. By minimizing the reconstruction error, the autoencoder generates accurate encodings that represent the input data with lower dimensions. To construct an autoencoder for dimensionality reduction, the architecture should include a bottleneck layer with smaller dimensionality than the input and output layers. During training, the autoencoder learns to encode the input data in a way that minimizes the reconstruction error. After training, the encoder can be used to obtain the lower-dimensional representation of new input data, which can be utilized for various downstream tasks such as clustering, classification, or visualization. To evaluate the effectiveness of the autoencoder for dimensionality reduction, the reconstruction error of the decoder is assessed, which provides a measure of how accurately the autoencoder can reconstruct the original input from the lower-dimensional encoding.

The autoencoder is trained using the Adam optimizer with a mean squared error (MSE) loss function, with the batch size and the number of epochs set. As the autoencoder learns, it encodes the input data in a way that minimizes the MSE loss between the reconstructed input and the original input. To optimize the weights of the network, the error is backpropagated through the network and the weights are updated using the gradients computed by the optimizer. The \textit{Autoencoder} function (see Methods) summarizes the steps performed during autoencoder training and takes five different inputs, including the network parameters for the encoder $f_{\theta}$ and decoder $f_{\beta}$, the network weights of encoder $w_{\theta}$ and decoder $w_{\beta}$, and dataset for autoencoder training represented with $D$. The network parameters for the encoder $f_{\theta}$ and decoder $f_{\beta}$ include the structure of the respective networks, such as the number of layers, types of activation function, and the number of neural units in each layer. Additionally, the target feature size, which becomes the neuron size of the last layer of the encoder for reducing the input dimension to the desired feature size, is also included in the network parameters. In each epoch and batch of the training, the reconstruction loss value is calculated, and the encoder and decoder network weights, $w_{\theta}$ and $w_{\beta}$, are optimized using the Adam optimizer to improve the model's performance.

\subsection{Local Training Step for RUL Prediction}
Predicting RUL is a regression task that can be achieved using various approaches, including linear regression (LR) and deep neural networks (DNNs). However, LR may have specific limitations in accurately predicting remaining useful life due to its assumptions of a linear relationship between input and output variables, independence and homoscedasticity of errors, and constant relationships over time. On the other hand, DNNs can effectively capture non-linear and complex relationships in the data, making them well-suited for FL approaches in RUL prediction. The reduced set of features obtained from the autoencoder are used as inputs to the DNN, which can be designed with a specific architecture including the number of hidden layers, neurons in each layer, and activation functions. The model is then trained using a dataset that includes compressed features and their corresponding RUL values. To optimize the model's performance, a loss function is defined to measure the difference between the predicted and actual RUL values. The model weights are then updated using the Adam optimizer, and the model is trained for a certain number of epochs. Finally, the trained model is evaluated on a test dataset using metrics such as mean squared error to assess its performance.

The \textit{RUL} function  (see Methods) is designed to train the DNN for predicting the remaining useful lifetime of lithium-ion batteries. This function takes four different inputs: network parameters $f_{\gamma}$, network weights $w_{\gamma}$, compressed battery features, $\mathcal{C}$, and target variables for prediction (actual RUL values), $\mathcal{P}$. Initially, the number of epochs and batch size are set initially, and for each epoch and batch, the input data is fed to the DNN. The loss values are then calculated between predicted RUL values $\hat{\mathcal{P}}$ and actual RUL values $\mathcal{P}$. Based on the loss values, the DNN weights $w_{\gamma}$ are optimized.

\subsection{Federated Prognosis Algorithm}
Our federated prognosis algorithm consists of several steps that involve applying the federated autoencoder and federated RUL prediction sequentially in our FL framework. First, we randomly split the battery dataset \{$D_{1},.., D_{\mathcal{M}}$\} into train \{$D_{1},.., D_{\mathcal{K}}$\} and test sets \{$D_{1},.., D_{\mathcal{L}}$\} and normalize the raw input datasets before initiating the FL framework and prepare the corresponding target RUL values,  $\mathcal{P}_{i}$, $i=1,..,\mathcal{M}$. Next, we set the hyperparameters of the FL algorithm, such as the number of rounds for federated autoencoder and RUL prediction, $\mathcal{T}_{autoencoder}$ and $\mathcal{T}_{RUL}$. Based on our preliminary experiments, we determine the network parameters for the encoder, decoder, and DNN for RUL prediction, $f_{\theta}$, $f_{\beta}$, and $f_{\gamma}$, respectively.

Our federated autoencoder operates in rounds, where each round $t$ involves training a separate model for a random subset of $\mathcal{S}$ batteries from the training set. For each battery $s$ in the subset, we apply the \textit{Autoencoder} function to a randomly sampled R\% of its own data. This function returns the trained weights of the encoder and decoder, $w^{ts}_{\theta}$ and $w^{ts}_{\beta}$, respectively, at round $t$ for battery $s$. These trained weights are then shared with the central node, which aggregates them using a Federated Averaging (FedAvg) function. This function assumes that all participating batteries carry equal weight. The aggregated model weights, $w^{t+1}_{\theta}$ and $w^{t+1}_{\beta}$, are then sent back to the batteries, and a new federation round begins. This process continues until all federation rounds are completed. Once finished,  the encoder and decoder weights are frozen, and we use them to transform all training data into a compressed feature representation. With the compressed feature set, we then initiate federated learning for RUL prediction using the training data.

\begin{figure*}[tb]
\centering
\includegraphics[width = 0.98\textwidth]{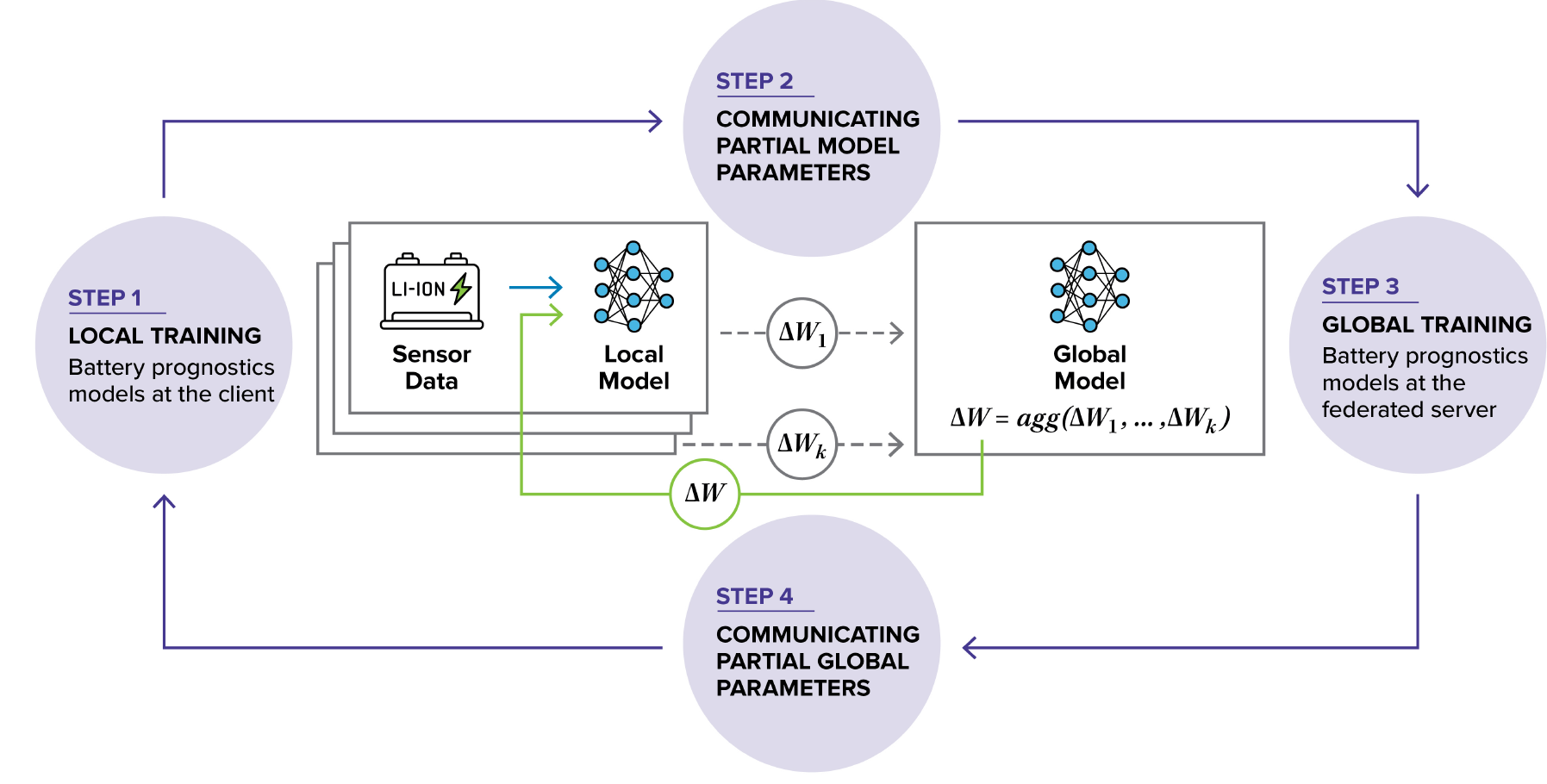}
\caption{Procedural stages of federated learning for battery diagnosis and prognosis}
\label{fig2}
\end{figure*}

In the RUL prediction task, we begin by sampling batteries and their associated datasets that will contribute to the current federation round. We then train a DNN separately for each sampled battery $s$ by calling the \textit{RUL} function. This function returns the trained DNN weights, $w^{ts}_{\gamma}$, for battery $s$ at round $t$. Once all sampled batteries have completed the execution of the RUL function, the DNN weights are aggregated using the FedAvg function until the convergence criteria are satisfied. The federated prognosis weight aggregation process is summarized in Fig. \ref{fig2}. After completing the training for federated RUL prediction, we then test our approach on the test data.  We first feed the test data into the encoder to obtain transformed features $\mathcal{C}$ for every test battery.  We then call the DNN, $f_{\gamma}$, for every battery $l$ to achieve RUL predictions $\hat{\mathcal{P}}_{l}$. Finally, we report the performance measures shared in Section \ref{cost}. Algorithm \ref{alg1} (see Methods) outlines the steps of the federated prognosis algorithm used in the FL process.  

\section{FL Performance Measures}
\label{4}
Our evaluation metric mechanism evaluates the performance in terms of both the predictive accuracy and economic impact. In this regard, two fundamental evaluation metrics are established: \textit{prediction error} \cite{gebraeel2006sensory} and \textit{long-run average cost} \cite{elwany2008sensor}. In Section \ref{cost}, we delve into the intricate details of the performance measures and cost calculation methodology that we have meticulously employed in our HOPE-FED framework. 

Furthermore, beyond the initial development phase, it is crucial to ensure the ongoing {replacement strategies} and monitoring of the HOPE-FED approach. This involves diligently observing its performance to guarantee its continued effectiveness. If the performance level experiences a substantial decline due to notable deviations in the upcoming data stream, it may be necessary to retrain the HOPE-FED model. To address this important aspect, Section \ref{maint} outlines our proposed deployment strategies, designed to sustain the performance and stability of the HOPE-FED framework.

\subsection{FL Performance Measures \& Cost Calculation} \label{cost}
We evaluate the effectiveness of HOPE-FED framework by utilizing two distinct metrics. Firstly, we compute the prediction error for each prediction made throughout the lifespan of the batteries, providing us with insights into the accuracy of our model's predictions. Secondly, we determine the long-run average cost per battery by striking a balance between the costs associated with overestimating and underestimating the battery's RUL. By considering both metrics, we are able to comprehensively evaluate the performance of HOPE-FED framework.

Our first metric to evaluate the efficacy of HOPE-FED involves the computation of prediction errors, which is proposed by \cite{gebraeel2006sensory}. This method facilitates the assessment of the prediction error for each specific forecast rendered at any period during the complete life cycle of the battery. Considering that we predict the RUL of each battery at multiple time periods throughout its lifespan, we calculate the prediction error, $E^k_{i}$, for battery $i$ at time period $k$ using the following formula:
\begin{equation}
    E^k_{i}=\dfrac{(p^{k}_{i}+t^{k}_{i})-t_{fi}}{t_{fi}}
\end{equation}
where $p^k_{i}$, $t^{k}_{i}$, $t_{fi}$ refer to the RUL prediction for battery $i$ at time period $k$, its current age at time period $k$, and its failure time, respectively. The $E^k_{i}$ measure indicates the deviation of our RUL prediction from the true failure time, while also considering the current age of the battery. This measure encompasses both positive and negative values, indicating instances of overestimation or underestimation of the failure time, respectively. In an industrial context, particularly in applications like electric vehicles, overestimating the battery's lifespan can have undesired consequences, leading to unexpected failures without prior warning, resulting in costly battery replacements and reputational damage for the manufacturer. Conversely, premature battery replacements can burden the client with unnecessary expenses, which is unsustainable from an economic standpoint. Hence, our investigation delves into the implications of our predictions and incorporates the calculation of cost rates associated with both overestimation and underestimation scenarios.

Our second metric, the long-run average cost, encompasses the total cost incurred per battery per period, including both the battery cost itself and { associated replacement expenses}. In assessing the effectiveness of our prediction policies, it is crucial to consider not only the battery cost but also the significant role played by { replacement} costs. The prediction of the RUL presents a valuable opportunity for devising efficient { battery replacement} strategies, optimizing resource allocation, and enhancing overall system reliability.

Traditionally, {{ battery replacement} activities have been carried out based on {periodic policies}, where fixed intervals dictate when {replacement} is performed, regardless of the battery's health status. In contrast, predicting RUL enables the implementation of {predictive replacement policies}, which offers a dual advantage of cost-effectiveness and improved user satisfaction. Specifically, when the RUL prediction for a battery reaches a certain threshold, {replacement} activities can be automatically triggered. This approach empowers users to avoid unnecessary replacement costs by automatically triggering { replacement} activities when the RUL prediction for a battery reaches a certain threshold. It ensures proactive {replacement strategies}, minimizing the risk of failures and optimizing the lifespan of batteries.

Nevertheless, it is important to acknowledge that even with {periodic and predictive replacement policies} in place, there is still a possibility of some batteries experiencing failures before the m{ replacement activities} are triggered. In such circumstances, catastrophic failures can occur, rendering the battery unavailable while awaiting the arrival of the {service crew}. These situations result in costly {replacement} operations compared to cases involving early replacement. Therefore, it highlights the significance of effective prediction policies in minimizing such costly occurrences and ensuring the smooth operation of battery systems.

The long-run average cost metric is computed by considering distinct cost values for the early replacement case $c_r$ and catastrophic failure case $c_f$. When the RUL prediction reaches the threshold value, $\delta$, {replacement operations} are triggered. Let's assume this event occurs at time $t^*$, and the {service crew} arrives after $t_{c}$ time periods. If the battery is still functional upon its arrival, early replacement is performed, effectively preventing catastrophic failures. However, in scenarios where the battery fails before the RUL prediction reaches the threshold value, $\delta$, or while the crew is en route, catastrophic failures arise, resulting in significantly higher replacement costs. 
Moreover, premature battery replacement is considered undesirable as it undermines the optimal utilization time of the batteries. Hence, we employ the calculation of the long-run average cost, considering its lifespan and replacement time, to ensure optimal utilization of battery life while minimizing replacement costs. Algorithm \ref{alg2} (see Methods) provides a detailed outline of the steps involved in computing the long-run average cost per battery when implementing predictive {replacement} policies. This approach enables effective decision-making in selecting appropriate {service crew} replacement policy}.

\subsection{Deployment Strategies for FL} \label{maint}
Deployment of FL algorithms is a critical aspect of ensuring their continued performance and effectiveness. As data changes over time, FL models may become outdated and lose accuracy, requiring regular updates to the algorithms. This process typically involves monitoring the model's performance, identifying and addressing any issues that may arise, and retraining the model with updated datasets. 

To maintain the efficacy of HOPE-FED, it is essential to establish a set of rules for triggering retraining. We use the long-run average cost per battery measure, as explained in Algorithm \ref{alg2} (see Methods), as a key indicator of performance. An increase in cost rates signifies potential issues with the algorithm's performance. To address this, we set a threshold value $\alpha$, for the long-run average cost rate. When the long-run average cost exceeds $\alpha$, it serves as an alert for the decision-maker, prompting the retraining of the model to enhance its performance. By implementing this approach, we ensure that HOPE-FED remains a reliable tool for accurately predicting the RUL of lithium-ion batteries, consistently delivering valuable insights for {replacement} strategies.


\section{Computational Experiments}
\label{5}
\begin{figure*}[tb]
\centering
\includegraphics[width=0.98\textwidth]{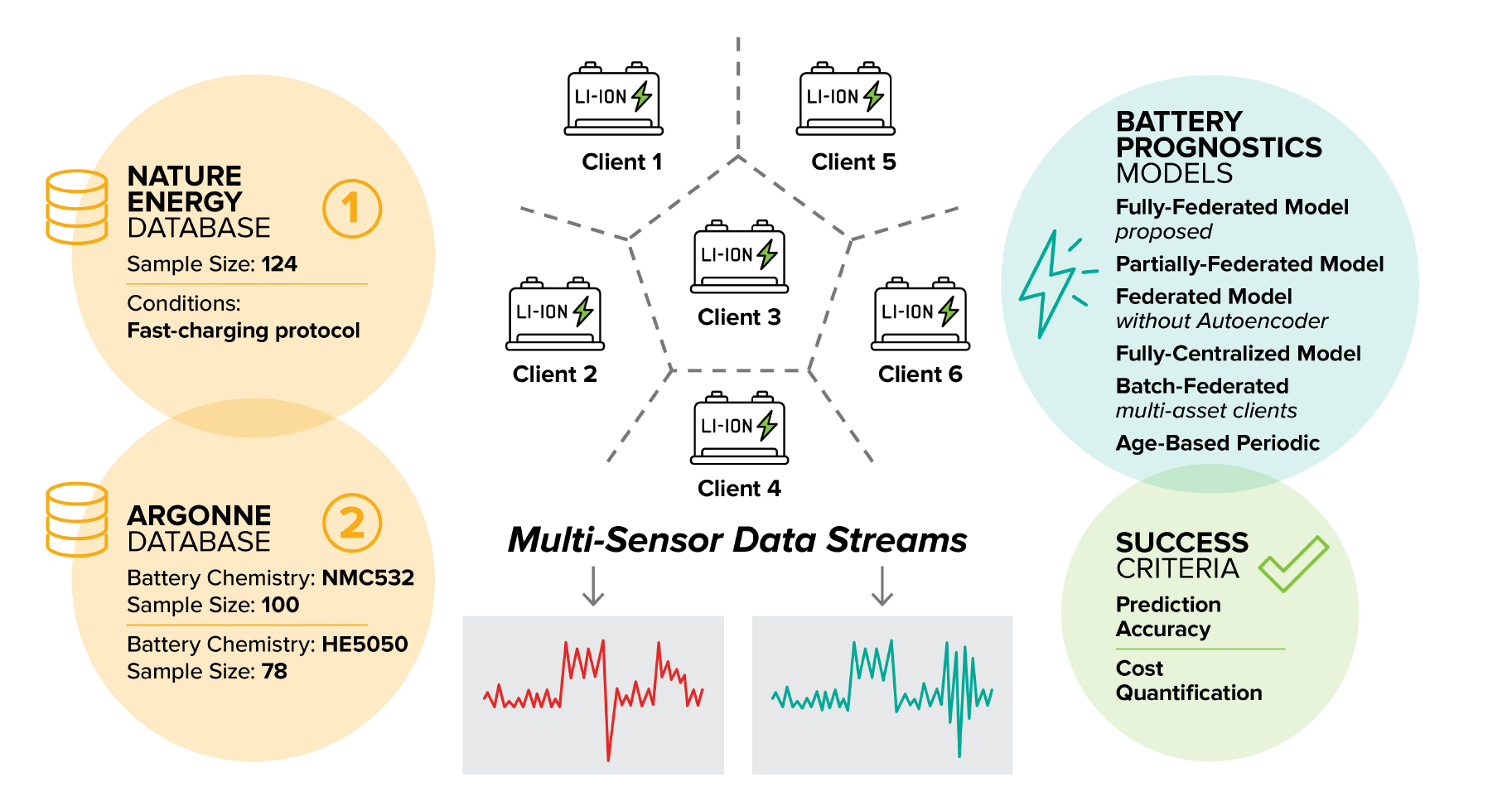}
\caption{Synopses of computational experiments} 
\label{fig3}
\end{figure*}

Our evaluation of the \textit{HOPE-FED} encompasses a meticulous analysis conducted on two distinct databases, Nature Energy \cite{severson2019data} and Argonne Databases \cite{paulson2022feature}, where we compare our findings against a range of benchmark studies. In the first set of benchmark studies, we assess the performance of both {\textit{age-based periodic replacement policy}} and {\textit{HOPE-FED based predictive replacement policy}}. The HOPE-FED based {predictive replacement} policy leverages the prediction information obtained through the \textit{fully-federated model}. Turning to the second set of benchmark studies, we conduct a comprehensive comparison involving the \textit{HOPE-FED approach} and three distinct benchmark studies. The first benchmark study entails executing all dimensionality reduction and RUL prediction tasks using a centralized approach, following the traditional paradigm of machine learning referred to as the \textit{fully-centralized model}. In the second benchmark study, the computational experiments are performed without dimensionality reduction via autoencoder to demonstrate the effect of autoencoder in the performance. This particular case study is referred to as \textit{FL model w/o autoencoder}. Moving on to the third benchmark study, we adopt a \textit{partially-federated model}, wherein the autoencoder is applied in a centralized manner while RUL estimation is performed using a federated approach. Lastly, in the final benchmark study, we implement a partial application of the federated learning (FL) framework by partitioning batteries into groups and preserving the privacy of each group, thereby representing the \textit{batch-federated approach}. This approach serves as a middle ground, striking a balance between the fully federated and fully centralized approaches. In the final analysis, we assess the performance of HOPE-FED in contrast to the benchmark studies, using two essential success criteria: prediction error and long-run average cost. The overview of our computational experiments is depicted in Fig. \ref{fig3}, encapsulating the comprehensive computational experiments.



\subsection{Experimental Setup}
\label{arch}


To establish a robust FL architecture, we utilize a high-performance computing (HPC) cluster, where each client's server is represented by separate nodes. This design allows an HPC node to access the data of a specific client and train its corresponding model. By employing a distributed computing framework using the Message Passing Interface (MPI), we allocate different nodes in the HPC cluster to represent each client. MPI serves as a standardized means of information transfer between multiple devices, facilitating synchronization and communication among parallel nodes in a constrained setting. For our implementation, we employed OpenMPI along with mpi4py to initiate and manage multiple distributed memory client processes, accurately simulating user devices in real-world field scenarios.

Within our FL framework, we leveraged the Keras and TensorFlow libraries to construct the autoencoder and DNN models for RUL prediction. The autoencoder structure includes two layers for both the encoder and decoder, excluding the input layers. 
For RUL prediction, the DNN consists of seven layers. We performed hyperparameter tuning on both the federated autoencoder and RUL prediction tasks to improve their performances. 
Furthermore, we tailored the hyperparameters of the FL algorithm by exploring variations in the number of federation rounds, the ratio of sampled batteries, and the ratio of sampled data points from each battery per round. Throughout both phases, we employed the Adam optimizer and utilized the mean squared error loss function. Our initial experiments guided us to set the target feature size for the encoder as 30 for the Nature Energy Database and 40 for the Argonne Database.
\subsection{Datasets}
\label{impl}
We conducted computational experiments using two distinct datasets. The first dataset, referred to as the Nature Energy Database throughout the paper, was provided by Severson et al. \cite{severson2019data}. This publicly available dataset comprises 124 commercial lithium-ion batteries that were subjected to fast-charging conditions. The second dataset, referred as the Argonne database, was obtained from Paulson et al. \cite{paulson2022feature} and was collected at the Argonne Cell Analysis, Modeling, and Prototyping (CAMP) facility. The Argonne database consists of 300 batteries with six different metal oxide cathode chemistries, namely NMC111, NMC532, NMC622, NMC811, HE5050, and 5Vspinel. For our experiments, we focused on the HE5050 and NMC532 chemistries and developed separate FL models for each of them. For more detailed information about the Nature Energy and Argonne Databases, please refer to Appendix \ref{dataa} and \ref{datab}, respectively.

Using our computational approach, we predicted the RUL of lithium-ion batteries at each cycle, allowing us to continuously monitor their health status. To generate the input data for our FL framework, we incorporated various relevant features, including charge and discharge capacity, temperature, and charging time. Furthermore, we derived additional features from the raw data of each cycle, such as the minimum discharge capacity observed. For more detailed information about the process of feature generation, please refer to Appendix \ref{fg}.

For both datasets, we followed similar experimental settings. Since the cycle lives of the batteries in both datasets exceeded 100, we did not activate the RUL prediction during the first 100 cycles. Instead, we collected information from these initial cycles and created various features for the dimensionality reduction task. However, we note that this threshold can be adjusted for different datasets, as RUL prediction can be done even in the early stages of a battery's lifetime, as indicated by Severson et al. Additionally, we normalized both datasets prior to leveraging dimensionality reduction. We split both datasets into train and test sets, with the train set consisting of 75\% of the data and the test set consisting of 25\%.

\subsection{HOPE-FED computational experiments}
\label{comp1}
\begin{figure*}
\centering
\includegraphics[width=.33\textwidth]{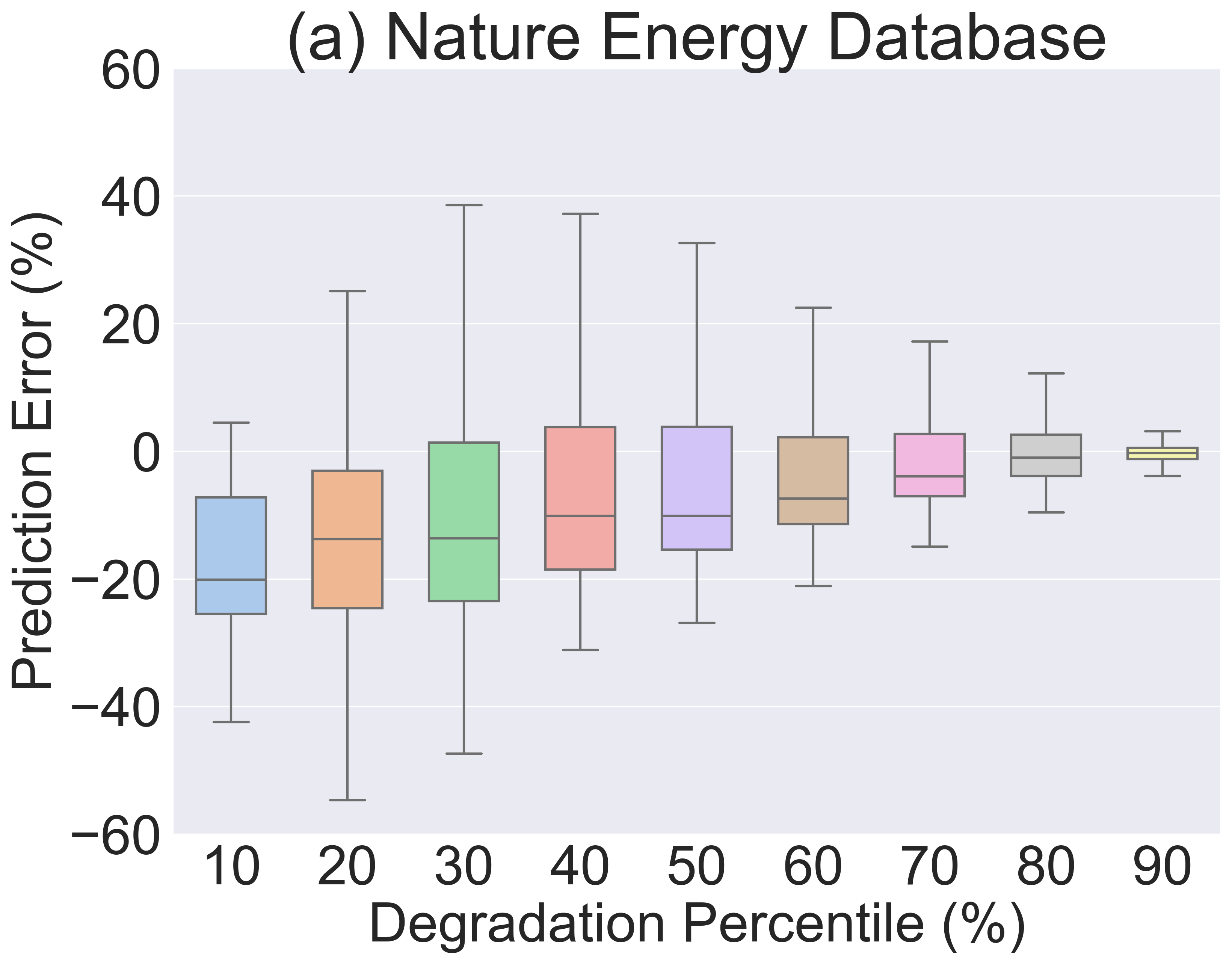}\hfill
\includegraphics[width=.33\textwidth]{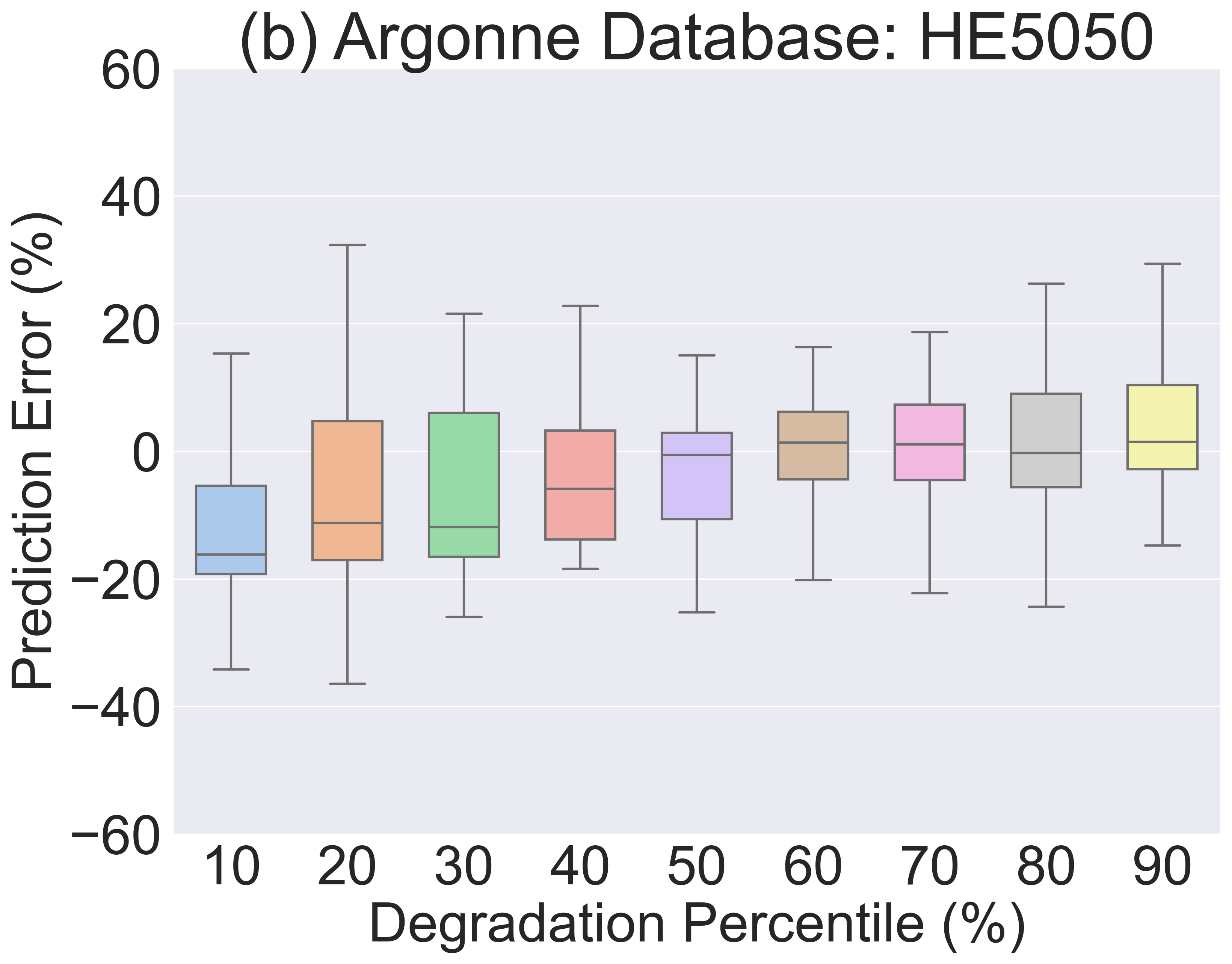}\hfill
\includegraphics[width=.33\textwidth]{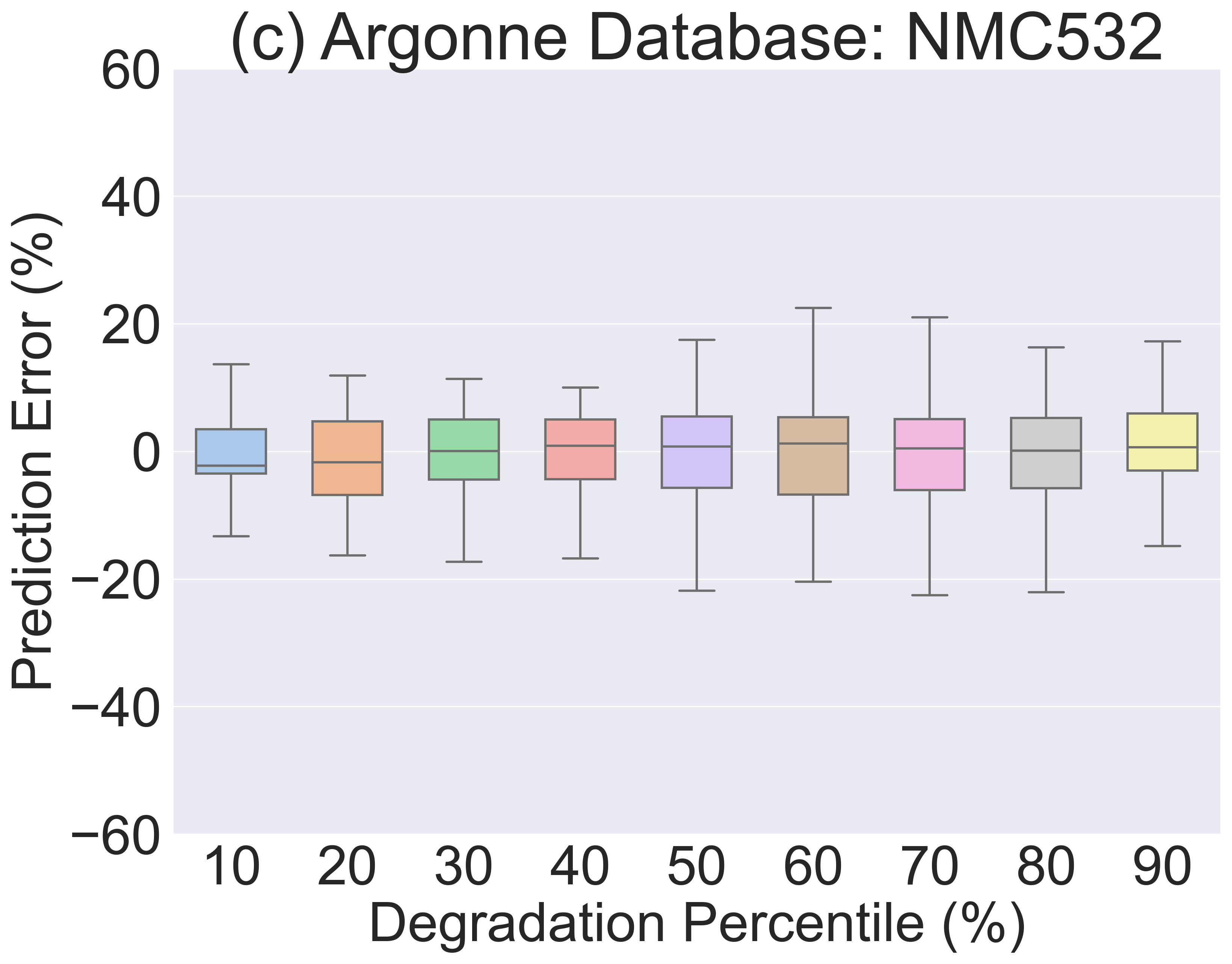}
\caption{Prediction error values across different degradation percentiles for the HOPE-FED approach}
\label{fig4}
\end{figure*}

The HOPE-FED approach was evaluated using test sets generated from two databases, as discussed in Section \ref{impl}. Since the Argonne Database comprises two different chemistries, three separate results were presented to demonstrate the effectiveness of the HOPE-FED approach. The prediction error was assessed across three distinct datasets and plotted against various degradation levels. Examining the prediction error at different stages of degradation is crucial because RUL prediction can be performed at any point in a battery's lifetime. Lower degradation levels indicate that the battery is in its early stages, similar to a brand-new condition. Conversely, higher degradation levels signify that the battery is approaching failure, making accurate predictions more critical for effective {replacement} scheduling. Unsuccessful predictions at higher degradation levels can result in catastrophic failures in some batteries. Fig. \ref{fig4} displays the prediction error values across degradation percentiles ranging from 10\% to 90\%. The absolute average prediction errors for the Nature Energy, Argonne:HE5050, and Argonne:NMC532 test sets are 11.3\%, 13.0\%, and 6.9\%, respectively, indicating the performance of the HOPE-FED approach.

In the Nature Energy Database, the absolute average prediction error values consistently decrease as the degradation percentile increases. This trend indicates that the predictions improve as the battery approaches failure. Such a circumstance is advantageous for conducting { replacement} activities efficiently, as it allows for balancing the tradeoff between battery replacement cost and the opportunity cost associated with early battery replacements. By accurately predicting the RUL of batteries nearing failure, decision-makers can optimize their {replacement} strategies and make informed choices regarding replacement.

In the case of the Argonne Database, no specific trend is observed between the absolute average prediction error and the degradation percentile. However, when considering the higher degradation percentile levels for both the HE5050 and NMC532 chemistries, the absolute average prediction error shows a positive value. This suggests that the HOPE-FED approach tends to slightly overestimate the RUL of batteries nearing failure. While this may seem counterintuitive, it actually serves a beneficial purpose by helping to prevent catastrophic failures. Given that catastrophic failures incur significantly higher costs, a slight overestimation in the lifetime of these batteries can be advantageous for minimizing the overall cost rates. By inclining on the side of caution and providing a buffer in the predicted RUL, the HOPE-FED approach effectively mitigates the risk of costly failures and supports more informed decision-making in {replacement} operations.


In Sections \ref{comp2}, \ref{comp3}, and \ref{comp4}, the HOPE-FED approach is compared to selected benchmark policies. This comparison is based on average long-run average cost rates, as determined by Algorithm \ref{alg2} (see Methods). Additionally, we provide insights into the number of early replacements and catastrophic failures, as well as the average unused life and average number of unavailable days. The average unused life refers to the extent to which battery replacement occurs before failure, indicating the level of earliness. Conversely, the average number of unavailable days represents the average duration during which a battery remains inactive due to replacement operations. This calculation takes into account both the time required for battery replacement and the waiting time for the crew. For a comprehensive understanding of the algorithms used to calculate the average unused life and average number of unavailable days, please refer to Algorithms \ref{alg4} and \ref{alg5} in Appendix \ref{A2}. These algorithms provide detailed steps for determining these important metrics, which contribute to evaluating the performance and effectiveness of the HOPE-FED approach in comparison to the benchmark policies.

\subsection{Comparative Analysis: HOPE-FED Approach vs. Age-based Periodic {Replacement} Policy}
\label{comp2}
\textit{Age-based periodic {replacement} policy} (APRP) dictates performing the {replacement} activities in the same time period for all of the batteries, without considering any prediction information to determine the {replacement} trigger time. To establish the threshold level  $t^*$ for triggering {replacement}, the long-run average cost rate for training set is minimized by following the steps of Algorithm \ref{alg3} (see Methods). For each potential {replacement} trigger time $t^*$, we calculate the total cost rate over all training batteries. The long-run average cost rates are computed in the same manner as the long-run average cost rate calculation in the predictive {replacement} policy, as described in Algorithm \ref{alg2} (see Methods) which include both catastrophic failure and early replacement costs. We select the $t^*$ value that minimizes the total long-run cost rate as the optimal {replacement} trigger time, and all battery {replacement} activities for the test datasets are scheduled to occur at time $t^*$.

We compare the performances of age-based { periodic-replacement and predictive-replacement policies} at Table \ref{periodic} for Nature Energy Database, and Argonne Database for chemistries HE5050 and NMC532, respectively. The best-performing thresholds are reported in Table \ref{periodic},  while more comprehensive results across various threshold levels can be found in Tables \ref{periodic-nature}, \ref{periodic-he5050}, and \ref{periodic-nmc532} in Appendix Section \ref{periodic_additional}. Table \ref{periodic} presents the {replacement} trigger time, the number of preventive and corrective {replacements}, the average unused life, the average number of days unavailable due to {replacement activities}, and the long-run average cost rates across various policies and datasets. The best performing HOPE-FED based predictive-{replacement} policy demonstrated a long-run average cost rate improvement of 38\%, 21\%, and 22\% compared to the age-based periodic-{replacement} policy for the Nature Energy Database, Argonne Database with HE5050 chemistry, and Argonne Database with NMC532 chemistry, respectively. For the Nature Energy Database, the number of catastrophic failures, average unused life, and average number of unavailable days were reduced compared to the \textit{age-based periodic-{replacement} policy}.
For the HE5050 chemistry in the Argonne Database, the HOPE-FED based { predictive-replacement} policy exhibited the same number of catastrophic failures, number of early replacements, and average number of unavailable days as the { periodic-replacement policy}. However, the average unused life improved, indicating that the customized predictions per battery helped prevent premature {replacement}. For NMC532 chemistry, the number of catastrophic failures increased by 1, whereas average unused life improved significantly which paves the way for improving the long-run average cost. 

Overall, the HOPE-FED based { predictive-replacement} outperformed the age-based {periodic-replacement} policy in all datasets. By leveraging predictions to schedule {replacement} activities, we observed enhanced cost rates and improved performance measures. This approach allowed for the creation of customized {replacement} plans for each battery by monitoring {standard current-voltage-time-usage information} effectively.

\begin{table*}[h]
  \centering
  \caption{Comparison of age-based periodic replacement vs. fully-federated prediction-based replacement approaches}
  \resizebox{\textwidth}{!}{
    \begin{tabular}{p{0.25\linewidth}  p{0.09\linewidth} p{0.09\linewidth} p{0.09\linewidth} p{0.09\linewidth} p{0.09\linewidth} p{0.09\linewidth} }
    \hline  
     \multicolumn{1}{l}{} &\multicolumn{2}{c}{\textbf{Nature Energy}} & \multicolumn{2}{c}{\textbf{Argonne: HE5050}} & \multicolumn{2}{c}{\textbf{Argonne: NMC532}}\\
    \hline
   \multicolumn{1}{l}{} & \multicolumn{1}{c}{\textbf{APRP}} & \multicolumn{1}{c}{\textbf{HOPE-FED}} & \multicolumn{1}{c}{\textbf{APRP}} & \multicolumn{1}{c}{\textbf{HOPE-FED}} & \multicolumn{1}{c}{\textbf{APRP}} & \multicolumn{1}{c}{\textbf{HOPE-FED}} \\
   \hline
    \textbf{Trigger Time} & 451   & \multicolumn{1}{l}{Predicted} & 1013 & \multicolumn{1}{l}{Predicted} & 593 & \multicolumn{1}{l}{Predicted} \\
    \hline
    \textbf{\# Preventive} & 29   & 30    & 10    & 10   & 18 & 17    \\
    \hline
    \textbf{\# Corrective} & 2 &1 & 9 &9 & 7 &8 \\
    \hline
    \textbf{Unused Life} & 444.5 & 20.3 & 262.9 &114.2 & 551.8 &95.8\\
    \hline
    \textbf{Unavailable Days} & 1.3 & 1.2 & 3.4 &3.4 & 2.4 & 2.2 \\
    \hline
    \textbf{Cost Rate} & \textbf{20.3} & \textbf{12.6} & \textbf{32.5} & \textbf{25.7} & \textbf{26.5}  & \textbf{20.6}\\
    \hline
    \end{tabular}}
  \label{periodic}%
\end{table*}%
\subsection{Comparative Analysis: HOPE-FED Approach vs. Major Benchmarks}
\label{comp3}
\begin{figure*}

\centering
\includegraphics[width=.25\textwidth]{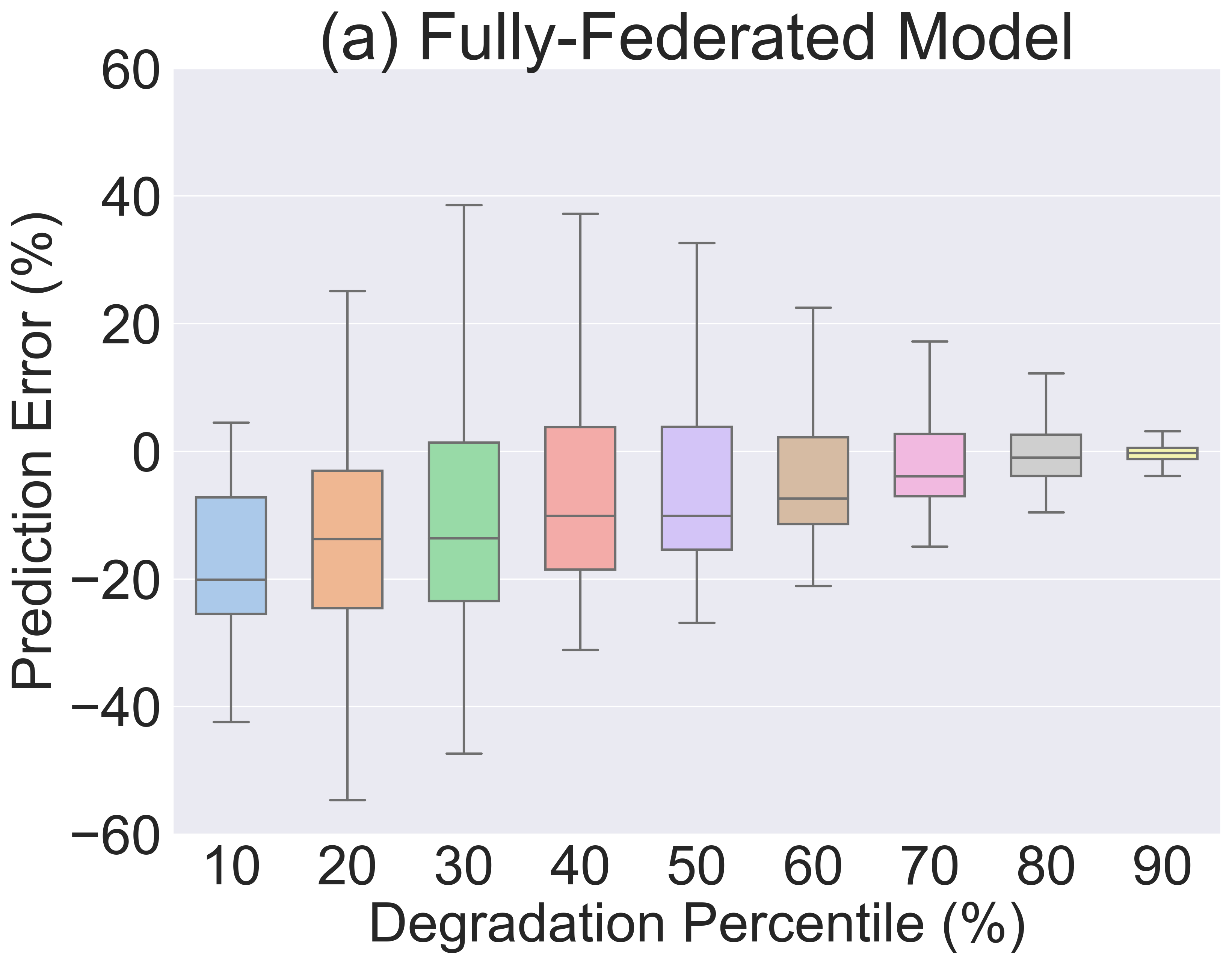}\hfill
\includegraphics[width=.25\textwidth]{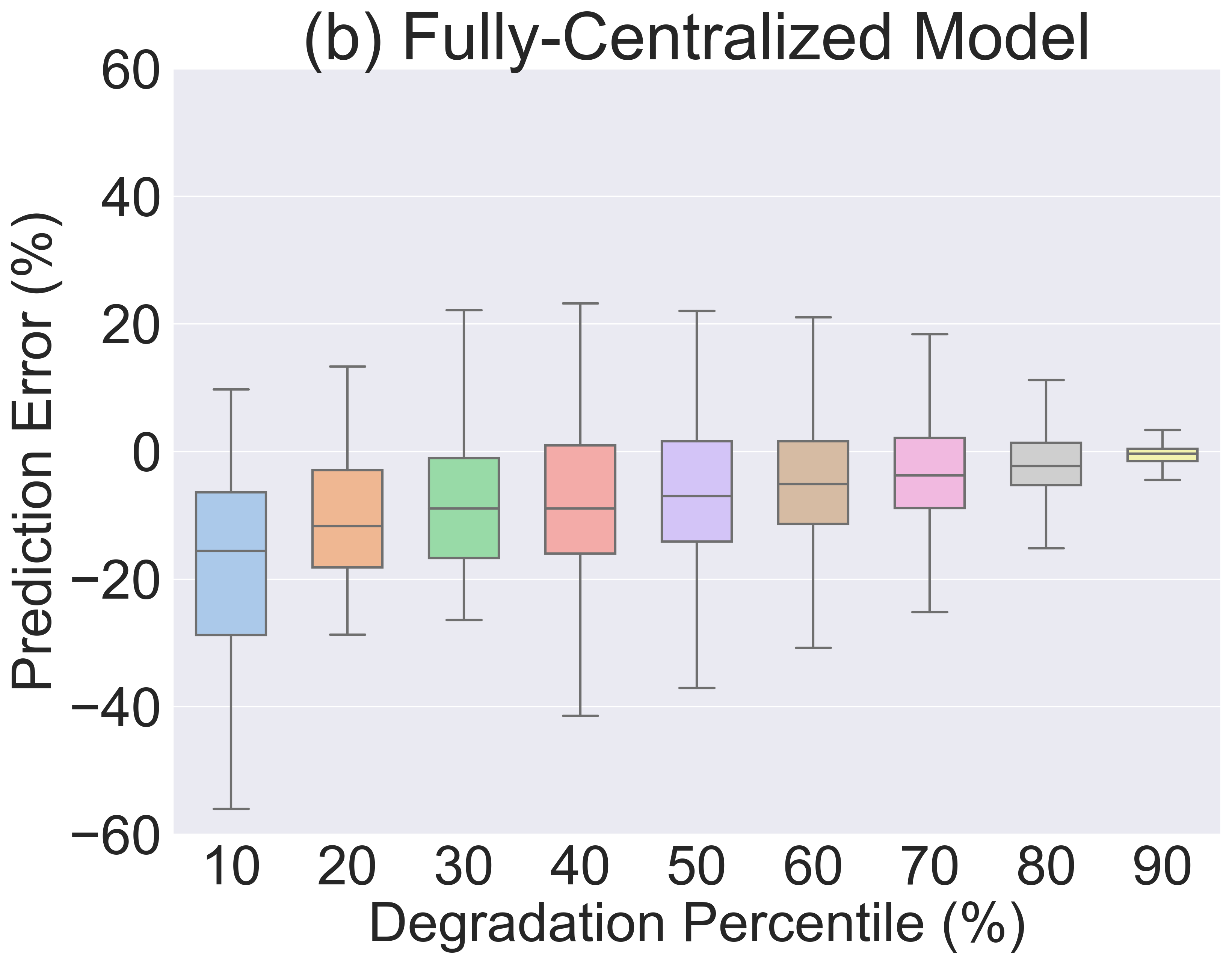}\hfill
\includegraphics[width=.25\textwidth]{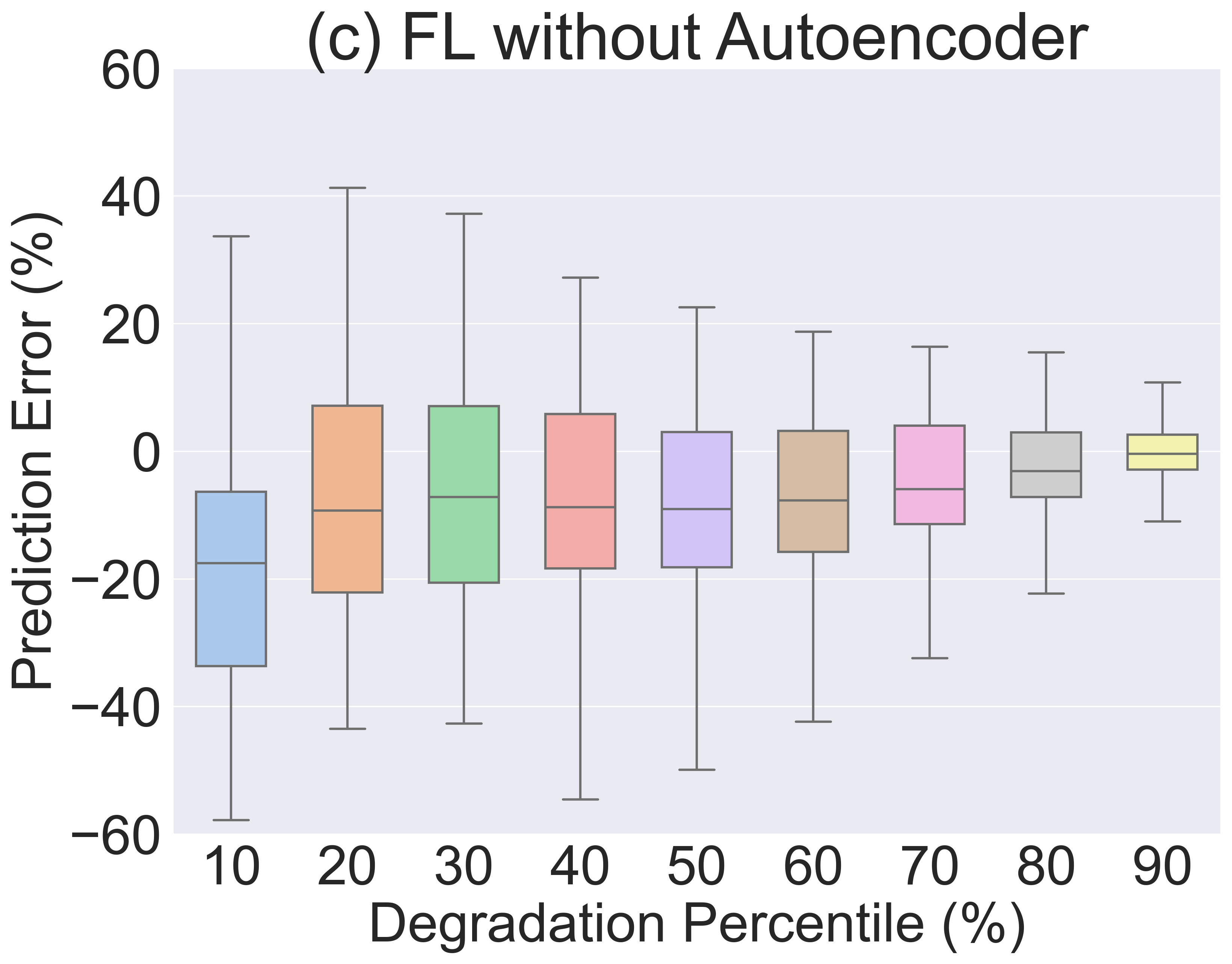}\hfill
\includegraphics[width=.25\textwidth]{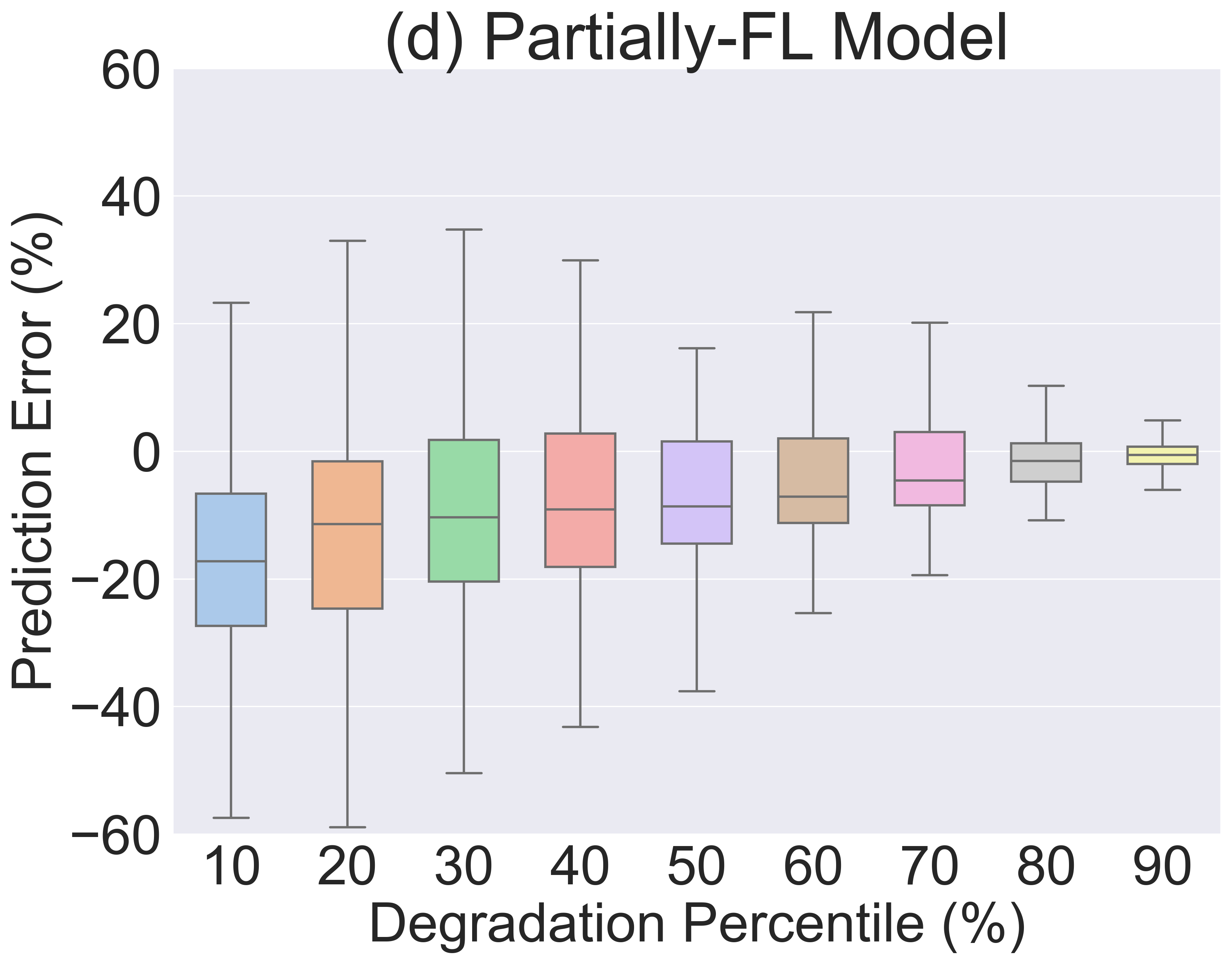}

\caption{Comparison of benchmark models for Nature Energy Database}
\label{fig:bench_2_nature}
\end{figure*}

\begin{figure*}
\centering
\includegraphics[width=.25\textwidth]{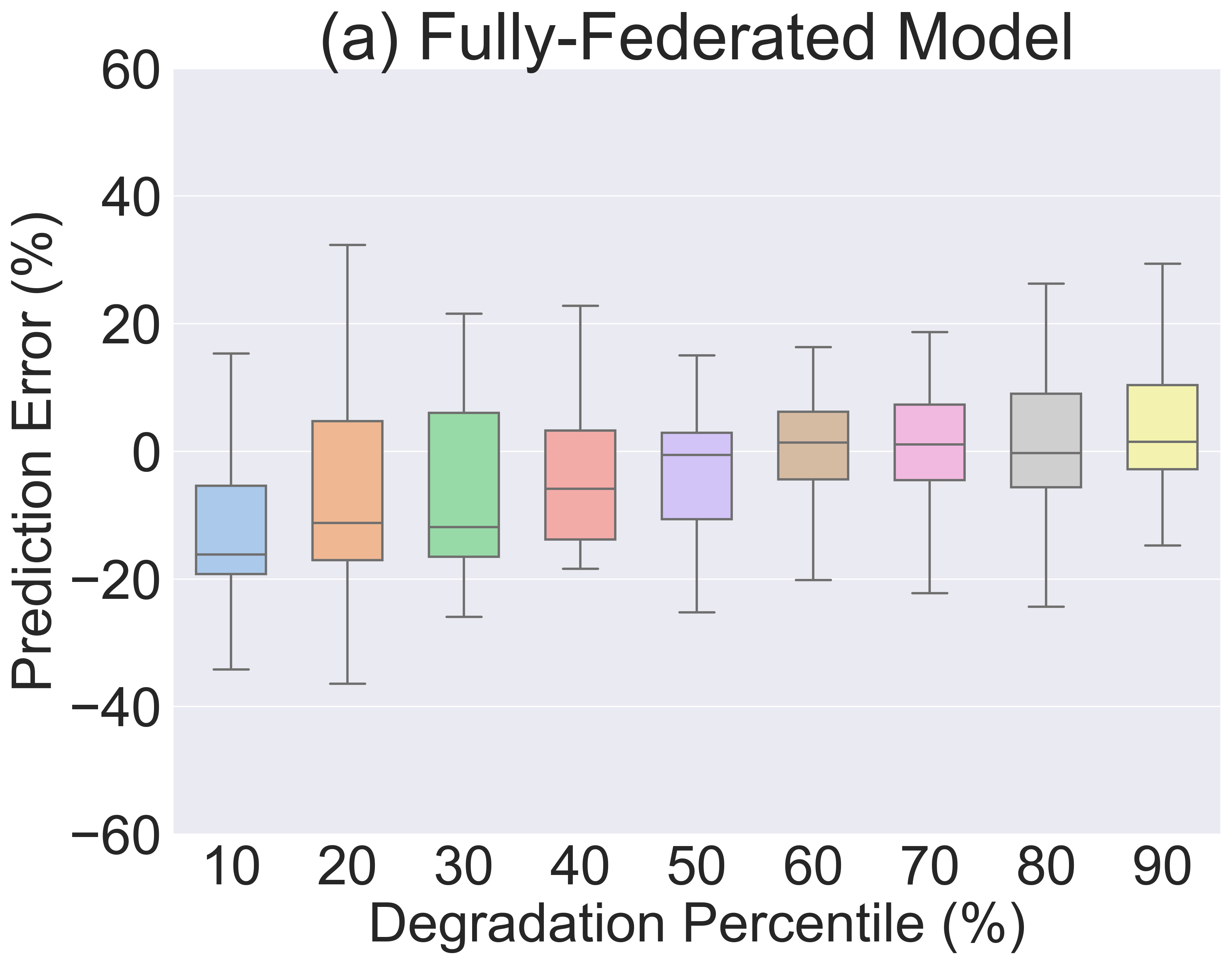}\hfill
\includegraphics[width=.25\textwidth]{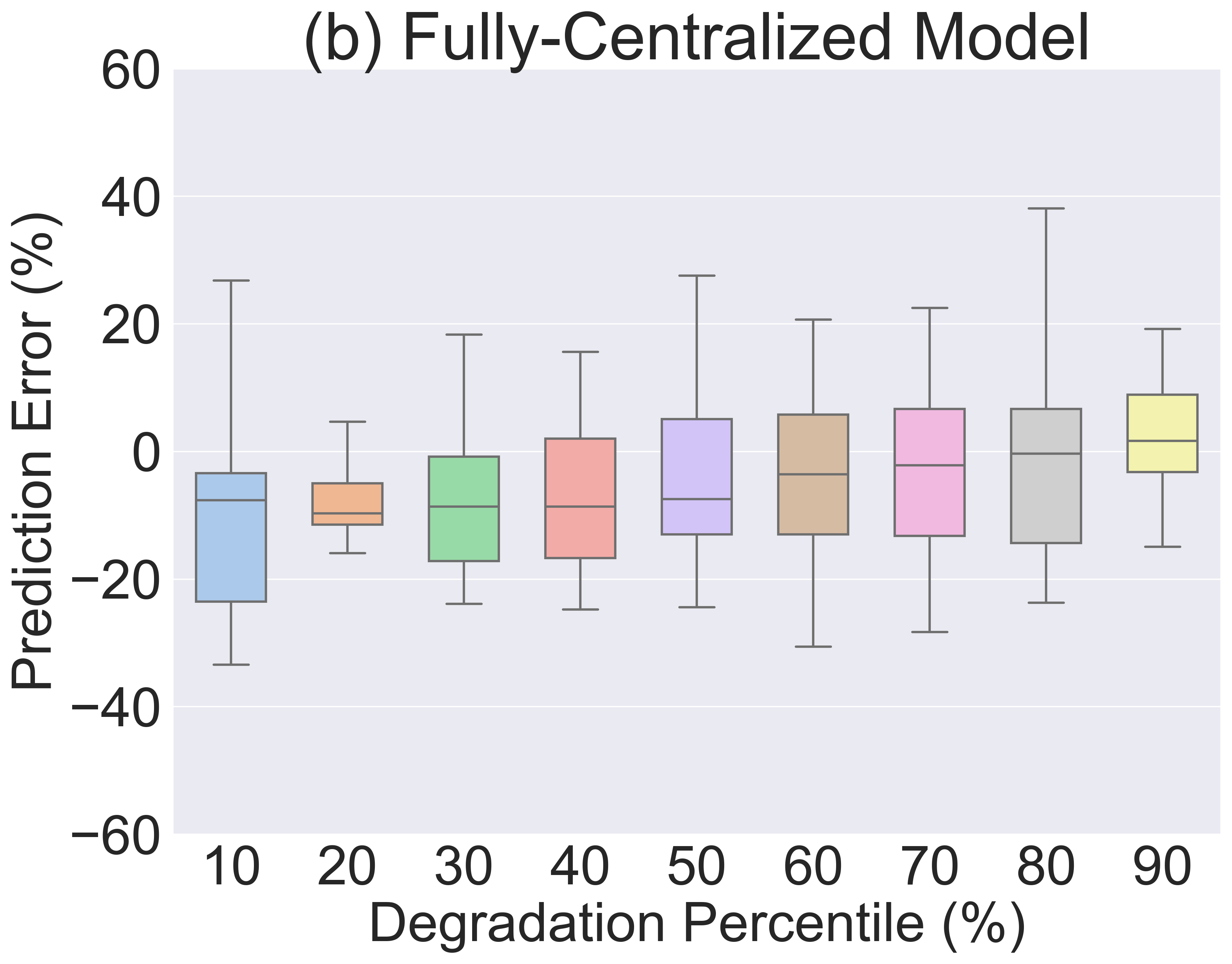}\hfill
\includegraphics[width=.25\textwidth]{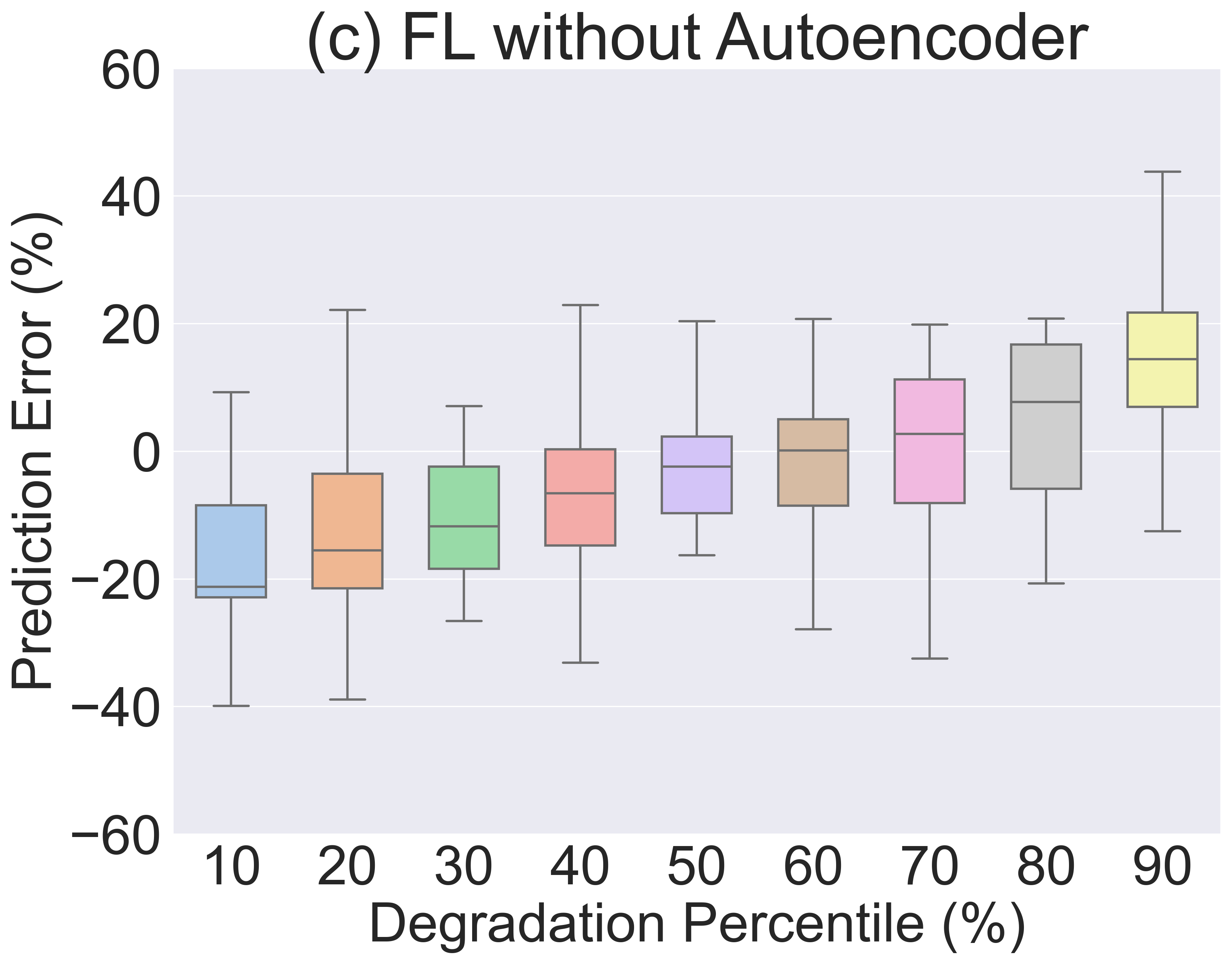}\hfill
\includegraphics[width=.25\textwidth]{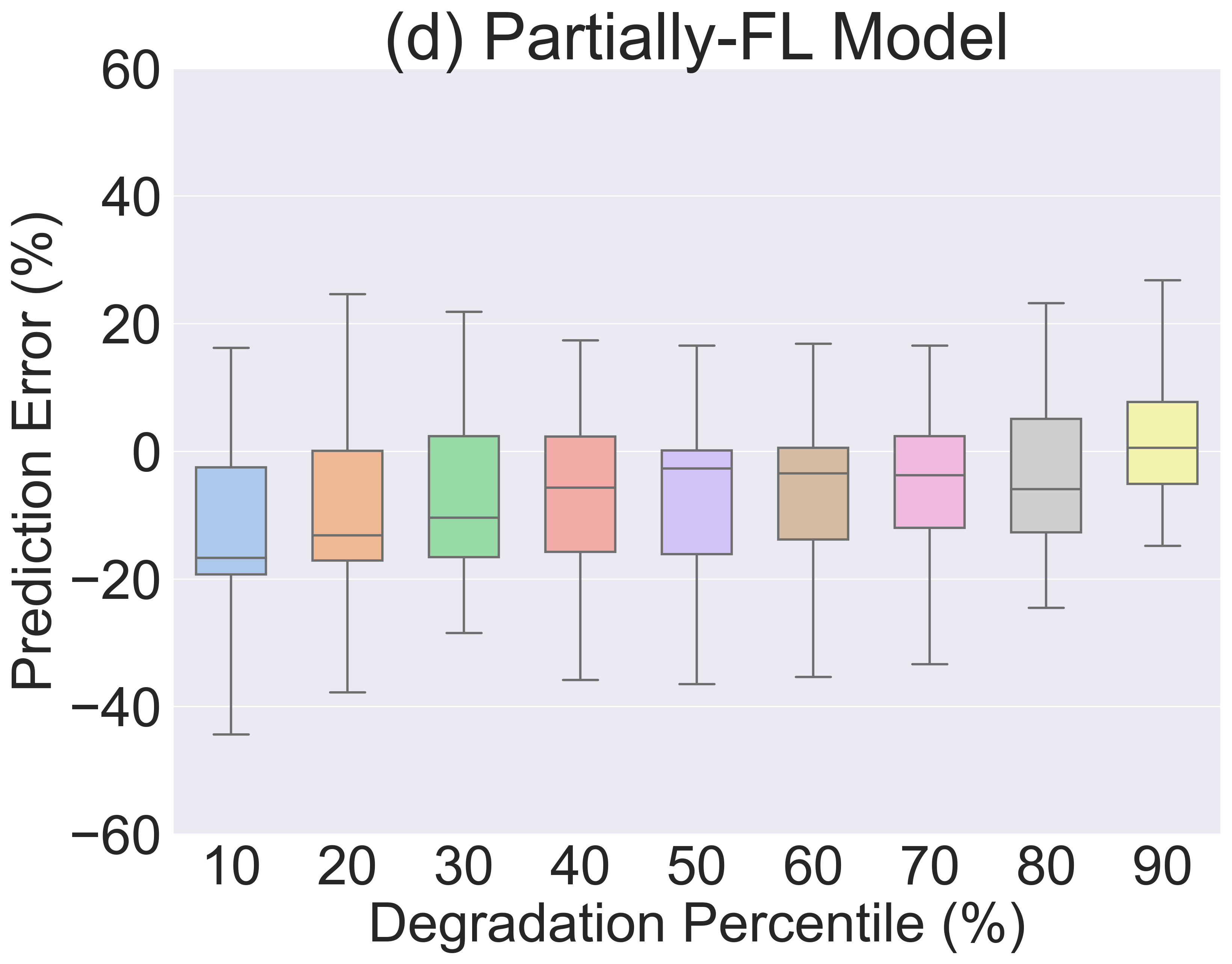}
\caption{Comparison of benchmark models for Argonne Database:HE5050}
\label{fig:bench_2_he5050}
\end{figure*}

\begin{figure*}

\centering
\includegraphics[width=.25\textwidth]{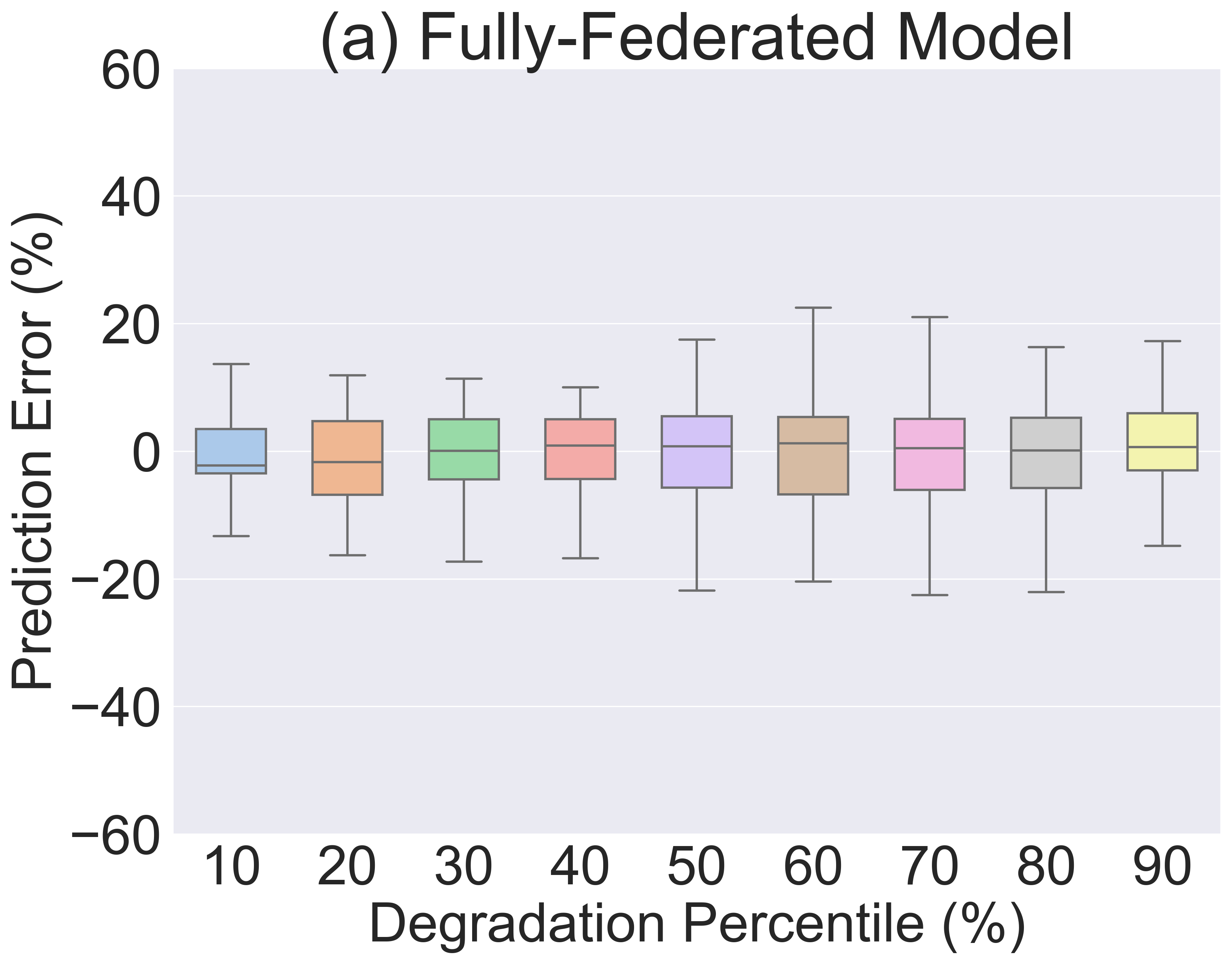}\hfill
\includegraphics[width=.25\textwidth]{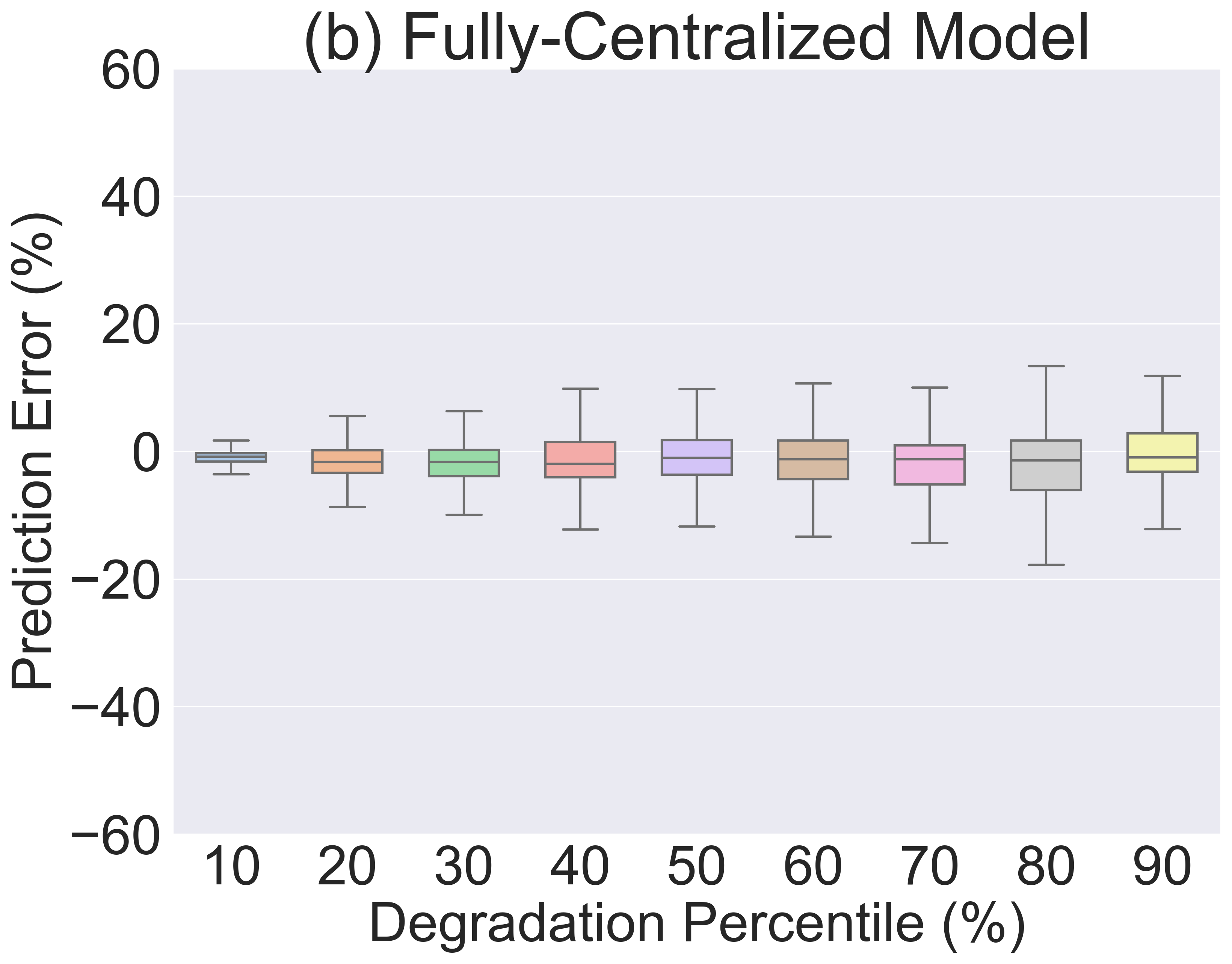}\hfill
\includegraphics[width=.25\textwidth]{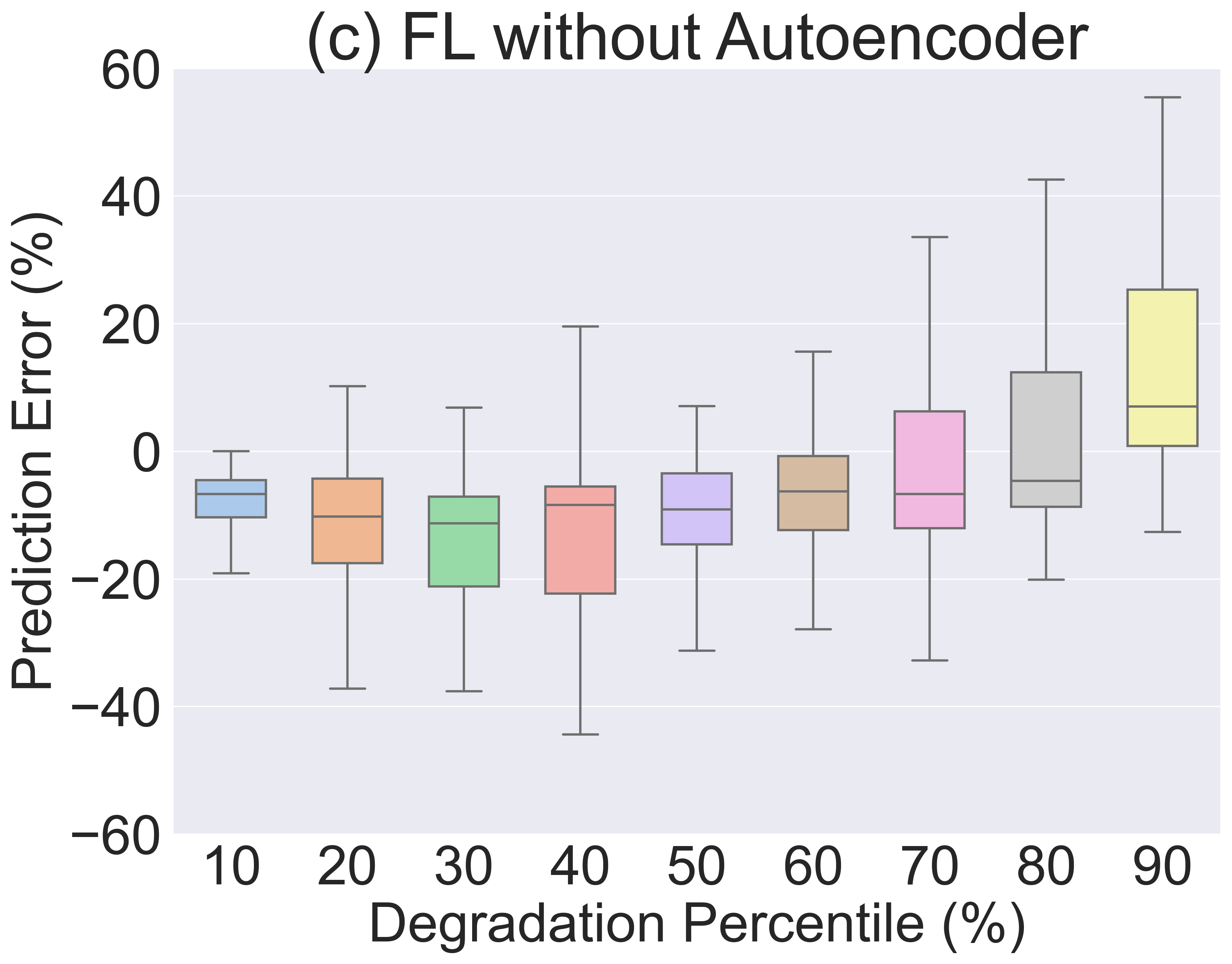}\hfill
\includegraphics[width=.25\textwidth]{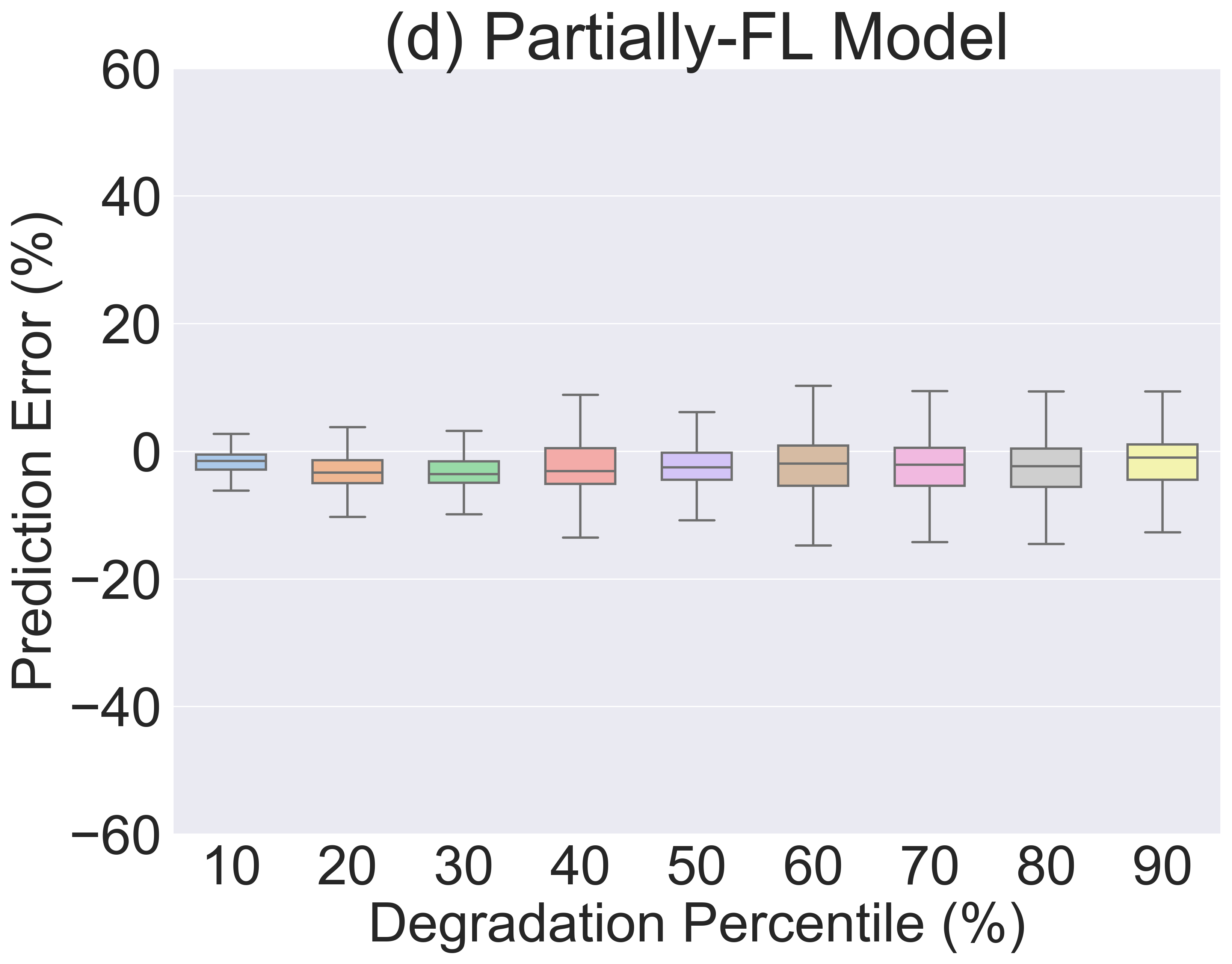}
\caption{Comparison of benchmark models for Argonne Database:NMC532}
\label{fig:bench_2_nmc532}
\end{figure*}

We conducted a comparative analysis of our approach, \textit{HOPE-FED}, against three different benchmarks: \textit{FL w/o autoencoder}, \textit{fully-centralized approach}, and \textit{partially-federated approach}. Detailed results can be found in Tables \ref{nature-main-comp}, \ref{he5050-main-comp}, and \ref{nmc532-main-comp} for the Nature Energy Database, Argonne Database (HE5050 and NMC532 chemistries). A summary of the prediction error plots across different degradation percentiles can be seen in Fig. \ref{fig:bench_2_nature}, \ref{fig:bench_2_he5050}, and \ref{fig:bench_2_nmc532}.

In the first benchmark study, we performed experiments without implementing the autoencoder to assess its effectiveness. This involved directly applying the federated RUL tasks by embedding all battery features. The results showed that leveraging federated autoencoders improved the long-run average cost rate by 24\%, 19\%, and 14\% for the Nature Energy Database, Argonne Database (HE5050 and NMC532 chemistries), respectively, while keeping hyperparameters constant in both experimental settings. Notably, we excluded the federated autoencoder module and maintained the same hyperparameters for the federated RUL task in both experiments.

The second benchmark study focused on a fully centralized model, where both the autoencoder and RUL estimation were performed in a centralized fashion using traditional ML techniques. All battery features were aggregated before applying the autoencoder, and the transformed features were used together to train the RUL estimation DNN. While data residency and privacy were not preserved, the long-run average cost rates improved for the Nature Energy Database, Argonne Database (HE5050 and NMC532 chemistries), respectively. Overall, the long-run average cost rates obtained through the HOPE-FED approach were close to the results of fully-centralized models. HOPE-FED provides a seamless methodology for protecting clients' privacy while achieving similar performance levels to the fully centralized approach. Additionally, the fully centralized approach reduced the average unused life for the Argonne Database.

The third benchmark study examined a partially federated model. The autoencoder module was performed by aggregating all batteries' datasets, similar to centralized approaches. However, it was followed by a federated RUL estimation module. Preprocessing steps such as dimensionality reduction via autoencoders are often overlooked in the literature when implemented in a centralized manner, despite their significance. However, implementing autoencoders centrally already compromises the privacy of the clients. We compared the long-run average cost rates of the partially-federated and HOPE-FED approaches. Our computational experiments demonstrated that HOPE-FED and the partially federated model achieved very similar levels of long-run average cost rates across all databases. For the Argonne Database, the partially federated approach had fewer catastrophic failures compared to HOPE-FED. However, the average unused life was improved in the partially federated approach compared to HOPE-FED for the Nature Energy Database.


In summary, the FL w/o autoencoder approach performed the worst for all the databases, indicating that autoencoders are powerful tools for obtaining high-quality features to improve the prediction performance of ML algorithms. Fully centralized models slightly outperformed fully federated models in terms of cost rates for all three datasets, as expected. A slight increase in the cost rates demonstrates the trade-off for achieving fully preserved privacy, highlighting the strength of FL models. Partially federated models achieved similar levels of long-run average cost rates to fully federated models, indicating that applying dimensionality reduction tasks in a federated manner does not negatively impact performance measures.

\newcommand{\highlight}[2][red]{\textcolor{#1}{#2}}
\begin{table*}[h]
  \centering
  \caption{Comparison of performance measures for benchmark policies and HOPE-FED on Nature Energy Database}
   \resizebox{\textwidth}{!}{
    \begin{tabular}{p{0.3\linewidth}  p{0.125\linewidth} p{0.125\linewidth} p{0.125\linewidth} p{0.125\linewidth}  }
    \hline 
   \multicolumn{1}{l}{} & \multicolumn{1}{c}{\textbf{Fully-Centralized}} & \multicolumn{1}{c}{\textbf{FL w/o Autoencoder}} & \multicolumn{1}{c}{\textbf{Partially-Federated}} & \multicolumn{1}{c}{\textbf{Fully-Federated}} \\
   \hline
    \textbf{Optimal Threshold} & \multicolumn{1}{c} {25}   & \multicolumn{1}{c}{25} & \multicolumn{1}{c}{25} & \multicolumn{1}{c}{25} \\
    \hline
    \textbf{\# Preventive} & \multicolumn{1}{c}{30} & \multicolumn{1}{c}{25} &\multicolumn{1}{c}{29} &\multicolumn{1}{c}{30}   \\
    \hline
    \textbf{\# Corrective} &\multicolumn{1}{c}{1} &\multicolumn{1}{c}{6} &\multicolumn{1}{c}{2} &\multicolumn{1}{c}{1} \\
    \hline
    \textbf{Unused Life} & \multicolumn{1}{c}{22.3} &\multicolumn{1}{c}{71.1} & \multicolumn{1}{c}{20.1} & \multicolumn{1}{c}{20.3} \\
    \hline
    \textbf{Unavailable Days} & \multicolumn{1}{c}{1.03} & \multicolumn{1}{c}{2.0} & \multicolumn{1}{c}{1.3} &\multicolumn{1}{c}{1.2}  \\
    \hline
    \textbf{Cost Rate} & \multicolumn{1}{c}{\textbf{12.3}} & \multicolumn{1}{c}{\textbf{16.5}} & \multicolumn{1}{c}{\textbf{12.5}} & \multicolumn{1}{c}{\textbf{12.6}}   \\
    \hline
    \end{tabular}}
  \label{nature-main-comp}%
\end{table*}%

\begin{table*}[h]
  \centering
  \caption{Comparison of performance measures for benchmark policies and HOPE-FED on Argonne Database: HE5050}
  \resizebox{\textwidth}{!}{
    \begin{tabular}{p{0.3\linewidth}  p{0.125\linewidth} p{0.125\linewidth} p{0.125\linewidth} p{0.125\linewidth}  }
    \hline  
   \multicolumn{1}{l}{} & \multicolumn{1}{c}{\textbf{Fully-Centralized}} & \multicolumn{1}{c}{\textbf{FL w/o Autoencoder}} & \multicolumn{1}{c}{\textbf{Partially-Federated}} & \multicolumn{1}{c}{\textbf{Fully-Federated}} \\
   \hline
    \textbf{Optimal Threshold} & \multicolumn{1}{c} {25}   & \multicolumn{1}{c}{50} & \multicolumn{1}{c}{25} & \multicolumn{1}{c}{25} \\
    \hline
    \textbf{\# Preventive} & \multicolumn{1}{c}{11} & \multicolumn{1}{c}{8} &\multicolumn{1}{c}{12} &\multicolumn{1}{c}{10}   \\
    \hline
    \textbf{\# Corrective} &\multicolumn{1}{c}{8} &\multicolumn{1}{c}{11} &\multicolumn{1}{c}{7} &\multicolumn{1}{c}{9} \\
    \hline
    \textbf{Unused Life} & \multicolumn{1}{c}{96.9} &\multicolumn{1}{c}{288.0} & \multicolumn{1}{c}{123.2} & \multicolumn{1}{c}{114.2} \\
    \hline
    \textbf{Unavailable Days} & \multicolumn{1}{c}{3.1} & \multicolumn{1}{c}{3.9} & \multicolumn{1}{c}{2.8} &\multicolumn{1}{c}{3.4}  \\
    \hline
    \textbf{Cost Rate} & \multicolumn{1}{c}{\textbf{24.5}} & \multicolumn{1}{c}{\textbf{31.9}} & \multicolumn{1}{c}{\textbf{24.8}} & \multicolumn{1}{c}{\textbf{25.7}}   \\
    \hline
    \end{tabular}}
  \label{he5050-main-comp}%
\end{table*}%

\begin{table*}[h]
  \centering
  \caption{Comparison of performance measures for benchmark policies and HOPE-FED on Argonne Database: NMC532}
  \resizebox{\textwidth}{!}{
    \begin{tabular}{p{0.3\linewidth}  p{0.125\linewidth} p{0.125\linewidth} p{0.125\linewidth} p{0.125\linewidth}  }
    \hline  
   \multicolumn{1}{l}{} & \multicolumn{1}{c}{\textbf{Fully-Centralized}} & \multicolumn{1}{c}{\textbf{FL w/o Autoencoder}} & \multicolumn{1}{c}{\textbf{Partially-Federated}} & \multicolumn{1}{c}{\textbf{Fully-Federated}} \\
   \hline
    \textbf{Optimal Threshold} & \multicolumn{1}{c} {25}   & \multicolumn{1}{c}{100} & \multicolumn{1}{c}{50} & \multicolumn{1}{c}{50} \\
    \hline
    \textbf{\# Preventive} & \multicolumn{1}{c}{19} & \multicolumn{1}{c}{14} &\multicolumn{1}{c}{22} &\multicolumn{1}{c}{19}   \\
    \hline
    \textbf{\# Corrective} &\multicolumn{1}{c}{6} &\multicolumn{1}{c}{11} &\multicolumn{1}{c}{3} &\multicolumn{1}{c}{6} \\
    \hline
    \textbf{Unused Life} & \multicolumn{1}{c}{59.8} &\multicolumn{1}{c}{136.4} & \multicolumn{1}{c}{108.8} & \multicolumn{1}{c}{95.8} \\
    \hline
    \textbf{Unavailable Days} & \multicolumn{1}{c}{2.2} & \multicolumn{1}{c}{3.0} & \multicolumn{1}{c}{1.6} &\multicolumn{1}{c}{2.2}  \\
    \hline
    \textbf{Cost Rate} & \multicolumn{1}{c}{\textbf{18.8}} & \multicolumn{1}{c}{\textbf{22.3}} & \multicolumn{1}{c}{\textbf{19.2}} & \multicolumn{1}{c}{\textbf{19.1}}   \\
    \hline
    \end{tabular}}
  \label{nmc532-main-comp}%
\end{table*}%

\subsection{Comparative Analysis: HOPE-FED Approach vs. Batch-Federated Approach}
\label{comp4}
Batch-federated learning is a collaborative training approach that proves beneficial for multi-asset clients aiming to leverage the advantages of FL while addressing specific requirements. In batch-federated learning, a subset of the clients within the multi-asset environment participates in aggregating their data during the model training process. Rather than involving all clients, this approach allows for the consolidation of data from a specific group. By aggregating data from these selected clients, batch-federated learning enables the creation of a unified and representative training dataset for model updates. This method strikes a balance between sharing data for collaborative learning and preserving privacy and operational efficiency specific to multi-asset environments.

Batch-federated learning can be highly advantageous for multi-asset clients, such as wind farms, seeking to leverage the benefits of collaborative model training while ensuring data privacy and operational efficiency. By aggregating data from a subset of turbines within the wind farm, batch-federated learning enables the creation of a more diverse and representative training dataset, leading to improved model accuracy and performance. Additionally, the reduced communication overhead in transmitting aggregated data enhances the overall efficiency of the training process, especially in scenarios where bandwidth is limited. With faster convergence enabled by a larger combined dataset, wind farms can derive valuable insights and optimize their operations, maintenance, and energy production across their entire fleet of turbines.

To implement the batch-federated learning approach, the batteries were randomly divided into 5, 20, and 30 clusters. Within each cluster, the datasets of the associated batteries were aggregated. Increasing the number of clusters brings the approach closer to our proposed fully federated approach, HOPE-FED. Conversely, fewer clusters align the approach more closely with the fully-centralized approach. The long-run average cost rates and other relevant metrics for the Argonne Database, specifically for the NMC532 chemistry, are documented in Table \ref{batch-federated-nmc532}. Additionally, Figure \ref{fig:nmc532-batch} in Appendix \ref{batch} presents a box plot illustrating the prediction error across different degradation percentiles.

\begin{table*}[h]
  \centering
  \caption{Comparison of batch-federated vs. fully-federated approaches for the Argonne Database: NMC532}
  \resizebox{\textwidth}{!}{
    \begin{tabular}{p{0.3\linewidth}  p{0.1\linewidth} p{0.1\linewidth} p{0.1\linewidth} p{0.1\linewidth} p{0.1\linewidth} }
    \hline  
   \multicolumn{1}{l}{} & \multicolumn{1}{c}{\textbf{Fully-Centralized}} & \multicolumn{1}{c}{\textbf{5-Clusters}} & \multicolumn{1}{c}{\textbf{20-Clusters}} & \multicolumn{1}{c}{\textbf{30-Clusters}} & \multicolumn{1}{c}{\textbf{Fully-Federated}}  \\
   \hline
    \textbf{Optimal Threshold} & \multicolumn{1}{c} {25}   & \multicolumn{1}{c}{25} & \multicolumn{1}{c}{50} & \multicolumn{1}{c}{25} & \multicolumn{1}{c}{50}  \\
    \hline
    \textbf{\# Preventive} & \multicolumn{1}{c}{19}    & \multicolumn{1}{c}{19}    & \multicolumn{1}{c}{21}    & \multicolumn{1}{c}{18}    & \multicolumn{1}{c}{19}    \\
    \hline
    \textbf{\# Corrective} & \multicolumn{1}{c}{6}     & \multicolumn{1}{c}{6}   & \multicolumn{1}{c}{4}     & \multicolumn{1}{c}{7}     & \multicolumn{1}{c}{6}     \\
    \hline
    \textbf{Unused Life} & \multicolumn{1}{c}{59.8} & \multicolumn{1}{c}{85.6} & \multicolumn{1}{c}{101.3} & \multicolumn{1}{c}{61.5} &\multicolumn{1}{c}{95.8} \\
    \hline
    \textbf{Unavailable Days} & \multicolumn{1}{c}{2.2} & \multicolumn{1}{c}{2.0} & \multicolumn{1}{c}{1.8} & \multicolumn{1}{c}{2.4} & \multicolumn{1}{c}{2.2} \\
    \hline
    \textbf{Cost Rate} & \multicolumn{1}{c}{\textbf{18.8}} & \multicolumn{1}{c}{\textbf{19.0}} & \multicolumn{1}{c}{\textbf{18.9}} & \multicolumn{1}{c}{\textbf{18.5}} & \multicolumn{1}{c}{\textbf{19.1}}  \\
    \hline
    \end{tabular}}
  \label{batch-federated-nmc532}%
\end{table*}%
The findings presented in Table \ref{batch-federated-nmc532} demonstrate that utilizing 5 and 20 clusters yields long-run average cost rates that fall between those of the fully centralized and fully federated approaches. Conversely, employing 30 clusters slightly enhances the cost-rate performance of both the fully centralized and fully federated approaches. These results affirm the competitive capability of the HOPE-FED approach in terms of performance. Furthermore, the batch-federated approach becomes beneficial in reducing computational load, particularly in scenarios where clients have the flexibility to aggregate datasets from multiple assets they possess.
\section{Conclusion}
\label{6}
In conclusion, this paper proposes a FL approach for predicting the RUL of lithium-ion batteries. The proposed HOPE-FED approach addresses the challenges associated with predicting RUL by implementing dimensionality reduction and prediction tasks in a distributed-federated manner. This approach ensures that each battery's data is kept on its device, thus addressing privacy concerns. The use of a federated autoencoder for dimensionality reduction further enhances the accuracy of the prediction. 
The proposed approach offers notable performance advantages when compared to traditional methods like age-based periodic {replacement}. Furthermore, the HOPE-FED approach demonstrates similar performance levels to benchmark models, including fully centralized, partially federated, and batch-federated approaches, all while ensuring data residency and privacy is maintained. Overall, the proposed approach presents a novel and efficient solution for predicting the RUL of lithium-ion batteries while addressing privacy and cost concerns.

The proposed HOPE-FED approach for predicting the RUL of lithium-ion batteries opens up new avenues for further research. There is potential for exploring the use of other types of neural network models in the FL setting for predicting the RUL of lithium-ion batteries, such as recurrent neural networks (RNNs) or long short-term memory (LSTM) networks. Finally, future research could also investigate the performance of our proposed FL approach on other types of battery systems.

\section{Methods} 
This section presents the methods utilized in our HOPE-FED framework.
\subsection{Autoencoder and RUL Functions}
Our HOPE-FED framework consists of two main stages: federated autoencoder and federated RUL estimation. To achieve these tasks, we have developed the \textit{Autoencoder} and \textit{RUL} functions, which are integral components of our federated battery prognosis framework. The \textit{Autoencoder} function summarizes the steps involved in the dimensionality reduction task and is employed within the Federated Battery Prognosis Algorithm. Each sampled battery executes its own autoencoder function, ensuring the preservation of the federated learning structure. It is important to note that the autoencoder structure and hyperparameters remain consistent across all sampled batteries. Similarly, the \textit{RUL} function is individually invoked for each sampled battery within the federated battery prognosis framework to train the deep neural network for predicting the RUL of lithium-ion batteries.
\begin{algorithm}[H]
\begin{algorithmic}
\State \textbf{function} AUTOENCODER($f_{\theta},f_{\beta}, w_{\theta},w_{\beta},\mathcal{D}$)
\vspace{0.1cm}

\State \quad Set number of epochs to $\mathcal{E}$
\vspace{0.1cm}
\vspace{0.1cm}
\State  \quad  \textbf{for} epoch=1,2,..., $\mathcal{E}$
\vspace{0.1cm}
\State \quad \quad  \textbf{apply} ENCODER:
\vspace{0.1cm}
\State \quad \quad  \ \quad Compress $\mathcal{D}$ to $\mathcal{C}$ by leveraging $f_{\theta}$
\vspace{0.1cm}
\State \quad \quad  \quad \quad $\mathcal{C} \gets f_{\theta}(\mathcal{D};w_{\theta})$
\vspace{0.1cm}
\State \quad \quad  \textbf{apply} DECODER:
\vspace{0.1cm}
\State \quad \quad   \quad  Reconstruct $\mathcal{C}$ by leveraging $f_{\beta}$ to obtain $\hat{\mathcal{D}}$
\vspace{0.1cm}
\State \quad \quad   \quad \quad $
\hat{\mathcal{D}} \gets f_{\beta}(\mathcal{C};w_{\beta})$
\vspace{0.1cm}
\State \quad \quad Compute loss function:
\vspace{0.1cm}
\State \quad   \quad \quad $L(w_{\beta}) =\dfrac{1}{m} \displaystyle \sum^{m}_{i=1} (\mathcal{D}_{i} -f_{\beta}(\mathcal{C}_{i};w_{\beta}))^2$
\vspace{0.1cm}
\State \quad   \quad Update the weights $w_{\theta}$ and $w_{\beta}$
\vspace{0.1cm}
\State  \quad \textbf{end for}
\vspace{0.1cm}
\State \textbf{return} {$w_{\theta},w_{\beta}$}
\end{algorithmic}
\end{algorithm}

\begin{algorithm}[H]
\begin{algorithmic}
\State \textbf{function} 
RUL($f_{\gamma},w_{\gamma},\mathcal{C},\mathcal{P}$)
\vspace{0.1cm}
\State \quad Set number of epochs to $\mathcal{E}$
\vspace{0.1cm}
\vspace{0.1cm}
\State  \quad  \textbf{for} epoch=1,2,..., $\mathcal{E}$
\vspace{0.1cm}
\State \quad \quad Feed input data $C$ to the DNN
\vspace{0.1cm}
\State \quad \quad \quad $
\hat{\mathcal{P}} \gets f_{\gamma}(\mathcal{C};w_{\gamma})$
\vspace{0.1cm}
\State \quad \quad Compute loss function: 
\vspace{0.1cm}
\State \quad   \quad \quad $L(w_{\gamma}) =\dfrac{1}{m} \displaystyle \sum^{m}_{i=1} (\mathcal{P}_{i} -f_{\gamma}(\mathcal{C}_{i};w_{\gamma}))^2$
\vspace{0.1cm}
\State \quad  \quad Update the network weights $w_{\gamma}$ by calling Adam optimizer
\vspace{0.1cm}
\State \quad \textbf{end for}
\vspace{0.1cm}
\State \textbf{return} {$w_{\gamma}$}
\end{algorithmic}
\end{algorithm}

\subsection{Main Algorithm}
Algorithm \ref{alg1} presents federated prognosis steps within HOPE-FED framework which sequentially applies federated dimensionality reduction and federated RUL prediction.
\begin{algorithm}[H]
\caption{Federated Prognosis Algorithm}\label{alg1}
\begin{algorithmic}
\State \textbf{Notation:} 
\State $\mathcal{M}$: set of batteries 
\State $\mathcal{K}$: set of training batteries 
\State $\mathcal{L}$: set of test batteries 
\State $\mathcal{S}$: number of sampled batteries at each round
\State $\mathcal{R}$: sampling ratio of data points at each round
\State $\mathcal{D}_{m}$: input dataset for battery $m$, $m \in \{1,2,..,\mathcal{M}\}$
\State $\mathcal{P}_{m}$: target dataset for battery $m$, $m \in \{1,2,..,\mathcal{M}\}$ which consist of actual values of RUL
\State $\mathcal{T}_{auteoencoder}$: number of rounds for federated autoencoder
\State $\mathcal{T}_{RUL}$: number of rounds for federated RUL prediction
\State $f_{\theta}$: neural network function for encoder
\State $f_{\beta}$: neural network function for decoder
\State $f_{\gamma}$: neural network function for decoder for RUL prediction
\State $w_{\theta}$: network weights of encoder
\State $w_{\beta}$: network weights of decoder
\State $w_{\gamma}$: network weights of RUL prediction
\vspace{0.1cm}
\State \textbf{for} each round $t=1,2,..,\mathcal{T}_{autoencoder}$ \textbf{do}:
\vspace{0.1cm}
\State \quad Randomly select $\mathcal{S}$ batteries from training set $\mathcal{K}$
\vspace{0.1cm}
\State  \quad \textbf{for} each selected battery $s=1,2,..,\mathcal{S}$, in parallel \textbf{do}:
\vspace{0.1cm}
\State \quad  \quad Sample from dataset $\mathcal{D}_{s}$ with $\mathcal{R}$\% and obtain $\mathcal{D}_{s'}$ 
\vspace{0.1cm}
\State \quad \quad \quad $w^{ts}_{\theta},w^{ts}_{\beta} \gets $ \textbf{AUTOENCODER}($f_{\theta}, f_{\beta},w^{ts}_{\theta},w^{ts}_{\beta}, \mathcal{D}_{s'}$)
\vspace{0.1cm}
\State \quad \textbf{end for}
\vspace{0.1cm}
\State \quad Apply federated averaging:
\vspace{0.1cm}
\State \quad \quad \quad $w^{t+1}_{\theta} \gets \dfrac{1}{\| \mathcal{S} \|} \sum^{\mathcal{S}}_{s=1} w^{ts}_{\theta}$
\vspace{0.1cm}
\State \quad \quad \quad $w^{t+1}_{\beta} \gets \dfrac{1}{\| \mathcal{S} \|} \sum^{\mathcal{S}}_{s=1} w^{ts}_{\beta}$
\vspace{0.1cm}
\State \textbf{end for}
\vspace{0.1cm}
\State Freeze encoder and decoder weights, $w_{\theta}$ and $w_{\beta}$ 
\vspace{0.1cm}
\State Transform train inputs using encoder:
\vspace{0.1cm}
\State  \textbf{for} each train battery $k=1,2,..,\mathcal{K}$ in parallel \textbf{do}:
\vspace{0.1cm}
\State \quad $\mathcal{C}_{k} \gets f_{\theta}(\mathcal{D}_{k};w_{\theta})$
\vspace{0.1cm}
\State  \textbf{end for}
\vspace{0.1cm}
\State \textbf{for} each round $t=1,2,..,\mathcal{T}_{RUL}$ in parallel \textbf{do}:
\vspace{0.1cm}
\State \quad Randomly select $\mathcal{S}$ batteries from training set $\mathcal{K}$
\vspace{0.1cm}
\State \quad  Sample from dataset $\mathcal{C}_{s}$ with $\mathcal{R}$\% and obtain $\mathcal{C}_{s'}$ 
\vspace{0.1cm}
\State  \quad \textbf{for} each selected battery $s=1,2,..,\mathcal{S}$, in parallel \textbf{do}:
\vspace{0.1cm}
\State \quad \quad \quad $w^{ts}_{\gamma} \gets$ \textbf{RUL}($f_{\gamma},w^{ts}_{\gamma},\mathcal{C}_{s'},\mathcal{P}_{s'}$)
\vspace{0.1cm}
\State \quad \textbf{end for}
\vspace{0.1cm}
\State \quad Apply federated averaging:
\vspace{0.1cm}
\State \quad \quad \quad $w^{t+1}_{\gamma} \gets \dfrac{1}{\| \mathcal{S} \|} \sum^{\mathcal{S}}_{s=1} w^{ts}_{\gamma}$
\vspace{0.1cm}
\State \textbf{end for}
\vspace{0.1cm}
\State Freeze RUL prediction weights, $w_{\gamma}$ 
\vspace{0.1cm}
\State \textbf{for} each test battery $l=1,2,..,\mathcal{L}$ 
\textbf{do}:
\vspace{0.1cm}
\State \quad Transform inputs $\mathcal{D}_{l}$ using encoder:
\vspace{0.1cm}
\State \quad \quad $\mathcal{C}_{l} \gets f_{\theta}(\mathcal{D}_{l};w_{\theta})$
\vspace{0.1cm}
\State \quad Make RUL predictions:
\vspace{0.1cm}
\State \quad \quad $\hat{\mathcal{P}_{l}} \gets f_{\gamma}(\mathcal{C}_{l},w_{\gamma})$
\vspace{0.1cm}
\State \quad Report performance measures for the test set $D_{l}$
\vspace{0.1cm}
\State \textbf{end for}
\end{algorithmic}
\end{algorithm}

\subsection{Long-run Average Cost Rate Calculation Framework}
Algorithm \ref{alg2} provides a comprehensive overview of the long-run average cost rate calculation for each battery, enabling the quantification of prediction efficacy to facilitate comparisons between benchmark studies and our HOPE-FED approach.
\begin{algorithm}[H]
\caption{Long-run Average Cost Rate Calculation Algorithm}\label{alg2}
\begin{algorithmic}
\State \textbf{Notation:} 
\State $\boldsymbol{t_{fi}}$: failure time of battery $i$, $i \in {1,2,..,\mathcal{M}}$ 
\State  $\boldsymbol{t_{c}}$: time for crew to initiate a battery replacement operation after {replacement} is 
\State \quad  \mbox{ } triggered
\State  $\boldsymbol{\delta}$:\mbox{ } threshold remaining useful lifetime level for triggering a replacement operation 
\State \quad \mbox{ } before a failure happens
\State $\boldsymbol{t^{*}_{i}}$: age of battery $i$, $i \in {1,2,..,\mathcal{M}}$ when remaining useful lifetime prediction hits 
\State \quad  \mbox{ } under $\delta$ for the first time
\State  $\boldsymbol{c_r}$: the cost of replacing a battery before it experiences catastrophic failure
\State  $\boldsymbol{c_f}$: the cost of replacing a battery after it experiences catastrophic failure

\State  $\boldsymbol{C_{i}}$: long-run average cost rate per battery $i$

\\
\State  \textbf{for} each battery $i=1,2,...,\mathcal{M}$ \textbf{do}:
\vspace{0.1cm}
\State \quad  Calculate long-run average cost rate of prediction $C_{i}$:
\vspace{0.1cm}
\State \quad \quad  \textbf{if} $t^{*}_{i}+t_{c}<t_{fi}$:
\vspace{0.1cm}
\State \quad \quad \quad  $C_{i}=\dfrac{c_r}{t^*_{i}+t_{c}}$
\vspace{0.1cm}
\State \quad \quad  \textbf{else}:
\vspace{0.1cm}
\State \quad \quad \quad  $C_{i}=\dfrac{c_f}{t_{fi}}$
\vspace{0.1cm}
\State \textbf{end for}
\end{algorithmic}
\end{algorithm}

\subsection{Long-run Average Cost Rate Calculation for Age-based Periodic Replacement}
Algorithm \ref{alg3} introduces the long-run average cost rate calculation for age-based periodic {replacement}, which plays a crucial role in determining the optimal time period to initiate {replacement} activities. By minimizing the long-run average cost rate among the available candidate time periods, this algorithm aids in the selection of the most effective {replacement} triggering point.
\begin{algorithm}[H]
\caption{Benchmark Long-run Average Cost Rate Calculation}\label{alg3}
\begin{algorithmic}
\State  \textbf{for} each time period  $t^*=1,2,...,\mathcal{T}$ \textbf{do}:
\State  \quad \textbf{for} each battery $i^*=1,2,...,\mathcal{K}$ in the training set \textbf{do}:
\State \quad  \quad Calculate long-run average cost rate of prediction $C_{it^{*}}$:
\State \quad \quad  \quad  \textbf{if} $t^{*}+t_{c}<t_{fi}$:
\vspace{0.1cm}
\State \quad \quad \quad \quad $C_{it^{*}}=\dfrac{c_r}{t^*+t_{c}}$
\State \quad \quad \quad  \textbf{else}:
\State \quad \quad  \quad \quad $C_{it^{*}}=\dfrac{c_f}{t_{fi}}$
\State \quad \textbf{end for}
\State \textbf{end for}
\State Select $t^*$ value that minimizes $\displaystyle \sum_{i} C_{it^{*}}$
\end{algorithmic}
\end{algorithm}

\backmatter

\pagebreak
\begin{appendices}

\section{Data Processing}
\label{ap}
To implement our HOPE-FED framework, we employed two datasets, namely the Nature Energy and Argonne databases, which are detailed in Sections \ref{dataa} and \ref{datab}, respectively. The feature generation procedure utilized for these datasets is explained in Section \ref{fg}.
\subsection{Nature Energy Database}
\label{dataa}
\sloppy
In our computational experiments, we employed one of the most extensive publicly available datasets for lithium-ion batteries, which was introduced by Severson et al. \footnote[1]{The dataset is retrieved from \burl{https://data.matr.io/1/projects/5c48dd2bc625d700019f3204}}.
This dataset comprises 124 commercial lithium-ion phosphate (LFP) / graphite cells with a nominal capacity of 1.1 Ah and a nominal voltage of 3.3 V. The cycle lives of these cells range from 150 to 2300 cycles, with cycle life defined as the number of cycles completed corresponding to 80\% of the nominal capacity. The dataset generated by Severson et al. contains approximately 96,700 cycle data points from 124 commercial lithium-ion batteries. The average cycle life is 806 cycles with a standard deviation of 377 cycles. Among the batteries, 48 have a cycle life of less than 550 cycles, while 90 batteries fall within the range of 550 to 1200 cycles.

During the data collection process, the batteries were subjected to 72 distinct fast-charging conditions while maintaining identical discharging conditions (4.0 C / 2.0 V). The fast-charging rates varied between 3.6 C and 6.0 C for a duration of 10 minutes, and the batteries were charged until reaching 80\% state-of-charge (SOC) conditions using one or two different fast-charging rates. After the completion of fast-charging, the batteries were further charged until reaching 100\% SOC using a 1C CC-CV charge, ramping up to 3.6 V with a C/50 charge cutoff. Throughout the data generation process, the voltage, current, internal resistance, and cell temperature were recorded. It is important to note that although the cell temperature was initially set to $30^\circ C$, charging and discharging operations could cause the cell temperature to fluctuate by up to $10^\circ C$. For more comprehensive details regarding the data collection process, we refer the reader to Severson et al.'s work \cite{severson2019data}.

\subsection{Argonne Database}
\label{datab}
\sloppy
Our second battery dataset is obtained from the Argonne Cell Analysis, Modeling, and Prototyping (CAMP) facility \cite{paulson2022feature} \footnote[2]{Processing steps of the dataset can be found in \cite{git} and dataset can be downloaded from \burl{https://acdc.alcf.anl.gov/mdf/detail/camp_2023_v3.5/}.}. This dataset comprises 300 batteries with six distinct metal oxide cathode chemistries, namely NMC111, NMC532, NMC622, NMC811, HE5050, and 5Vspinel. The selection of batteries for this dataset was based on specific criteria: they utilize graphite as the active material, have charging rates equal to or less than 1C, and have undergone performance testing for a minimum of 100 cycles.  It is worth noting that batteries belonging to different chemistries exhibit varying cycle life values, and even within the same chemistry, properties such as porosity, loadings, and materials can vary.  In our analysis, we focus on the NMC532 and HE5050 chemistries, which are the largest chemistries within the dataset, and present our experimental results for these two chemistries. For further in-depth information on the CAMP dataset, please refer to \cite{paulson2022feature}.
\subsection{Feature Generation}
\label{fg}
The Nature Energy Database, as presented in \cite{severson2019data}, encompasses a wide range of features that are categorized into summary data and cycle data. The summary data provides per-cycle information regarding charge and discharge capacity, internal resistance, charging time, cycle number, and temperature statistics. On the other hand, the cycle data captures detailed information within each cycle, including the data stream of various features such as charge and discharge capacity, temperature, voltage, and current.

By leveraging these diverse time-series cycle features, it becomes possible to generate new features that contribute to a better understanding of the underlying causes of degradation in lithium-ion batteries. These additional features ultimately enhance our prediction task by providing valuable insights into the degradation behavior. Through the utilization of different time-series cycle features, we can create new features specifically designed to improve our understanding of the underlying reasons behind the degradation behavior exhibited by lithium-ion batteries.

Capacity is a pivotal health indicator for lithium-ion batteries \cite{diao2019algorithm}, providing a time frame for the operation of a fully charged battery under current environmental conditions. Capacity fade curves are frequently generated to observe the degradation behavior of batteries \cite{saxena2022convolutional}, \cite{honkura2011capacity}. The capacity itself, along with its temporal changes, serves as a powerful feature for estimating the remaining useful life of batteries. Furthermore, capacity values can be evaluated in conjunction with voltage values, and analyzing their time-series behavior within cycles can offer insights into degradation mechanisms. Notably, Severson et al. \cite{severson2019data} predict the lifetime classification of different cells by leveraging features derived from observed changes in the capacity curve. In a more intricate approach, they incorporate additional features such as temperature, internal resistance, and charge time to enhance their prediction task.


In our study, we have carefully engineered a set of 68 diverse features to predict the remaining useful life (RUL) of lithium-ion batteries. These features encompass information extracted from both summary and cycle data sources. The summary features provide valuable insights into the internal resistance, charge and discharge capacities, average, minimum, and maximum temperature, as well as charging times observed for each cycle. Additionally, features representing the charging policy have been incorporated into the feature set. To further enrich the feature set, we have calculated statistical measures such as mean, variance, skewness, and kurtosis by comparing the cycle data from previous cycles with the current cycle. This comprehensive set of features enables us to capture various aspects of battery behavior and maximize the predictive capabilities for RUL estimation.

In the case of the Argonne Database \cite{paulson2022feature}, a similar feature generation procedure is applied, resulting in the creation of a total of 74 features that are embedded to the dimensionality reduction task. The dataset includes discharge energy and capacity, as well as charge energy and capacity values for each cycle. Additionally, raw feature values are available, which can be leveraged to generate additional features. By utilizing the voltage and current values from the raw data, we computed statistical measures such as mean and standard deviation, resulting in the creation of supplementary features. Furthermore, temporal features and statistical measures were derived from the discharge and charge capacities and energies within the database. This comprehensive feature set captures various aspects of the battery behavior and enables effective dimensionality reduction techniques to be employed for further analysis and modeling.


\section{Algorithms in Online Supplement}
\label{A2}
Algorithm \ref{alg4} outlines the step-by-step process for computing the average unused life across all batteries in the test data.

\begin{algorithm}[H]
\caption{Calculation of average unused life}\label{alg4}
\begin{algorithmic}
\State \textbf{Notation:} 
\State $\boldsymbol{e_{i}}$: number of unavailable days for lithium-ion battery $i$
\State $\boldsymbol{t_m}$: duration of {replacement} activities in terms of days
\State $\mathcal{R}$: set of replaced batteries
\State \textbf{for} each battery $i={1,2,...,\mathcal{L}}$
\State \quad \quad  \textbf{if} $t^{*}+t_{c}<t_{fi}$:
\State \quad \quad \quad Extend battery $i$ to set $R$
\State \quad \quad \quad $e_{i}=t_{fi}-(t^{*}+t_{c}+t_{m})$
\State Calculate average unused periods:
\State \quad \quad  $\bar{e} = \dfrac {\displaystyle \sum_{{i}\in \mathcal{R}}  e_{i}}{|\mathcal{R}|}$
\end{algorithmic}
\end{algorithm}

Algorithm \ref{alg5} presents the sequential steps involved in calculating the average number of days that batteries are unavailable due to {replacement} operations across the entire test data set.

\begin{algorithm}[H]
\caption{Calculation of average number of unavailable days}\label{alg5}
\begin{algorithmic}
\State \textbf{Notation:} 
\State $\boldsymbol{u_{i}}$: number of unavailable periods for lithium-ion battery $i$
\State $\boldsymbol{t_m}$: duration of {replacement} activities in terms of days
\State \textbf{for} each battery $i={1,2,...,\mathcal{L}}$ \textbf{do}:
\State \quad   \textbf{if} $t^{*}+t_{c}<t_{fi}$:
\State \quad \quad $u_{i}$=$t_m$
\State \quad   \textbf{else if}: $t^{*}<t_{fi}$: 
\State \quad   \quad $u_{i}=(t_{c}+t_m)-(t_{fi}-t^{*})$
\State \quad  \textbf{else}:
\State \quad   \quad $u_{i}=t_{c}+t_{m}$
\State \textbf{end for}
\State Calculate average unavailable periods:
\State \quad   $\bar{u} = \dfrac {\displaystyle \sum_{i} u_{i}}{|\mathcal{L}|}$
\end{algorithmic}
\end{algorithm}

\section{Age-based Periodic {Replacement} Policy Benchmark Results}
Tables \ref{periodic-nature}, \ref{periodic-he5050}, and \ref{periodic-nmc532} provide a detailed comparison of the age-based periodic {replacement} policy and the HOPE-FED framework under different threshold values for triggering {replacement} activities, specifically for the Nature Energy Database and the Argonne Database with HE5050 and NMC532 chemistries, respectively.
\label{periodic_additional}
\begin{table*}[h]
  \centering
  \caption{Comparison of age-based periodic replacement vs. fully-federated prediction-based replacement approaches on Nature Energy Database}
  \resizebox{\textwidth}{!}{
    \begin{tabular}{p{0.3\linewidth}  p{0.1\linewidth} p{0.1\linewidth} p{0.1\linewidth} p{0.1\linewidth} p{0.1\linewidth}}
    \hline
    \multicolumn{1}{l}{} & \multicolumn{1}{c}{\textbf{Benchmark Policy}} & \multicolumn{1}{c}{\textbf{Threshold=10}} & \multicolumn{1}{c}{\textbf{Threshold=25}} & \multicolumn{1}{c}{\textbf{Threshold=50}} & \multicolumn{1}{c}{\textbf{Threshold=100}}\\
    \hline
    \textbf{Trigger Time} & 451   & \multicolumn{1}{c}{Prediction-based} & \multicolumn{1}{c}{Prediction-based} & \multicolumn{1}{c}{Prediction-based} & \multicolumn{1}{c}{Prediction-based} \\
    \hline
    \textbf{\# Preventive} & 29    & 25    & 30    & 31 & 31\\
    \hline
   \textbf{\# Corrective} & 2     & 6    & 1     & 0 & 0\\
    \hline
    \textbf{Unused Life} & 444.5     &5.4    & 20.3    & 46 & 114.3\\
    \hline
    \textbf{Unavailable Days} & 1.3     &1.6    & 1.2     & 1 & 1\\
    \hline
    \textbf{Cost Rate} & \textbf{20.3} & \textbf{13.4} & \textbf{12.6} & \textbf{12.9} & \textbf{16.1} \\
    \hline
    \end{tabular}}
  \label{periodic-nature}%
\end{table*}%

\begin{table*}[h]
  \centering
  \caption{Comparison of age-based periodic replacement vs. fully-federated prediction-based replacement approaches on Argonne Database: HE5050}
  \resizebox{\textwidth}{!}{
    \begin{tabular}{p{0.3\linewidth}  p{0.1\linewidth} p{0.1\linewidth} p{0.1\linewidth} p{0.1\linewidth} p{0.1\linewidth} }
    \hline  
   \multicolumn{1}{l}{} & \multicolumn{1}{c}{\textbf{Benchmark Policy}} & \multicolumn{1}{c}{\textbf{Threshold=10}} & \multicolumn{1}{c}{\textbf{Threshold=25}} & \multicolumn{1}{c}{\textbf{Threshold=50}} & \multicolumn{1}{c}{\textbf{Threshold=100}}  \\
    \hline
    \textbf{Trigger Time} & 1013  & \multicolumn{1}{c}{Prediction-based} & \multicolumn{1}{c}{Prediction-based} & \multicolumn{1}{c}{Prediction-based} & \multicolumn{1}{c}{Prediction-based}  \\
    \hline
    \textbf{\# Preventive} & 10    & 7     & 10    & 10    & 13    \\
    \hline
    \textbf{\# Corrective} & 9     & 12    & 9     & 9     & 6      \\
    \hline
    \textbf{Unused Life} & 262.9 & 148.9 & 114.2 & 143.8 & 156 \\
    \hline
    \textbf{Unavailable Days} & 3.4     & 3.9    & 3.4     & 3.4 & 2.6\\
    \hline
    \textbf{Cost Rate} & \textbf{32.5} & \textbf{30.4} & \textbf{25.7} & \textbf{27.5} & \textbf{28.8}  \\
    \hline
    \end{tabular}}
  \label{periodic-he5050}%
\end{table*}%

\begin{table*}[h]
  \centering
  \caption{Comparison of age-based periodic replacement vs. fully-federated prediction-based replacement approaches on Argonne Database: NMC532}
  \resizebox{\textwidth}{!}{
    \begin{tabular}{p{0.3\linewidth}  p{0.1\linewidth} p{0.1\linewidth} p{0.1\linewidth} p{0.1\linewidth} p{0.1\linewidth} }
    \hline  
   \multicolumn{1}{l}{} & \multicolumn{1}{c}{\textbf{Benchmark Policy}} & \multicolumn{1}{c}{\textbf{Threshold=10}} & \multicolumn{1}{c}{\textbf{Threshold=25}} & \multicolumn{1}{c}{\textbf{Threshold=50}} & \multicolumn{1}{c}{\textbf{Threshold=100}}  \\
   \hline
    \textbf{Trigger Time} & 593   & \multicolumn{1}{c}{Prediction-based} & \multicolumn{1}{c}{Prediction-based} & \multicolumn{1}{c}{Prediction-based} & \multicolumn{1}{c}{Prediction-based} \\
    \hline
    \textbf{\# Preventive} & 18    & 13    & 16    & 17    & 21     \\
    \hline
    \textbf{\# Corrective} & 7     & 12    & 9     & 8     & 4     \\
    \hline
    \textbf{Unused Life} & 551.8 & 78.2 & 73.69 & 95.8 &124.6 \\
    \hline
    \textbf{Unavailable Days} & 2.4 & 3.9 & 2.5 & 2.2 & 1.6 \\
    \hline
    \textbf{Cost Rate} & \textbf{26.5} & \textbf{22.7} & \textbf{21.2} & \textbf{20.6} & \textbf{21.4}  \\
    \hline
    \end{tabular}}
  \label{periodic-nmc532}%
\end{table*}%

\section{Batch-federated Results for Argonne Database}
\label{batch}
\begin{figure}[H]
\centering
\includegraphics[width=.2\textwidth]{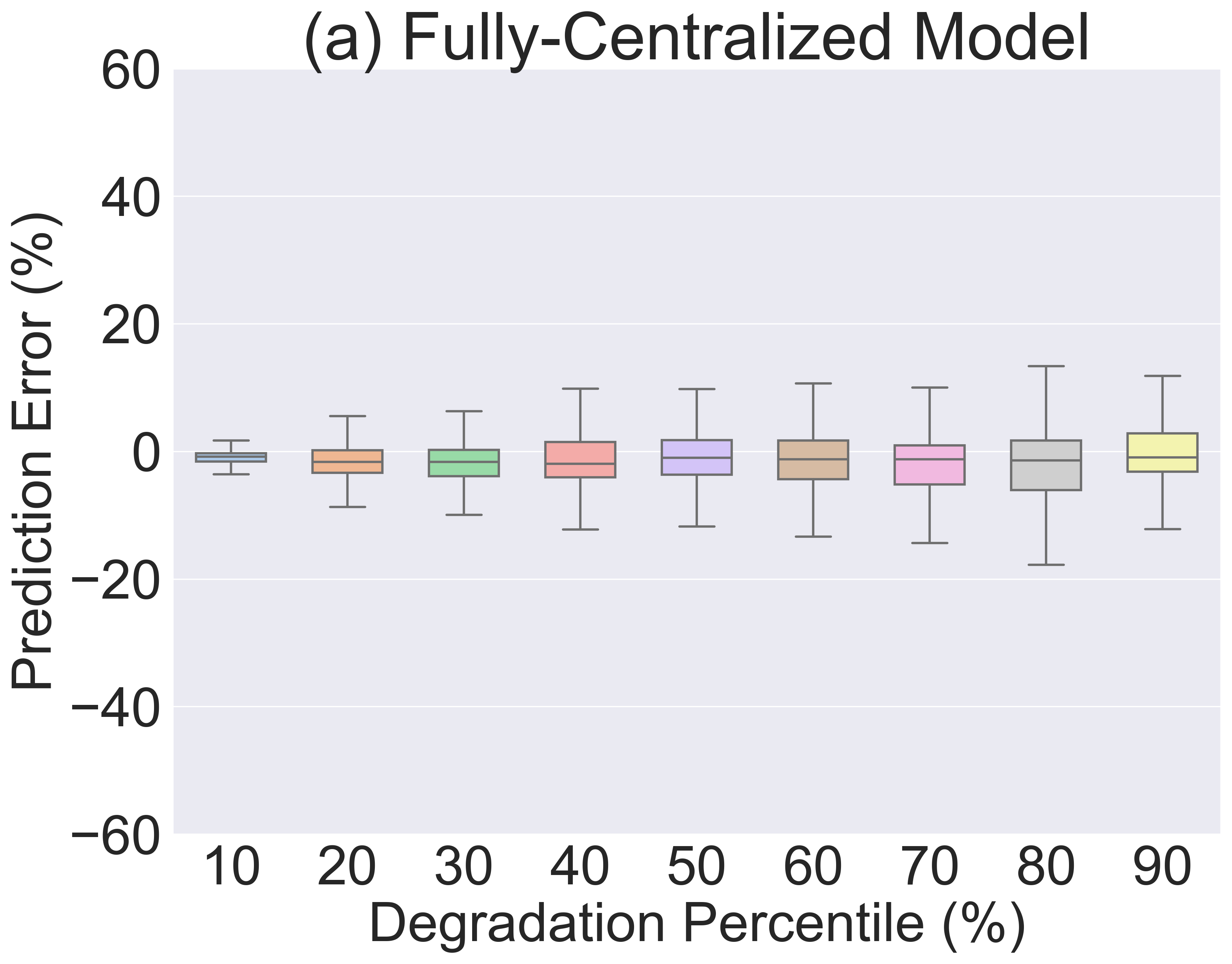}\hfill
\includegraphics[width=.2\textwidth]{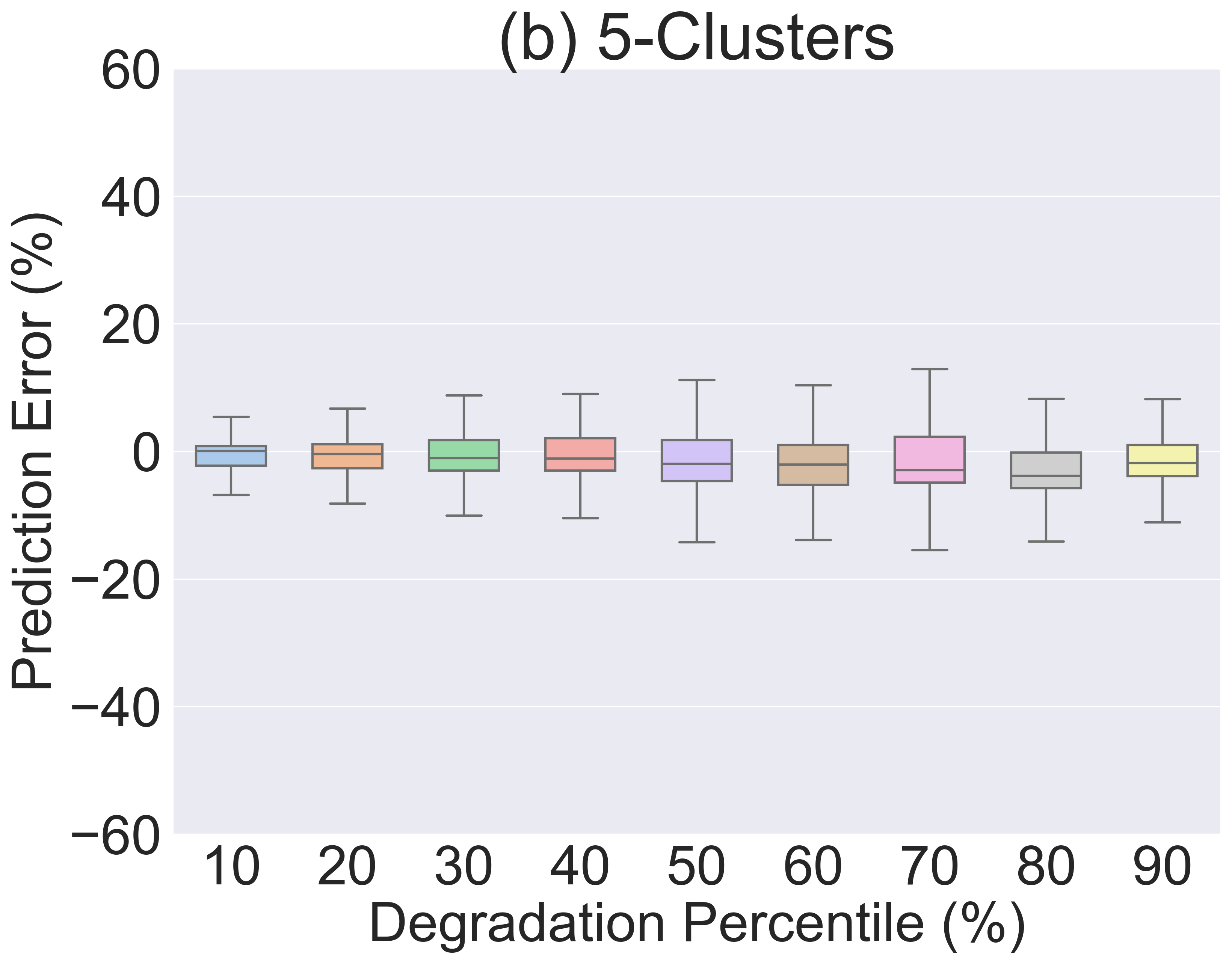}\hfill
\includegraphics[width=.2\textwidth]{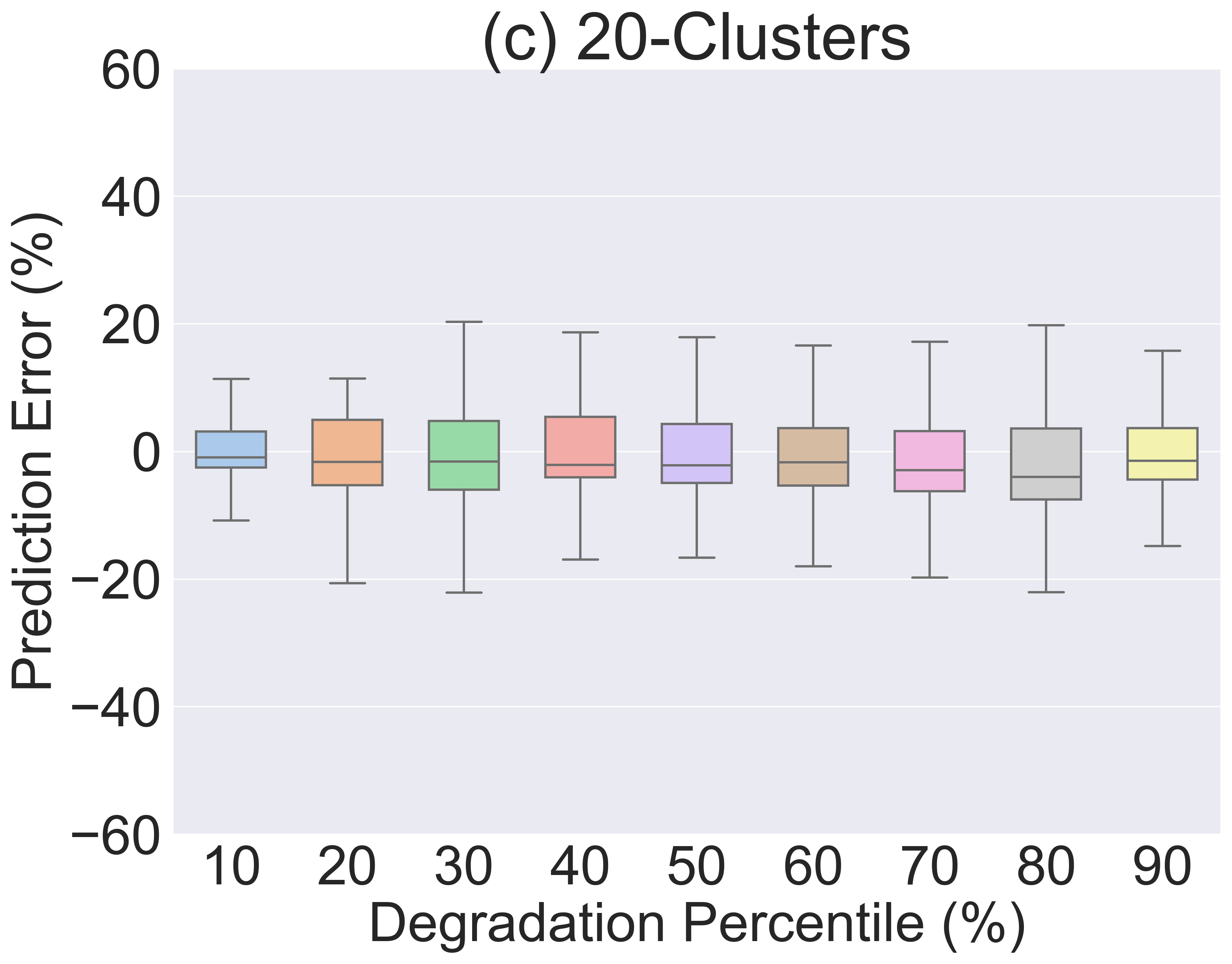}\hfill
\includegraphics[width=.2\textwidth]
{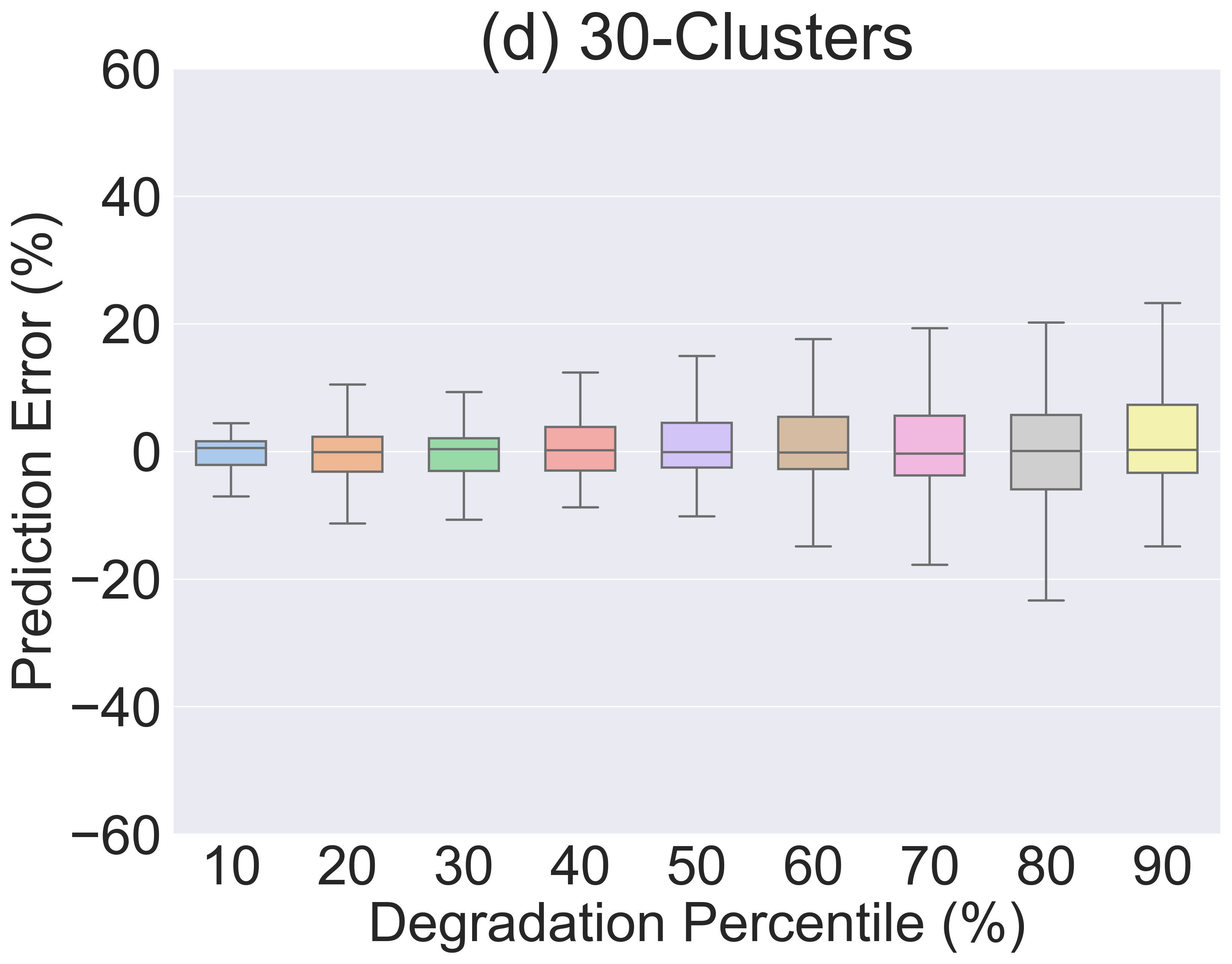}\hfill
\includegraphics[width=.2\textwidth]{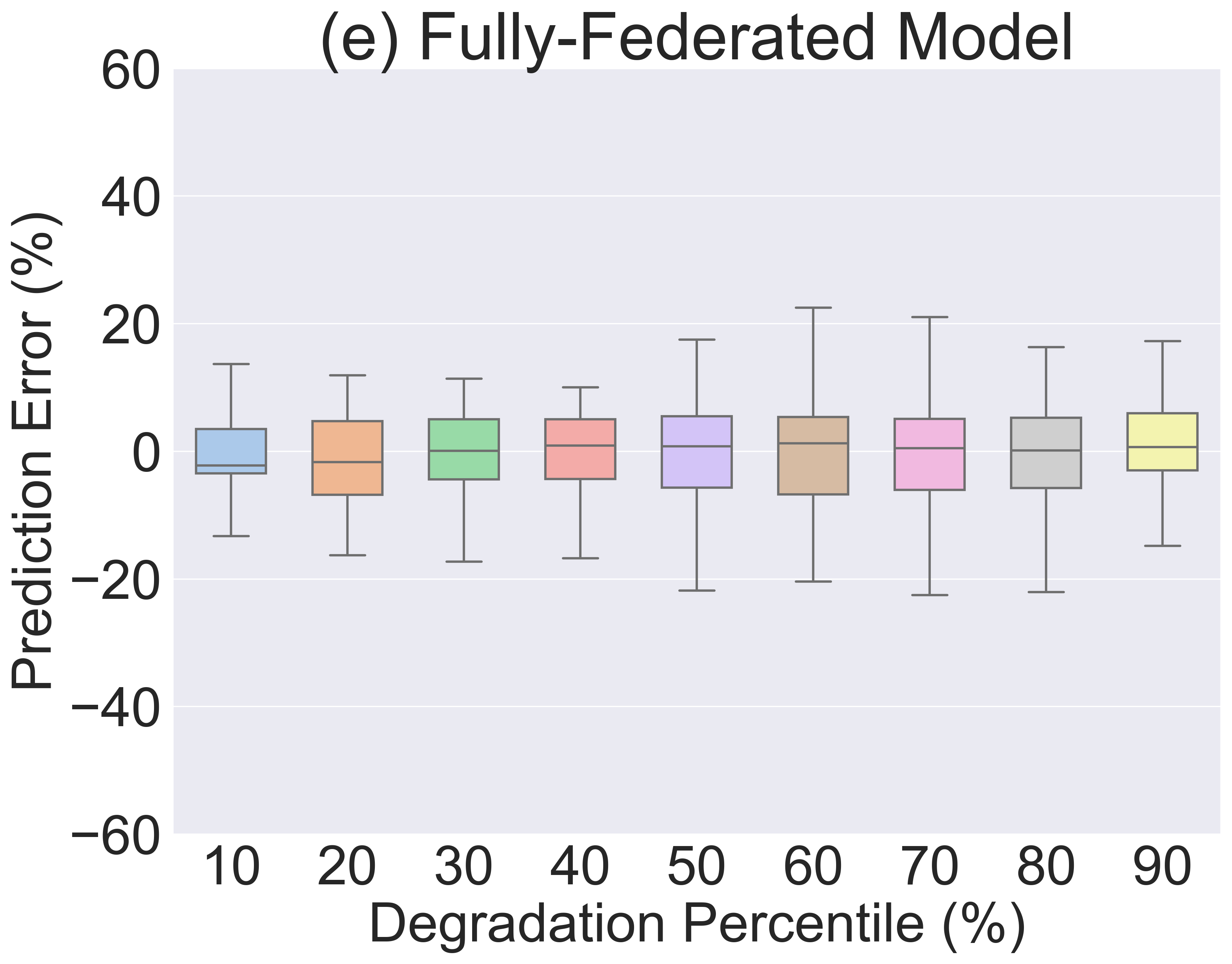}
\caption{Batch-federated computational results for Argonne Database: NMC532}
\label{fig:nmc532-batch}
\end{figure}

Fig. \ref{fig:nmc532-batch} displays the box plots depicting the results of batch-federated experiments conducted on the NMC532 chemistry from the Argonne database. The x-axis represents degradation percentiles, while the y-axis represents prediction errors. Notably, the fully-centralized approach exhibits lower levels of error compared to other approaches. As the number of clusters increases, indicating a decrease in data aggregation, the model approaches the fully federated approach. Consequently, the prediction error generally increases as the model moves closer to the fully federated approach, aligning with expectations.

\end{appendices}

\label{bnm}

\pagebreak
\bibliography{main}


\begin{thebibliography}{54}
\ifx \bisbn   \undefined \def \bisbn  #1{ISBN #1}\fi
\ifx \binits  \undefined \def \binits#1{#1}\fi
\ifx \bauthor  \undefined \def \bauthor#1{#1}\fi
\ifx \batitle  \undefined \def \batitle#1{#1}\fi
\ifx \bjtitle  \undefined \def \bjtitle#1{#1}\fi
\ifx \bvolume  \undefined \def \bvolume#1{\textbf{#1}}\fi
\ifx \byear  \undefined \def \byear#1{#1}\fi
\ifx \bissue  \undefined \def \bissue#1{#1}\fi
\ifx \bfpage  \undefined \def \bfpage#1{#1}\fi
\ifx \blpage  \undefined \def \blpage #1{#1}\fi
\ifx \burl  \undefined \def \burl#1{\textsf{#1}}\fi
\ifx \doiurl  \undefined \def \doiurl#1{\url{https://doi.org/#1}}\fi
\ifx \betal  \undefined \def \betal{\textit{et al.}}\fi
\ifx \binstitute  \undefined \def \binstitute#1{#1}\fi
\ifx \binstitutionaled  \undefined \def \binstitutionaled#1{#1}\fi
\ifx \bctitle  \undefined \def \bctitle#1{#1}\fi
\ifx \beditor  \undefined \def \beditor#1{#1}\fi
\ifx \bpublisher  \undefined \def \bpublisher#1{#1}\fi
\ifx \bbtitle  \undefined \def \bbtitle#1{#1}\fi
\ifx \bedition  \undefined \def \bedition#1{#1}\fi
\ifx \bseriesno  \undefined \def \bseriesno#1{#1}\fi
\ifx \blocation  \undefined \def \blocation#1{#1}\fi
\ifx \bsertitle  \undefined \def \bsertitle#1{#1}\fi
\ifx \bsnm \undefined \def \bsnm#1{#1}\fi
\ifx \bsuffix \undefined \def \bsuffix#1{#1}\fi
\ifx \bparticle \undefined \def \bparticle#1{#1}\fi
\ifx \barticle \undefined \def \barticle#1{#1}\fi
\bibcommenthead
\ifx \bconfdate \undefined \def \bconfdate #1{#1}\fi
\ifx \botherref \undefined \def \botherref #1{#1}\fi
\ifx \url \undefined \def \url#1{\textsf{#1}}\fi
\ifx \bchapter \undefined \def \bchapter#1{#1}\fi
\ifx \bbook \undefined \def \bbook#1{#1}\fi
\ifx \bcomment \undefined \def \bcomment#1{#1}\fi
\ifx \oauthor \undefined \def \oauthor#1{#1}\fi
\ifx \citeauthoryear \undefined \def \citeauthoryear#1{#1}\fi
\ifx \endbibitem  \undefined \def \endbibitem {}\fi
\ifx \bconflocation  \undefined \def \bconflocation#1{#1}\fi
\ifx \arxivurl  \undefined \def \arxivurl#1{\textsf{#1}}\fi
\csname PreBibitemsHook\endcsname

\bibitem[\protect\citeauthoryear{{United Nations Department of Economic and Social Affairs}}{2023}]{UN2023}
\begin{botherref}
\oauthor{\bsnm{{United Nations Department of Economic and Social Affairs}}}:
Frontier Technology Issues: Lithium-ion Batteries - A Pillar for a Fossil Fuel-Free Economy
(2023).
\url{https://www.un.org/development/desa/dpad/publication/frontier-technology-issues-lithium-ion-batteries-a-pillar-for-a-fossil-fuel-free-economy/}
\end{botherref}
\endbibitem

\bibitem[\protect\citeauthoryear{Ward et~al.}{2022}]{ward2022principles}
\begin{barticle}
\bauthor{\bsnm{Ward}, \binits{L.}},
\bauthor{\bsnm{Babinec}, \binits{S.}},
\bauthor{\bsnm{Dufek}, \binits{E.J.}},
\bauthor{\bsnm{Howey}, \binits{D.A.}},
\bauthor{\bsnm{Viswanathan}, \binits{V.}},
\bauthor{\bsnm{Aykol}, \binits{M.}},
\bauthor{\bsnm{Beck}, \binits{D.A.}},
\bauthor{\bsnm{Blaiszik}, \binits{B.}},
\bauthor{\bsnm{Chen}, \binits{B.-R.}},
\bauthor{\bsnm{Crabtree}, \binits{G.}}, \betal:
\batitle{Principles of the battery data genome}.
\bjtitle{Joule}
\bvolume{6}(\bissue{10}),
\bfpage{2253}--\blpage{2271}
(\byear{2022})
\end{barticle}
\endbibitem

\bibitem[\protect\citeauthoryear{Severson et~al.}{2019}]{severson2019data}
\begin{barticle}
\bauthor{\bsnm{Severson}, \binits{K.A.}},
\bauthor{\bsnm{Attia}, \binits{P.M.}},
\bauthor{\bsnm{Jin}, \binits{N.}},
\bauthor{\bsnm{Perkins}, \binits{N.}},
\bauthor{\bsnm{Jiang}, \binits{B.}},
\bauthor{\bsnm{Yang}, \binits{Z.}},
\bauthor{\bsnm{Chen}, \binits{M.H.}},
\bauthor{\bsnm{Aykol}, \binits{M.}},
\bauthor{\bsnm{Herring}, \binits{P.K.}},
\bauthor{\bsnm{Fraggedakis}, \binits{D.}}, \betal:
\batitle{Data-driven prediction of battery cycle life before capacity degradation}.
\bjtitle{Nature Energy}
\bvolume{4}(\bissue{5}),
\bfpage{383}--\blpage{391}
(\byear{2019})
\end{barticle}
\endbibitem

\bibitem[\protect\citeauthoryear{{U.S. Department of Energy}}{2019}]{EnergyGov2018}
\begin{botherref}
\oauthor{\bsnm{{U.S. Department of Energy}}}:
Spotlight: Solving Energy Challenges in Energy Storage
(2019).
\url{https://www.energy.gov/sites/default/files/2019/07/f64/2018-OTT-Energy-Storage-Spotlight.pdf}
\end{botherref}
\endbibitem

\bibitem[\protect\citeauthoryear{OECD}{2019}]{oecdenhancing}
\begin{bbook}
\bauthor{\bsnm{OECD}}:
\bbtitle{Enhancing Access to and Sharing of Data},
p. \bfpage{135}
(\byear{2019}).
\doiurl{10.1787/276aaca8-en} .
\burl{https://www.oecd-ilibrary.org/content/publication/276aaca8-en}
\end{bbook}
\endbibitem

\bibitem[\protect\citeauthoryear{Kaissis et~al.}{2020}]{kaissis2020secure}
\begin{barticle}
\bauthor{\bsnm{Kaissis}, \binits{G.A.}},
\bauthor{\bsnm{Makowski}, \binits{M.R.}},
\bauthor{\bsnm{R{\"u}ckert}, \binits{D.}},
\bauthor{\bsnm{Braren}, \binits{R.F.}}:
\batitle{Secure, privacy-preserving and federated machine learning in medical imaging}.
\bjtitle{Nature Machine Intelligence}
\bvolume{2}(\bissue{6}),
\bfpage{305}--\blpage{311}
(\byear{2020})
\end{barticle}
\endbibitem

\bibitem[\protect\citeauthoryear{Cheng et~al.}{2017}]{cheng2017enterprise}
\begin{barticle}
\bauthor{\bsnm{Cheng}, \binits{L.}},
\bauthor{\bsnm{Liu}, \binits{F.}},
\bauthor{\bsnm{Yao}, \binits{D.}}:
\batitle{Enterprise data breach: causes, challenges, prevention, and future directions}.
\bjtitle{Wiley Interdisciplinary Reviews: Data Mining and Knowledge Discovery}
\bvolume{7}(\bissue{5}),
\bfpage{1211}
(\byear{2017})
\end{barticle}
\endbibitem

\bibitem[\protect\citeauthoryear{Verbraeken et~al.}{2020}]{verbraeken2020survey}
\begin{barticle}
\bauthor{\bsnm{Verbraeken}, \binits{J.}},
\bauthor{\bsnm{Wolting}, \binits{M.}},
\bauthor{\bsnm{Katzy}, \binits{J.}},
\bauthor{\bsnm{Kloppenburg}, \binits{J.}},
\bauthor{\bsnm{Verbelen}, \binits{T.}},
\bauthor{\bsnm{Rellermeyer}, \binits{J.S.}}:
\batitle{A survey on distributed machine learning}.
\bjtitle{Acm computing surveys (csur)}
\bvolume{53}(\bissue{2}),
\bfpage{1}--\blpage{33}
(\byear{2020})
\end{barticle}
\endbibitem

\bibitem[\protect\citeauthoryear{Paulson et~al.}{2022}]{paulson2022feature}
\begin{barticle}
\bauthor{\bsnm{Paulson}, \binits{N.H.}},
\bauthor{\bsnm{Kubal}, \binits{J.}},
\bauthor{\bsnm{Ward}, \binits{L.}},
\bauthor{\bsnm{Saxena}, \binits{S.}},
\bauthor{\bsnm{Lu}, \binits{W.}},
\bauthor{\bsnm{Babinec}, \binits{S.J.}}:
\batitle{Feature engineering for machine learning enabled early prediction of battery lifetime}.
\bjtitle{Journal of Power Sources}
\bvolume{527},
\bfpage{231127}
(\byear{2022})
\end{barticle}
\endbibitem

\bibitem[\protect\citeauthoryear{Guha et~al.}{2017}]{guha2017remaining}
\begin{bchapter}
\bauthor{\bsnm{Guha}, \binits{A.}},
\bauthor{\bsnm{Patra}, \binits{A.}},
\bauthor{\bsnm{Vaisakh}, \binits{K.}}:
\bctitle{Remaining useful life estimation of lithium-ion batteries based on the internal resistance growth model}.
In: \bbtitle{2017 Indian Control Conference (ICC)},
pp. \bfpage{33}--\blpage{38}
(\byear{2017}).
\bcomment{IEEE}
\end{bchapter}
\endbibitem

\bibitem[\protect\citeauthoryear{Duan et~al.}{2020}]{duan2020remaining}
\begin{barticle}
\bauthor{\bsnm{Duan}, \binits{B.}},
\bauthor{\bsnm{Zhang}, \binits{Q.}},
\bauthor{\bsnm{Geng}, \binits{F.}},
\bauthor{\bsnm{Zhang}, \binits{C.}}:
\batitle{Remaining useful life prediction of lithium-ion battery based on extended kalman particle filter}.
\bjtitle{International Journal of Energy Research}
\bvolume{44}(\bissue{3}),
\bfpage{1724}--\blpage{1734}
(\byear{2020})
\end{barticle}
\endbibitem

\bibitem[\protect\citeauthoryear{Walker et~al.}{2015}]{walker2015comparison}
\begin{barticle}
\bauthor{\bsnm{Walker}, \binits{E.}},
\bauthor{\bsnm{Rayman}, \binits{S.}},
\bauthor{\bsnm{White}, \binits{R.E.}}:
\batitle{Comparison of a particle filter and other state estimation methods for prognostics of lithium-ion batteries}.
\bjtitle{Journal of Power Sources}
\bvolume{287},
\bfpage{1}--\blpage{12}
(\byear{2015})
\end{barticle}
\endbibitem

\bibitem[\protect\citeauthoryear{Miao et~al.}{2013}]{miao2013remaining}
\begin{barticle}
\bauthor{\bsnm{Miao}, \binits{Q.}},
\bauthor{\bsnm{Xie}, \binits{L.}},
\bauthor{\bsnm{Cui}, \binits{H.}},
\bauthor{\bsnm{Liang}, \binits{W.}},
\bauthor{\bsnm{Pecht}, \binits{M.}}:
\batitle{Remaining useful life prediction of lithium-ion battery with unscented particle filter technique}.
\bjtitle{Microelectronics Reliability}
\bvolume{53}(\bissue{6}),
\bfpage{805}--\blpage{810}
(\byear{2013})
\end{barticle}
\endbibitem

\bibitem[\protect\citeauthoryear{Ramadass et~al.}{2004}]{ramadass2004development}
\begin{barticle}
\bauthor{\bsnm{Ramadass}, \binits{P.}},
\bauthor{\bsnm{Haran}, \binits{B.}},
\bauthor{\bsnm{Gomadam}, \binits{P.M.}},
\bauthor{\bsnm{White}, \binits{R.}},
\bauthor{\bsnm{Popov}, \binits{B.N.}}:
\batitle{Development of first principles capacity fade model for li-ion cells}.
\bjtitle{Journal of the Electrochemical Society}
\bvolume{151}(\bissue{2}),
\bfpage{196}
(\byear{2004})
\end{barticle}
\endbibitem

\bibitem[\protect\citeauthoryear{Zhang et~al.}{2017}]{zhang2017remaining}
\begin{bchapter}
\bauthor{\bsnm{Zhang}, \binits{D.}},
\bauthor{\bsnm{Dey}, \binits{S.}},
\bauthor{\bsnm{Perez}, \binits{H.E.}},
\bauthor{\bsnm{Moura}, \binits{S.J.}}:
\bctitle{Remaining useful life estimation of lithium-ion batteries based on thermal dynamics}.
In: \bbtitle{2017 American Control Conference (ACC)},
pp. \bfpage{4042}--\blpage{4047}
(\byear{2017}).
\bcomment{IEEE}
\end{bchapter}
\endbibitem

\bibitem[\protect\citeauthoryear{Guha and Patra}{2017}]{guha2017state}
\begin{barticle}
\bauthor{\bsnm{Guha}, \binits{A.}},
\bauthor{\bsnm{Patra}, \binits{A.}}:
\batitle{State of health estimation of lithium-ion batteries using capacity fade and internal resistance growth models}.
\bjtitle{IEEE Transactions on Transportation Electrification}
\bvolume{4}(\bissue{1}),
\bfpage{135}--\blpage{146}
(\byear{2017})
\end{barticle}
\endbibitem

\bibitem[\protect\citeauthoryear{Wang et~al.}{2017}]{wang2017remaining}
\begin{barticle}
\bauthor{\bsnm{Wang}, \binits{Y.}},
\bauthor{\bsnm{Pan}, \binits{R.}},
\bauthor{\bsnm{Yang}, \binits{D.}},
\bauthor{\bsnm{Tang}, \binits{X.}},
\bauthor{\bsnm{Chen}, \binits{Z.}}:
\batitle{Remaining useful life prediction of lithium-ion battery based on discrete wavelet transform}.
\bjtitle{Energy Procedia}
\bvolume{105},
\bfpage{2053}--\blpage{2058}
(\byear{2017})
\end{barticle}
\endbibitem

\bibitem[\protect\citeauthoryear{Ren et~al.}{2018}]{ren2018remaining}
\begin{barticle}
\bauthor{\bsnm{Ren}, \binits{L.}},
\bauthor{\bsnm{Zhao}, \binits{L.}},
\bauthor{\bsnm{Hong}, \binits{S.}},
\bauthor{\bsnm{Zhao}, \binits{S.}},
\bauthor{\bsnm{Wang}, \binits{H.}},
\bauthor{\bsnm{Zhang}, \binits{L.}}:
\batitle{Remaining useful life prediction for lithium-ion battery: A deep learning approach}.
\bjtitle{IEEE Access}
\bvolume{6},
\bfpage{50587}--\blpage{50598}
(\byear{2018})
\end{barticle}
\endbibitem

\bibitem[\protect\citeauthoryear{Zhang et~al.}{2018}]{zhang2018long}
\begin{barticle}
\bauthor{\bsnm{Zhang}, \binits{Y.}},
\bauthor{\bsnm{Xiong}, \binits{R.}},
\bauthor{\bsnm{He}, \binits{H.}},
\bauthor{\bsnm{Pecht}, \binits{M.G.}}:
\batitle{Long short-term memory recurrent neural network for remaining useful life prediction of lithium-ion batteries}.
\bjtitle{IEEE Transactions on Vehicular Technology}
\bvolume{67}(\bissue{7}),
\bfpage{5695}--\blpage{5705}
(\byear{2018})
\end{barticle}
\endbibitem

\bibitem[\protect\citeauthoryear{Li et~al.}{2020}]{li2020state}
\begin{barticle}
\bauthor{\bsnm{Li}, \binits{P.}},
\bauthor{\bsnm{Zhang}, \binits{Z.}},
\bauthor{\bsnm{Xiong}, \binits{Q.}},
\bauthor{\bsnm{Ding}, \binits{B.}},
\bauthor{\bsnm{Hou}, \binits{J.}},
\bauthor{\bsnm{Luo}, \binits{D.}},
\bauthor{\bsnm{Rong}, \binits{Y.}},
\bauthor{\bsnm{Li}, \binits{S.}}:
\batitle{State-of-health estimation and remaining useful life prediction for the lithium-ion battery based on a variant long short term memory neural network}.
\bjtitle{Journal of power sources}
\bvolume{459},
\bfpage{228069}
(\byear{2020})
\end{barticle}
\endbibitem

\bibitem[\protect\citeauthoryear{Patil et~al.}{2015}]{patil2015novel}
\begin{barticle}
\bauthor{\bsnm{Patil}, \binits{M.A.}},
\bauthor{\bsnm{Tagade}, \binits{P.}},
\bauthor{\bsnm{Hariharan}, \binits{K.S.}},
\bauthor{\bsnm{Kolake}, \binits{S.M.}},
\bauthor{\bsnm{Song}, \binits{T.}},
\bauthor{\bsnm{Yeo}, \binits{T.}},
\bauthor{\bsnm{Doo}, \binits{S.}}:
\batitle{A novel multistage support vector machine based approach for li ion battery remaining useful life estimation}.
\bjtitle{Applied energy}
\bvolume{159},
\bfpage{285}--\blpage{297}
(\byear{2015})
\end{barticle}
\endbibitem

\bibitem[\protect\citeauthoryear{Saha et~al.}{2009}]{saha2009comparison}
\begin{barticle}
\bauthor{\bsnm{Saha}, \binits{B.}},
\bauthor{\bsnm{Goebel}, \binits{K.}},
\bauthor{\bsnm{Christophersen}, \binits{J.}}:
\batitle{Comparison of prognostic algorithms for estimating remaining useful life of batteries}.
\bjtitle{Transactions of the Institute of Measurement and Control}
\bvolume{31}(\bissue{3-4}),
\bfpage{293}--\blpage{308}
(\byear{2009})
\end{barticle}
\endbibitem

\bibitem[\protect\citeauthoryear{Li et~al.}{2019}]{li2019remaining}
\begin{barticle}
\bauthor{\bsnm{Li}, \binits{X.}},
\bauthor{\bsnm{Zhang}, \binits{L.}},
\bauthor{\bsnm{Wang}, \binits{Z.}},
\bauthor{\bsnm{Dong}, \binits{P.}}:
\batitle{Remaining useful life prediction for lithium-ion batteries based on a hybrid model combining the long short-term memory and elman neural networks}.
\bjtitle{Journal of Energy Storage}
\bvolume{21},
\bfpage{510}--\blpage{518}
(\byear{2019})
\end{barticle}
\endbibitem

\bibitem[\protect\citeauthoryear{Xue et~al.}{2020}]{xue2020remaining}
\begin{barticle}
\bauthor{\bsnm{Xue}, \binits{Z.}},
\bauthor{\bsnm{Zhang}, \binits{Y.}},
\bauthor{\bsnm{Cheng}, \binits{C.}},
\bauthor{\bsnm{Ma}, \binits{G.}}:
\batitle{Remaining useful life prediction of lithium-ion batteries with adaptive unscented kalman filter and optimized support vector regression}.
\bjtitle{Neurocomputing}
\bvolume{376},
\bfpage{95}--\blpage{102}
(\byear{2020})
\end{barticle}
\endbibitem

\bibitem[\protect\citeauthoryear{Zheng and Fang}{2015}]{zheng2015integrated}
\begin{barticle}
\bauthor{\bsnm{Zheng}, \binits{X.}},
\bauthor{\bsnm{Fang}, \binits{H.}}:
\batitle{An integrated unscented kalman filter and relevance vector regression approach for lithium-ion battery remaining useful life and short-term capacity prediction}.
\bjtitle{Reliability Engineering \& System Safety}
\bvolume{144},
\bfpage{74}--\blpage{82}
(\byear{2015})
\end{barticle}
\endbibitem

\bibitem[\protect\citeauthoryear{Hu et~al.}{2020}]{hu2020battery}
\begin{barticle}
\bauthor{\bsnm{Hu}, \binits{X.}},
\bauthor{\bsnm{Xu}, \binits{L.}},
\bauthor{\bsnm{Lin}, \binits{X.}},
\bauthor{\bsnm{Pecht}, \binits{M.}}:
\batitle{Battery lifetime prognostics}.
\bjtitle{Joule}
\bvolume{4}(\bissue{2}),
\bfpage{310}--\blpage{346}
(\byear{2020})
\end{barticle}
\endbibitem

\bibitem[\protect\citeauthoryear{Li et~al.}{2019}]{li2019data}
\begin{barticle}
\bauthor{\bsnm{Li}, \binits{Y.}},
\bauthor{\bsnm{Liu}, \binits{K.}},
\bauthor{\bsnm{Foley}, \binits{A.M.}},
\bauthor{\bsnm{Z{\"u}lke}, \binits{A.}},
\bauthor{\bsnm{Berecibar}, \binits{M.}},
\bauthor{\bsnm{Nanini-Maury}, \binits{E.}},
\bauthor{\bsnm{Van~Mierlo}, \binits{J.}},
\bauthor{\bsnm{Hoster}, \binits{H.E.}}:
\batitle{Data-driven health estimation and lifetime prediction of lithium-ion batteries: A review}.
\bjtitle{Renewable and sustainable energy reviews}
\bvolume{113},
\bfpage{109254}
(\byear{2019})
\end{barticle}
\endbibitem

\bibitem[\protect\citeauthoryear{Lipu et~al.}{2018}]{lipu2018review}
\begin{barticle}
\bauthor{\bsnm{Lipu}, \binits{M.H.}},
\bauthor{\bsnm{Hannan}, \binits{M.}},
\bauthor{\bsnm{Hussain}, \binits{A.}},
\bauthor{\bsnm{Hoque}, \binits{M.}},
\bauthor{\bsnm{Ker}, \binits{P.J.}},
\bauthor{\bsnm{Saad}, \binits{M.M.}},
\bauthor{\bsnm{Ayob}, \binits{A.}}:
\batitle{A review of state of health and remaining useful life estimation methods for lithium-ion battery in electric vehicles: Challenges and recommendations}.
\bjtitle{Journal of cleaner production}
\bvolume{205},
\bfpage{115}--\blpage{133}
(\byear{2018})
\end{barticle}
\endbibitem

\bibitem[\protect\citeauthoryear{Meng and Li}{2019}]{meng2019review}
\begin{barticle}
\bauthor{\bsnm{Meng}, \binits{H.}},
\bauthor{\bsnm{Li}, \binits{Y.-F.}}:
\batitle{A review on prognostics and health management (phm) methods of lithium-ion batteries}.
\bjtitle{Renewable and Sustainable Energy Reviews}
\bvolume{116},
\bfpage{109405}
(\byear{2019})
\end{barticle}
\endbibitem

\bibitem[\protect\citeauthoryear{Ng et~al.}{2020}]{ng2020predicting}
\begin{barticle}
\bauthor{\bsnm{Ng}, \binits{M.-F.}},
\bauthor{\bsnm{Zhao}, \binits{J.}},
\bauthor{\bsnm{Yan}, \binits{Q.}},
\bauthor{\bsnm{Conduit}, \binits{G.J.}},
\bauthor{\bsnm{Seh}, \binits{Z.W.}}:
\batitle{Predicting the state of charge and health of batteries using data-driven machine learning}.
\bjtitle{Nature Machine Intelligence}
\bvolume{2}(\bissue{3}),
\bfpage{161}--\blpage{170}
(\byear{2020})
\end{barticle}
\endbibitem

\bibitem[\protect\citeauthoryear{Kairouz et~al.}{2021}]{kairouz2021advances}
\begin{barticle}
\bauthor{\bsnm{Kairouz}, \binits{P.}},
\bauthor{\bsnm{McMahan}, \binits{H.B.}},
\bauthor{\bsnm{Avent}, \binits{B.}},
\bauthor{\bsnm{Bellet}, \binits{A.}},
\bauthor{\bsnm{Bennis}, \binits{M.}},
\bauthor{\bsnm{Bhagoji}, \binits{A.N.}},
\bauthor{\bsnm{Bonawitz}, \binits{K.}},
\bauthor{\bsnm{Charles}, \binits{Z.}},
\bauthor{\bsnm{Cormode}, \binits{G.}},
\bauthor{\bsnm{Cummings}, \binits{R.}}, \betal:
\batitle{Advances and open problems in federated learning}.
\bjtitle{Foundations and Trends{\textregistered} in Machine Learning}
\bvolume{14}(\bissue{1--2}),
\bfpage{1}--\blpage{210}
(\byear{2021})
\end{barticle}
\endbibitem

\bibitem[\protect\citeauthoryear{Hard et~al.}{2018}]{hard2018federated}
\begin{botherref}
\oauthor{\bsnm{Hard}, \binits{A.}},
\oauthor{\bsnm{Rao}, \binits{K.}},
\oauthor{\bsnm{Mathews}, \binits{R.}},
\oauthor{\bsnm{Ramaswamy}, \binits{S.}},
\oauthor{\bsnm{Beaufays}, \binits{F.}},
\oauthor{\bsnm{Augenstein}, \binits{S.}},
\oauthor{\bsnm{Eichner}, \binits{H.}},
\oauthor{\bsnm{Kiddon}, \binits{C.}},
\oauthor{\bsnm{Ramage}, \binits{D.}}:
Federated learning for mobile keyboard prediction.
arXiv preprint arXiv:1811.03604
(2018)
\end{botherref}
\endbibitem

\bibitem[\protect\citeauthoryear{Rieke et~al.}{2020}]{rieke2020future}
\begin{barticle}
\bauthor{\bsnm{Rieke}, \binits{N.}},
\bauthor{\bsnm{Hancox}, \binits{J.}},
\bauthor{\bsnm{Li}, \binits{W.}},
\bauthor{\bsnm{Milletari}, \binits{F.}},
\bauthor{\bsnm{Roth}, \binits{H.R.}},
\bauthor{\bsnm{Albarqouni}, \binits{S.}},
\bauthor{\bsnm{Bakas}, \binits{S.}},
\bauthor{\bsnm{Galtier}, \binits{M.N.}},
\bauthor{\bsnm{Landman}, \binits{B.A.}},
\bauthor{\bsnm{Maier-Hein}, \binits{K.}}, \betal:
\batitle{The future of digital health with federated learning}.
\bjtitle{NPJ digital medicine}
\bvolume{3}(\bissue{1}),
\bfpage{1}--\blpage{7}
(\byear{2020})
\end{barticle}
\endbibitem

\bibitem[\protect\citeauthoryear{Pfitzner et~al.}{2021}]{pfitzner2021federated}
\begin{barticle}
\bauthor{\bsnm{Pfitzner}, \binits{B.}},
\bauthor{\bsnm{Steckhan}, \binits{N.}},
\bauthor{\bsnm{Arnrich}, \binits{B.}}:
\batitle{Federated learning in a medical context: a systematic literature review}.
\bjtitle{ACM Transactions on Internet Technology (TOIT)}
\bvolume{21}(\bissue{2}),
\bfpage{1}--\blpage{31}
(\byear{2021})
\end{barticle}
\endbibitem

\bibitem[\protect\citeauthoryear{Huong et~al.}{2021}]{huong2021detecting}
\begin{barticle}
\bauthor{\bsnm{Huong}, \binits{T.T.}},
\bauthor{\bsnm{Bac}, \binits{T.P.}},
\bauthor{\bsnm{Long}, \binits{D.M.}},
\bauthor{\bsnm{Luong}, \binits{T.D.}},
\bauthor{\bsnm{Dan}, \binits{N.M.}},
\bauthor{\bsnm{Thang}, \binits{B.D.}},
\bauthor{\bsnm{Tran}, \binits{K.P.}}, \betal:
\batitle{Detecting cyberattacks using anomaly detection in industrial control systems: A federated learning approach}.
\bjtitle{Computers in Industry}
\bvolume{132},
\bfpage{103509}
(\byear{2021})
\end{barticle}
\endbibitem

\bibitem[\protect\citeauthoryear{Ge et~al.}{2021}]{ge2021failure}
\begin{botherref}
\oauthor{\bsnm{Ge}, \binits{N.}},
\oauthor{\bsnm{Li}, \binits{G.}},
\oauthor{\bsnm{Zhang}, \binits{L.}},
\oauthor{\bsnm{Liu}, \binits{Y.}}:
Failure prediction in production line based on federated learning: an empirical study.
Journal of Intelligent Manufacturing,
1--18
(2021)
\end{botherref}
\endbibitem

\bibitem[\protect\citeauthoryear{Mehta and Shao}{2022}]{mehta2022federated}
\begin{barticle}
\bauthor{\bsnm{Mehta}, \binits{M.}},
\bauthor{\bsnm{Shao}, \binits{C.}}:
\batitle{Federated learning-based semantic segmentation for pixel-wise defect detection in additive manufacturing}.
\bjtitle{Journal of Manufacturing Systems}
\bvolume{64},
\bfpage{197}--\blpage{210}
(\byear{2022})
\end{barticle}
\endbibitem

\bibitem[\protect\citeauthoryear{Saputra et~al.}{2019}]{saputra2019energy}
\begin{bchapter}
\bauthor{\bsnm{Saputra}, \binits{Y.M.}},
\bauthor{\bsnm{Hoang}, \binits{D.T.}},
\bauthor{\bsnm{Nguyen}, \binits{D.N.}},
\bauthor{\bsnm{Dutkiewicz}, \binits{E.}},
\bauthor{\bsnm{Mueck}, \binits{M.D.}},
\bauthor{\bsnm{Srikanteswara}, \binits{S.}}:
\bctitle{Energy demand prediction with federated learning for electric vehicle networks}.
In: \bbtitle{2019 IEEE Global Communications Conference (GLOBECOM)},
pp. \bfpage{1}--\blpage{6}
(\byear{2019}).
\bcomment{IEEE}
\end{bchapter}
\endbibitem

\bibitem[\protect\citeauthoryear{Savi and Olivadese}{2021}]{savi2021short}
\begin{barticle}
\bauthor{\bsnm{Savi}, \binits{M.}},
\bauthor{\bsnm{Olivadese}, \binits{F.}}:
\batitle{Short-term energy consumption forecasting at the edge: A federated learning approach}.
\bjtitle{IEEE Access}
\bvolume{9},
\bfpage{95949}--\blpage{95969}
(\byear{2021})
\end{barticle}
\endbibitem

\bibitem[\protect\citeauthoryear{Grammenos et~al.}{2020}]{grammenos2020federated}
\begin{barticle}
\bauthor{\bsnm{Grammenos}, \binits{A.}},
\bauthor{\bsnm{Mendoza~Smith}, \binits{R.}},
\bauthor{\bsnm{Crowcroft}, \binits{J.}},
\bauthor{\bsnm{Mascolo}, \binits{C.}}:
\batitle{Federated principal component analysis}.
\bjtitle{Advances in Neural Information Processing Systems}
\bvolume{33},
\bfpage{6453}--\blpage{6464}
(\byear{2020})
\end{barticle}
\endbibitem

\bibitem[\protect\citeauthoryear{Novoa-Paradela et~al.}{2022}]{novoa2022fast}
\begin{botherref}
\oauthor{\bsnm{Novoa-Paradela}, \binits{D.}},
\oauthor{\bsnm{Romero-Fontenla}, \binits{O.}},
\oauthor{\bsnm{Guijarro-Berdi{\~n}as}, \binits{B.}}:
Fast deep autoencoder for federated learning.
arXiv preprint arXiv:2206.05136
(2022)
\end{botherref}
\endbibitem

\bibitem[\protect\citeauthoryear{Banerjee et~al.}{2021}]{banerjee2021fed}
\begin{bchapter}
\bauthor{\bsnm{Banerjee}, \binits{S.}},
\bauthor{\bsnm{Elmroth}, \binits{E.}},
\bauthor{\bsnm{Bhuyan}, \binits{M.}}:
\bctitle{Fed-fis: a novel information-theoretic federated feature selection for learning stability}.
In: \bbtitle{International Conference on Neural Information Processing},
pp. \bfpage{480}--\blpage{487}
(\byear{2021}).
\bcomment{Springer}
\end{bchapter}
\endbibitem

\bibitem[\protect\citeauthoryear{Wang et~al.}{}]{wangsecure}
\begin{botherref}
\oauthor{\bsnm{Wang}, \binits{L.}},
\oauthor{\bsnm{Pang}, \binits{Q.}},
\oauthor{\bsnm{Wang}, \binits{S.}},
\oauthor{\bsnm{Song}, \binits{D.}}:
Secure federated feature selection
\end{botherref}
\endbibitem

\bibitem[\protect\citeauthoryear{Hu et~al.}{2022}]{hu2022multi}
\begin{botherref}
\oauthor{\bsnm{Hu}, \binits{Y.}},
\oauthor{\bsnm{Zhang}, \binits{Y.}},
\oauthor{\bsnm{Gong}, \binits{D.}},
\oauthor{\bsnm{Sun}, \binits{X.}}:
Multi-participant federated feature selection algorithm with particle swarm optimizaiton for imbalanced data under privacy protection.
IEEE Transactions on Artificial Intelligence
(2022)
\end{botherref}
\endbibitem

\bibitem[\protect\citeauthoryear{Gao et~al.}{2021}]{gao2021federated}
\begin{barticle}
\bauthor{\bsnm{Gao}, \binits{Y.}},
\bauthor{\bsnm{Zhang}, \binits{G.}},
\bauthor{\bsnm{Zhang}, \binits{C.}},
\bauthor{\bsnm{Wang}, \binits{J.}},
\bauthor{\bsnm{Yang}, \binits{L.T.}},
\bauthor{\bsnm{Zhao}, \binits{Y.}}:
\batitle{Federated tensor decomposition-based feature extraction approach for industrial iot}.
\bjtitle{IEEE Transactions on Industrial Informatics}
\bvolume{17}(\bissue{12}),
\bfpage{8541}--\blpage{8549}
(\byear{2021})
\end{barticle}
\endbibitem

\bibitem[\protect\citeauthoryear{Wang et~al.}{2021}]{wang2021electricity}
\begin{barticle}
\bauthor{\bsnm{Wang}, \binits{Y.}},
\bauthor{\bsnm{Bennani}, \binits{I.L.}},
\bauthor{\bsnm{Liu}, \binits{X.}},
\bauthor{\bsnm{Sun}, \binits{M.}},
\bauthor{\bsnm{Zhou}, \binits{Y.}}:
\batitle{Electricity consumer characteristics identification: A federated learning approach}.
\bjtitle{IEEE Transactions on Smart Grid}
\bvolume{12}(\bissue{4}),
\bfpage{3637}--\blpage{3647}
(\byear{2021})
\end{barticle}
\endbibitem

\bibitem[\protect\citeauthoryear{Cassar{\'a} et~al.}{2022}]{cassara2022federated}
\begin{botherref}
\oauthor{\bsnm{Cassar{\'a}}, \binits{P.}},
\oauthor{\bsnm{Gotta}, \binits{A.}},
\oauthor{\bsnm{Valerio}, \binits{L.}}:
Federated feature selection for cyber-physical systems of systems.
IEEE Transactions on Vehicular Technology
(2022)
\end{botherref}
\endbibitem

\bibitem[\protect\citeauthoryear{Xiao et~al.}{2021}]{xiao2021federated}
\begin{barticle}
\bauthor{\bsnm{Xiao}, \binits{Z.}},
\bauthor{\bsnm{Xu}, \binits{X.}},
\bauthor{\bsnm{Xing}, \binits{H.}},
\bauthor{\bsnm{Song}, \binits{F.}},
\bauthor{\bsnm{Wang}, \binits{X.}},
\bauthor{\bsnm{Zhao}, \binits{B.}}:
\batitle{A federated learning system with enhanced feature extraction for human activity recognition}.
\bjtitle{Knowledge-Based Systems}
\bvolume{229},
\bfpage{107338}
(\byear{2021})
\end{barticle}
\endbibitem

\bibitem[\protect\citeauthoryear{Gebraeel}{2006}]{gebraeel2006sensory}
\begin{barticle}
\bauthor{\bsnm{Gebraeel}, \binits{N.}}:
\batitle{Sensory-updated residual life distributions for components with exponential degradation patterns}.
\bjtitle{IEEE transactions on automation science and engineering}
\bvolume{3}(\bissue{4}),
\bfpage{382}--\blpage{393}
(\byear{2006})
\end{barticle}
\endbibitem

\bibitem[\protect\citeauthoryear{Elwany and Gebraeel}{2008}]{elwany2008sensor}
\begin{barticle}
\bauthor{\bsnm{Elwany}, \binits{A.H.}},
\bauthor{\bsnm{Gebraeel}, \binits{N.Z.}}:
\batitle{Sensor-driven prognostic models for equipment replacement and spare parts inventory}.
\bjtitle{IIe Transactions}
\bvolume{40}(\bissue{7}),
\bfpage{629}--\blpage{639}
(\byear{2008})
\end{barticle}
\endbibitem

\bibitem[\protect\citeauthoryear{Ward et~al.}{2021}]{git}
\begin{botherref}
\oauthor{\bsnm{Ward}, \binits{L.}},
\oauthor{\bsnm{Kubal}, \binits{J.}},
\oauthor{\bsnm{Blaiszik}, \binits{B.}},
\oauthor{\bsnm{Paulson}, \binits{N.}}:
Battery Data Toolkit
(2021).
\url{https://github.com/materials-data-facility/battery-data-toolkit}
\end{botherref}
\endbibitem

\bibitem[\protect\citeauthoryear{Diao et~al.}{2019}]{diao2019algorithm}
\begin{barticle}
\bauthor{\bsnm{Diao}, \binits{W.}},
\bauthor{\bsnm{Saxena}, \binits{S.}},
\bauthor{\bsnm{Han}, \binits{B.}},
\bauthor{\bsnm{Pecht}, \binits{M.}}:
\batitle{Algorithm to determine the knee point on capacity fade curves of lithium-ion cells}.
\bjtitle{Energies}
\bvolume{12}(\bissue{15}),
\bfpage{2910}
(\byear{2019})
\end{barticle}
\endbibitem

\bibitem[\protect\citeauthoryear{Saxena et~al.}{2022}]{saxena2022convolutional}
\begin{barticle}
\bauthor{\bsnm{Saxena}, \binits{S.}},
\bauthor{\bsnm{Ward}, \binits{L.}},
\bauthor{\bsnm{Kubal}, \binits{J.}},
\bauthor{\bsnm{Lu}, \binits{W.}},
\bauthor{\bsnm{Babinec}, \binits{S.}},
\bauthor{\bsnm{Paulson}, \binits{N.}}:
\batitle{A convolutional neural network model for battery capacity fade curve prediction using early life data}.
\bjtitle{Journal of Power Sources}
\bvolume{542},
\bfpage{231736}
(\byear{2022})
\end{barticle}
\endbibitem

\bibitem[\protect\citeauthoryear{Honkura et~al.}{2011}]{honkura2011capacity}
\begin{barticle}
\bauthor{\bsnm{Honkura}, \binits{K.}},
\bauthor{\bsnm{Takahashi}, \binits{K.}},
\bauthor{\bsnm{Horiba}, \binits{T.}}:
\batitle{Capacity-fading prediction of lithium-ion batteries based on discharge curves analysis}.
\bjtitle{Journal of power sources}
\bvolume{196}(\bissue{23}),
\bfpage{10141}--\blpage{10147}
(\byear{2011})
\end{barticle}
\endbibitem

\end{thebibliography}

\end{document}